\documentclass[a4paper]{cas-sc}

\usepackage[style=apa,backend=biber]{biblatex}
\DeclareLanguageMapping{english}{english-apa}
\usepackage{graphicx}

\usepackage{amsmath,amssymb,amsfonts}
\usepackage{dsfont}

\usepackage{booktabs}
\usepackage{tabularx}
\usepackage{multirow}
\usepackage{longtable}

\usepackage{algorithm}
\usepackage{algpseudocode}

\usepackage{caption}
\usepackage{subcaption}
\usepackage{placeins} 

\usepackage[dvipsnames]{xcolor}

\usepackage[title]{appendix}

\usepackage{hyperref}
\usepackage{tikz}
\usetikzlibrary{arrows.meta, positioning, calc, backgrounds}
\DeclareMathAlphabet{\pazocal}{OMS}{zplm}{m}{n}

\newcommand{\Cp}{\pazocal{C}}
\newcommand{\Lp}{\pazocal{L}}
\newcommand{\Xp}{\pazocal{X}}
\newcommand{\Fp}{\pazocal{F}}

\def\tsc#1{\csdef{#1}{\textsc{\lowercase{#1}}\xspace}}
\tsc{WGM}
\tsc{QE}

\definecolor{inputtop}{RGB}{90,90,90}
\definecolor{inputbot}{RGB}{230,230,230}

\definecolor{ensblueT}{RGB}{120, 170, 225}
\definecolor{ensblueB}{RGB}{210, 230, 245}
\definecolor{ensborder}{RGB}{40, 100, 180}

\definecolor{silgreentop}{RGB}{45, 135, 65}
\definecolor{silgreenbot}{RGB}{205, 240, 175}

\definecolor{laborangetop}{RGB}{215, 105, 25}
\definecolor{laborangebot}{RGB}{255, 215, 110}

\definecolor{cakepurpletop}{RGB}{120, 85, 170}
\definecolor{cakepurplebot}{RGB}{225, 220, 240}

\newcommand{\drawmat}[8]{
    \begin{scope}[shift={(#1,#2)}]
        \fill[top color=#6, bottom color=#7] (0,0) rectangle (#4,#5);
        \draw[xstep=#4/#3, ystep=0.3, very thin, draw=#8!70] (0,0) grid (#4,#5);
        \draw[thick, draw=#8] (0,0) rectangle (#4,#5);
    \end{scope}
}

\begin{document}

\shorttitle{\textit{CAKE: Confidence in Assignments via K-partition Ensembles}}    

\shortauthors{\textit{A. Semoglou and J. Pavlopoulos}}  

\title [mode = title]{CAKE: Confidence in Assignments via K-partition Ensembles}  

\affiliation[1]{organization={Department of Informatics, Athens University of Economics and Business},
            city={Athens},
            country={Greece}}

\affiliation[2]{organization={Archimedes Research Unit, Athena Research Center},
            city={Athens},
            country={Greece}}

\author[1,2]{Aggelos Semoglou}[orcid=0009-0005-3715-530X]
\cormark[1]
\ead{ang.semoglou@aueb.gr}

\author[1,2]{John Pavlopoulos}[orcid=0000-0001-9188-7425]
\ead{annis@aueb.gr}

\cortext[1]{Corresponding author}

\begin{abstract}
Clustering is widely used for unsupervised structure discovery, yet it offers limited insight into how reliable each individual assignment is. Diagnostics, such as convergence behavior or objective values, may reflect global quality, but they do not indicate whether particular instances are assigned confidently, especially for initialization-sensitive algorithms like \emph{k}-means. This assignment-level instability can undermine both accuracy and robustness. Ensemble approaches improve global consistency by aggregating multiple runs, but they typically lack tools for quantifying pointwise confidence in a way that combines cross-run agreement with geometric support from the learned cluster structure. This work introduces \textbf{CAKE} (Confidence in Assignments via K-partition Ensembles), a framework that evaluates each point using two complementary statistics computed over a clustering ensemble: assignment stability and consistency of local geometric fit. These are combined into a single, interpretable score in $[0,1]$. The theoretical analysis shows that CAKE remains effective under noise and separates stable from unstable points. Experiments on synthetic and real-world datasets indicate that CAKE effectively highlights ambiguous points and stable core members, providing a confidence ranking over instances that can be used for selection or prioritization in downstream clustering workflows.
\end{abstract}

\begin{keywords}
 Unsupervised learning \sep Ensemble learning \sep Clustering ensembles \sep Confidence estimation \sep Error detection \sep \medskip \textit{Software}:\sep \url{https://github.com/semoglou/cake} \sep \url{https://pypi.org/project/cake-ensemble/} \sep 
 \medskip \textit{Published in}: \sep Machine Learning with Applications, Vol. 24, 2026 \sep
 DOI: \url{https://doi.org/10.1016/j.mlwa.2026.100915} \sep License: CC BY-NC-ND 4.0
\end{keywords}

\maketitle

\section{Introduction}
\noindent Clustering is a fundamental task in unsupervised machine learning, widely used to uncover structure in unlabeled data~\parencite{aggarwal2013data, jain1999data}. It has core applications in pattern discovery, exploratory analysis, and decision-making across both scientific and applied domains~\parencite{kaufman2009finding}. However, unlike supervised learning, where confidence estimation techniques such as conformal prediction~\parencite{shafer2008tutorial} or calibrated probabilities~\parencite{guo2017calibration} are common, clustering methods typically do not offer reliable confidence scores for individual data point assignments. This makes it difficult to assess assignment trustworthiness, which is particularly important when downstream decisions rely on clustering results.

\medskip

\noindent In practice, clustering algorithms often exhibit sensitivity to initialization, susceptibility to local optima, and vulnerability to noisy or ambiguous data~\parencite{xu2005survey}. Consequently, outcomes may vary across executions, even under fixed algorithmic settings. For example, \emph{k}-means~\parencite{kmeans}, despite its popularity, can produce divergent results across runs due to random initialization~\parencite{bubeck2009}. Practitioners often mitigate this by running the algorithm multiple times with different random seeds and selecting the best result based on internal validation metrics. However, this approach only addresses global variability and offers no insight into the reliability of individual point assignments. Such assignment-level instability is particularly problematic in settings like anomaly detection or scientific discovery, where unreliable assignments can mislead interpretation or obscure important patterns.
Moreover, most existing evaluation methods assess clustering quality at a global or cluster level, using objective values or internal validation metrics such as the Silhouette score or Davies–Bouldin index~\parencite{arbelaitz2013extensive, vendramin2009relative}.

\medskip

\noindent 
Ensemble-based techniques like consensus clustering~\parencite{aktas2024,strehl2002,zhang2022} have also been explored to improve robustness by aggregating multiple partitions. However, they still lack interpretable, pointwise confidence scores that capture both cross-run agreement in label assignments and geometric fit. In response, many ensemble-style uncertainty heuristics focus primarily on \emph{agreement}, for example, counting aligned ``votes'' for each point's label across runs or measuring dispersion of assignments, to identify unstable samples~\parencite{ink2022,zhang25}. While useful, agreement alone does not reveal whether a point is \emph{geometrically} well supported by its assigned cluster, since a point can be consistently assigned due to a systematic bias or an overly rigid decision boundary even when it is isolated or weakly connected to the cluster structure. Conversely, purely geometric signals (such as the Silhouette score) computed within a single run can be overly optimistic in ambiguous regions: a point may appear well placed under that run’s geometry, yet still switch clusters across runs because it lies near a boundary or because multiple partitions explain it nearly equally well~\parencite{liu2022}. These complementary failure modes (Fig.~\ref{fig:failures}) motivate a confidence signal that fuses stability and geometry at the point level.

\begin{center}
    \includegraphics[width=\linewidth]{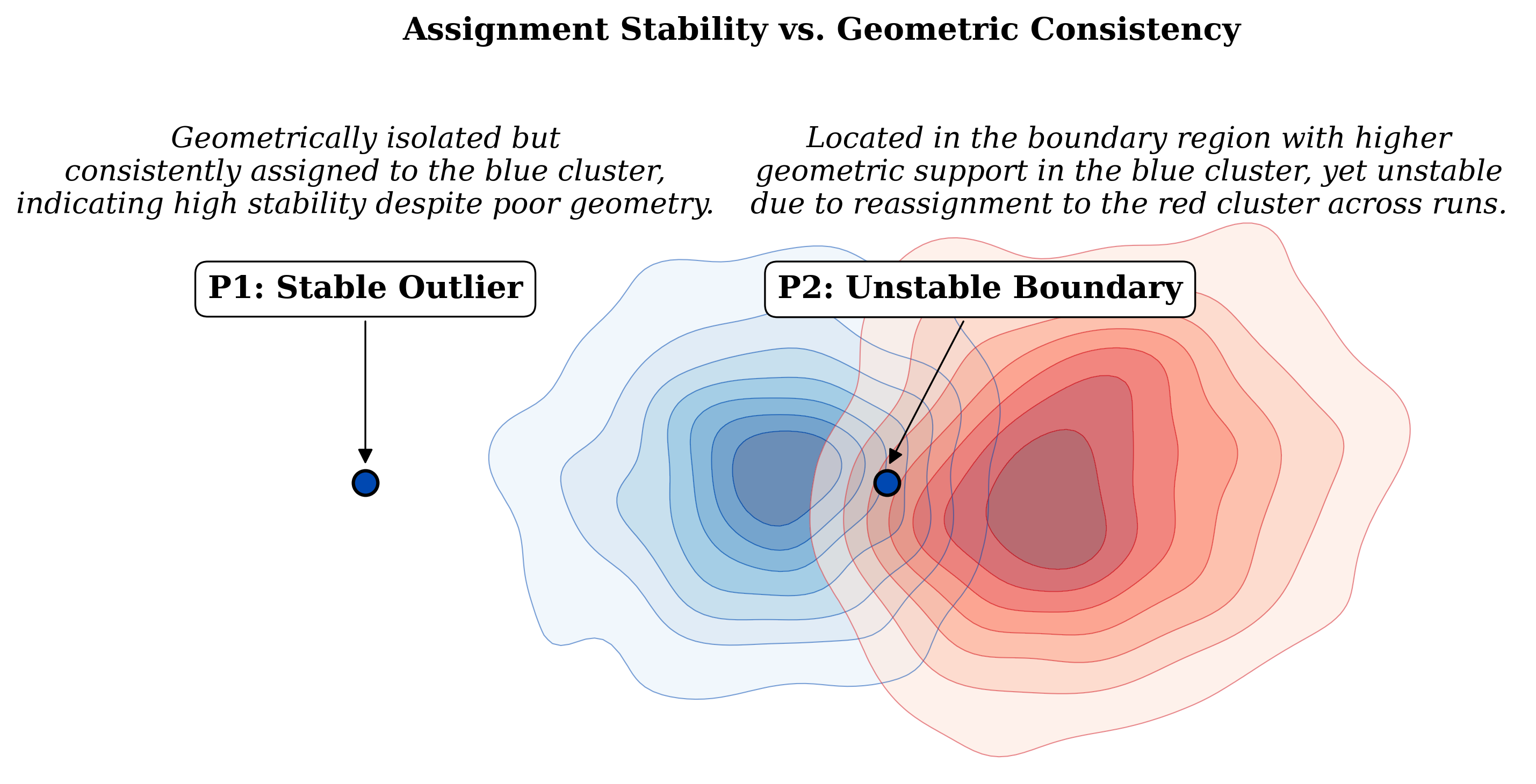}
    \captionsetup{hypcap=false}
    \captionof{figure}{Assignment stability and geometric consistency can fail in complementary ways. P1 (stable outlier; left) is consistently assigned (high stability) despite weak integration into its assigned cluster structure, while P2 (unstable boundary; right) shows higher within-cluster fit to one cluster in a single run yet switches labels across runs near the boundary. These cases suggest that reliable pointwise confidence should account for both signals jointly.}
    \label{fig:failures}
\end{center}

\noindent This raises a central question:

\medskip
\noindent\textit{How can each data point be assigned a confidence score that, across runs, reflects both the assignment stability and the consistency of its geometric support under the learned clustering structure?}

\medskip
\noindent To address this question, \textbf{CAKE} (Confidence in Assignments via K-partition Ensembles) is introduced as a framework that quantifies per-point confidence by combining two complementary statistics derived from an ensemble of clustering partitions (Fig.~\ref{fig:pipeline}): (\textit{i}) pairwise assignment agreement, computed via optimal label alignment using the Hungarian algorithm~\parencite{kuhn1955hungarian,meila2007}, and (\textit{ii}) local geometric consistency, measured through aggregated Silhouette statistics. These components are integrated into a single confidence score in $[0,1]$ for each point, indicating how strongly its cluster membership is supported and providing a fine-grained ranking that highlights ambiguous points and stable core members.
The paper provides theoretical guarantees for the statistical reliability of CAKE scores and demonstrates, across synthetic and real-world datasets, that CAKE effectively distinguishes between high- and low-confidence assignments. This enables selective removal or prioritization of points, improving clustering quality and interpretability in a fully label-free setting. Empirically, CAKE stabilizes with a modest number of runs, yielding a reliable confidence ranking without requiring large ensembles. This keeps the method lightweight and practical in iterative, exploratory clustering workflows. CAKE introduces a principled framework for per-instance confidence estimation in clustering, bridging ensemble diversity with pointwise assessment of assignment quality. More broadly, CAKE turns clustering ensembles into a practical, per-point diagnostic that complements global validation. 

\medskip

\noindent The CAKE implementation and experimental code are publicly available at \url{https://github.com/semoglou/cake}.

\noindent The Python package is available on PyPI at \url{https://pypi.org/project/cake-ensemble/}.

\begin{center}
    \begin{tikzpicture}[
        font=\sffamily,
        scale=0.88, 
        every node/.style={transform shape},
        >={Triangle[width=2.5mm,length=3mm]},
        arr/.style={->, line width=1.2pt, draw=black!55},
        arrG/.style={->, line width=1.2pt, draw=silgreentop},
        arrO/.style={->, line width=1.2pt, draw=laborangetop}
    ]
        \drawmat{0}{-1.05}{4}{1.2}{2.1}{inputtop}{inputbot}{black}
        \node[align=center] at (0.6, 1.8) {\small \textbf{Input} $\mathcal{X}$};
        \node at (0.6, -1.6) {\Large $\mathbb{R}^{n \times d}$};
        \draw[arr] (1.3, 0) -- (2.3, 0);

        \drawmat{2.9}{-0.75}{6}{1.8}{2.1}{ensblueT}{ensblueB}{ensborder}
        \drawmat{2.7}{-0.90}{6}{1.8}{2.1}{ensblueT}{ensblueB}{ensborder}
        \drawmat{2.5}{-1.05}{6}{1.8}{2.1}{ensblueT}{ensblueB}{ensborder}
        \node[align=center] at (3.4, 2.3) {\small \textbf{Clustering}\\\small\textbf{Ensemble}};
        \node at (3.4, -1.6) {$R$ Partitions};
        \draw[arr] (4.5, 0) -- (5.8, 2.3);
        \draw[arr] (4.5, 0) -- (5.8, -2.3);

        \drawmat{6}{1.25}{5}{1.5}{2.1}{silgreentop}{silgreenbot}{black}
        \node at (6.75, 3.8) {\small \textbf{Silhouette statistics}};
        \node at (6.75, 0.7) {\Large $\mathbb{R}^{n \times R}$};
        \draw[arr] (7.7, 2.3) -- (8.4, 2.3);
        \node[silgreentop] at (9.5, 3.8) {\small \textbf{Aggregation}};
        \draw[thick, draw=silgreentop, fill=silgreenbot!20] (9.5, 2.3) circle (0.9);
        \draw[very thick, draw=silgreentop] (8.8, 2.1) -- (10.2, 2.1); 
        \draw[very thick, draw=silgreentop] (8.9, 2.1)
            .. controls (9.2, 2.1) and (9.4, 2.8) .. (9.5, 2.8)
            .. controls (9.6, 2.8) and (9.8, 2.1) .. (10.1, 2.1);
        \draw[arr] (10.6, 2.3) -- (11.8, 2.3);
        \drawmat{12}{1.25}{1}{0.6}{2.1}{silgreentop}{silgreenbot}{black}
        \node at (12.3, 3.8) {\small \textbf{Geometric component}};
        \node at (12.3, 0.7) {\Large $\mathbb{R}^{n}$};

        \drawmat{6}{-3.35}{5}{1.5}{2.1}{laborangetop}{laborangebot}{black}
        \node at (6.75, -0.7) {\small \textbf{Label assignments}};
        \node at (6.75, -3.8) {\Large $\mathbb{R}^{n \times R}$};
        \draw[arr] (7.7, -2.3) -- (8.4, -2.3);
        \node[laborangetop] at (9.5, -0.7) {\small \textbf{Alignment}};
        \draw[thick, draw=laborangetop, fill=laborangebot!20] (9.5, -2.3) circle (0.9);
        \coordinate (L1) at (9.1, -1.9); \coordinate (L2) at (9.1, -2.3); \coordinate (L3) at (9.1, -2.7);
        \coordinate (R1) at (9.9, -2.1); \coordinate (R2) at (9.9, -2.5);
        \draw[very thick, laborangetop] (L1) -- (R2); \draw[very thick, laborangetop] (L2) -- (R1);
        \draw[very thick, laborangetop] (L2) -- (R2); \draw[very thick, laborangetop] (L3) -- (R1);
        \fill[laborangetop] (L1) circle (0.07); \fill[laborangetop] (L2) circle (0.07);
        \fill[laborangetop] (L3) circle (0.07); \fill[laborangetop] (R1) circle (0.07);
        \fill[laborangetop] (R2) circle (0.07);
        \draw[arr] (10.6, -2.3) -- (11.8, -2.3);
        \drawmat{12}{-3.35}{1}{0.6}{2.1}{laborangetop}{laborangebot}{black}
        \node at (12.3, -0.7) {\small \textbf{Stability component}};
        \node at (12.3, -3.8) {\Large $\mathbb{R}^{n}$};

        \draw[arrG] (12.8, 2.3) -- (15.3, 0.4);
        \draw[arrO] (12.8, -2.3) -- (15.3, -0.4);
        \node[cakepurpletop] at (14.0, 0) {\large \textbf{Fusion}};
        \drawmat{15.5}{-1.05}{1}{0.6}{2.1}{cakepurpletop}{cakepurplebot}{black}
        \node at (15.8, 1.6) {\small \textbf{Confidence score}};
        \node at (15.8, -1.6) {\Large $\mathbb{R}^{n}$};
    \end{tikzpicture}
    \captionsetup{hypcap=false}
    
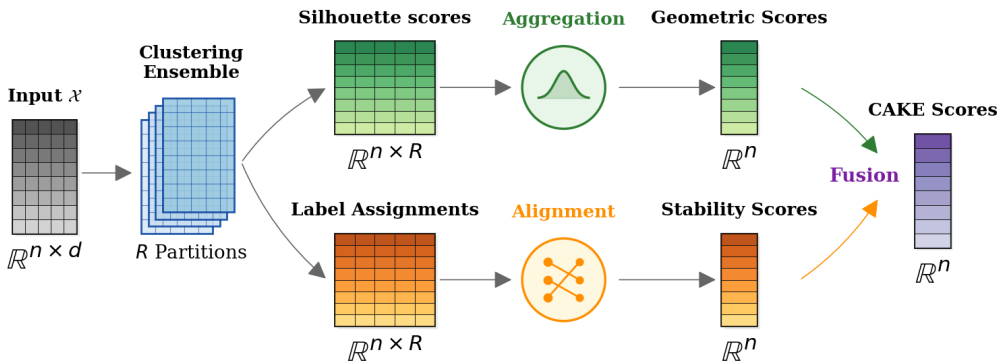
\captionof{figure}{CAKE framework overview. Across an ensemble of $R$ clustering runs (partitions), CAKE aggregates Silhouette statistics into a geometric component and (aligned) label assignments into a stability component, then fuses the two into a confidence score for each data point.}
    \label{fig:pipeline}
\end{center}

\section{Related Work}\label{sec:relwork}
\noindent Assessing the quality of clustering results has been the focus of extensive prior work. Classical validation metrics such as the Silhouette score~\parencite{dudek2020silhouette, pavlopoulos2024,rousseeuw1987}, Davies--Bouldin index~\parencite{davies1979cluster}, and Calinski--Harabasz criterion~\parencite{calinski1974dendrite} provide global or cluster-level assessments of compactness and separation, but they are not designed to quantify the reliability of individual assignments or to isolate unstable or ambiguous samples within a partition~\parencite{arbelaitz2013extensive}.

\medskip

\noindent 
Clustering ensembles aim to improve robustness by aggregating multiple partitions, using techniques such as co-association matrices or consensus functions~\parencite{fred2005, ink2022, strehl2002} and more recent ensemble-selection approaches~\parencite{golalipour2021ensemble_selection}. While effective at stabilizing global structure, these methods typically do not provide per-point confidence scores~\parencite{boongoen2018survey, zhang2022}. Their focus lies in producing a single consensus clustering, rather than quantifying how reliably each point is assigned across runs~\parencite{topchy2005clustering}. As a result, the consensus output can obscure localized disagreement in regions near cluster boundaries or in heterogeneous areas of the data, making it difficult to identify unstable points. When point-level information is available (e.g., via co-association counts), it is usually an intermediate artifact used to form the consensus rather than an explicit confidence score for diagnosing assignments. Moreover, ensemble aggregation is typically driven by partition agreement and does not incorporate pointwise geometric evidence of assignment quality into the confidence signal.

\medskip

\noindent 
Efforts to incorporate uncertainty into clustering include fuzzy clustering~\parencite{bezdek1981, bezdek1984fcm}, probabilistic mixture models, and bootstrap-based stability analysis~\parencite{benhur2002stability, hennig2007,lange2004stability, liu2022}. In related ensemble-stability work, per-point uncertainty is often summarized by the dispersion of aligned label ``votes'' across runs (e.g., entropy)~\parencite{ayad2008cumulative}. Recent deep clustering work also studies calibrated cluster confidence, for example via deep clustering networks explicitly trained for confidence calibration~\parencite{jia2025calibrated_deep_clustering}.
These approaches provide soft assignments or probability estimates, but may be less interpretable when applied to hard clustering results; in particular, agreement-focused resampling or label-alignment heuristics quantify per-point consistency but omit local geometric fit, leaving no unified stability--geometry measure. In addition, model-based techniques such as Gaussian Mixture Models~\parencite{reyn2009} require distributional assumptions, while bootstrapping methods can be computationally intensive and sensitive to sampling variability.

\medskip

\noindent 
Another relevant area involves the challenge of label alignment when comparing clustering results, often addressed using the Hungarian algorithm for optimal permutation matching~\parencite{kuhn1955hungarian,meila2007}.\\ While this alignment step is standard in ensemble clustering, it has been less commonly used to systematically quantify per-point assignment stability across an ensemble.

\medskip

\noindent 
Extending these approaches, CAKE combines optimal label alignment with local geometric analysis to provide a principled, interpretable confidence score for each data point, bridging ensemble-based robustness with pointwise reliability and grounding the ensemble consensus in geometric evidence.
CAKE draws inspiration from supervised ensemble techniques, where disagreement between classifiers informs uncertainty~\parencite{beluch2018power, gal2016dropout, lakshminarayanan2017simple}. Such methods leverage variation across independently trained models or stochastic passes (e.g., dropout) to estimate predictive confidence. In contrast, CAKE operates in an unsupervised regime, using clustering ensemble diversity and within-cluster geometric coherence to infer confidence scores without labels.

\section{Methodology}\label{sec:methodology}
\subsection{Setting \& Notation}
\label{subsec:notation}
\noindent Assume a dataset $\Xp = \left\{x_i \right\}_{i=1}^n \subset \mathbb{R}^d$ consisting of $n$ data points. Let $\Cp = \left\{C^{(1)}, C^{(2)}, \dots, C^{(R)}  \right\}$ denote a collection of $R$ clustering results obtained by repeatedly applying the same clustering algorithm to $\Xp$, each time using the same number of clusters $k$ but with a different random seed. 

\medskip

\noindent 
This collection forms a clustering ensemble by capturing multiple partitions of the dataset under stochastic variation. Although all runs use the same algorithm and parameters, random initialization introduces meaningful differences in the resulting partitions, reflecting uncertainty in how individual points are grouped. More generally, the ensemble could be constructed using resampling strategies (e.g., clustering on bootstrapped subsets of $\Xp$), or by aggregating partitions from different clustering algorithms, offering a broader perspective on assignment variability (Fig.~\ref{fig:app:moons_cake}). While the proposed framework is compatible with such (heterogeneous) ensemble construction strategies (Fig.~\ref{fig:ensembles-real-synth}), including resampling and multi-algorithm clustering ensembles, here, the focus is on (homogeneous) ensembles formed by repeated applications of a single clustering method, allowing assignment-level confidence to be quantified without interference from other sources of variation.

\medskip

\noindent 
To use these partitions quantitatively, let $\Lp = \left\{L^{(1)}, L^{(2)}, \dots, L^{(R)} \right\}$ denote the corresponding set of assignments. Each labeling $L^{(r)} \in \Lp$ assigns a cluster label to every data point $x_i \in \Xp$, such that
$L^{(r)} = \left\{L^{(r)}_1, L^{(r)}_2, \dots, L^{(r)}_n \right\}$, where $L^{(r)}_i \in \left\{1,2,\dots, k \right\}$: the label of $x_i$ in clustering run $r$.

\subsection{Ensemble Silhouette Statistics}
\label{subsec:silhouette}
\noindent For each point $x_i \in \Xp$, let $s_i^{(r)}$ denote its Silhouette score under clustering run $r \in \left\{1,2,\dots,R \right\}$, 
\begin{equation}\label{eq:sil}
s_i^{(r)} = \frac{b_i^{(r)} - a_i^{(r)}}{\max\left\{a_i^{(r)}, \ b_i^{(r)}\right\}} \in [-1,1],
\end{equation}
which quantifies the quality of its assignment by comparing its average intra-cluster distance $a_i^{(r)}$ to its minimum average inter-cluster distance $b_i^{(r)}$ under clustering run $r$.

\medskip

\noindent 
Computing all $s_i^{(r)}$ is $O(n^2 d)$ per clustering run. When the base method is centroidal (e.g., \emph{k}-means, \emph{k}-means++) and $n$ is large, a centroid-based Silhouette approximation~\parencite{wangsilapprox} can be computed instead by replacing
$\tilde a_i^{(r)} = \lVert x_i - \mu^{(r)}_{C_i}\rVert \ \text{ and } \ 
\tilde b_i^{(r)} = \min_{C \neq C_i} \lVert x_i - \mu^{(r)}_{C} \rVert,$
where $\mu^{(r)}_{C}$ is the centroid of cluster $C$ in run $r$, and $C_i$ denotes the cluster index of $x_i$ in run $r$ (i.e., $C_i=L^{(r)}_i$).
This reduces the per–run cost to $O(nkd)$ while tracking the exact score closely in centroidal settings.

\medskip

\noindent\textbf{Aggregating over the ensemble.} The mean Silhouette score and standard deviation (std) are computed for each point $x_i$ across all $R$ clustering runs in the ensemble: 
\begin{equation}\label{eq:silstas}
    \mu_i = \frac{1}{R} \sum_{r=1}^{R} s_i^{(r)}, \quad\sigma_i = \sqrt{ \frac{1}{R} \sum_{r=1}^{R} \left( s_i^{(r)} - \mu_i \right)^2}
\end{equation}
These two quantities are used to define a per‐point measure of \textit{Silhouette‐based reliability}. A high mean $\mu_i$ indicates that $x_i$ tends to be well-placed geometrically across clustering runs, while a low std $\sigma_i$ implies that this quality is consistent. Subtracting $\sigma_i$ from $\mu_i$ penalizes points with high variability in Silhouette values across runs, emphasizing both quality and stability of geometric fit. The resulting score is thresholded at zero to ensure non-negativity:\footnote{To preserve the signal from consistently negative means (e.g., $\mu_i < 0$ with small $\sigma_i$; a relatively uncommon regime in practice), $\mu_i-\sigma_i$ can be mapped to $[0,1]$ via $\left[(\mu_i - \sigma_i)+1\right]/2$}
\begin{equation}\label{eq:silhouette}
    \tilde S_i =  \left( \mu_i - \sigma_i \right)_+ = \max\left\{0, \ (\mu_i - \sigma_i) \right\} \in [0,1].
\end{equation}
$\tilde S_i$  rewards points that not only have high average Silhouette (good cluster fit) but also low variability across runs (stable geometric placement). However, a high $\tilde S_i$ does not guarantee that a point is consistently assigned to the same cluster across runs. It only indicates that the point maintains a consistently strong fit within its assigned cluster. To capture actual assignment stability, a complementary measure is introduced based on cluster agreement across the ensemble.

\subsection{Ensemble Assignment Stability}
\label{subsec:stability}
\medskip
\noindent\textbf{Label alignment.}
Given two clustering labelings $L^{(r_1)}, L^{(r_2)} \in \Lp$ from clustering runs $r_1$, $r_2$, the labels may not be aligned: 
that is, cluster label $j$ in $L^{(r_1)}$ and in $L^{(r_2)}$ may correspond to a different group of points.
To align $L^{(r_2)}$ onto $L^{(r_1)}$, a contingency matrix $M\in \mathbb{N}^{k\times k}$ is defined, where each entry $M_{i,j}$ counts the number of points assigned to cluster $i$ in $L^{(r_1)}$ and cluster $j$ in $L^{(r_2)}$: 
\begin{equation}
M_{i,j} = \sum_{p=1}^{n}\mathds{1}\left\{ \left(L_p^{(r_1)}=i\right) \wedge \left(L_p^{(r_2)}=j\right)\right\}
\end{equation}
where $\mathds{1}(\cdot)$ is the indicator function. Finding the optimal alignment corresponds to finding a permutation $\pi^*$, with $\pi^*:\left\{ 1, \dots, k\right\} \rightarrow \left\{ 1, \dots, k\right\}$, that maximizes agreement: 
\begin{equation}\label{eq:P}
    \pi^*=\arg\max_{\pi} \sum_{i=1}^{k}M_{i, \pi(i)},
\end{equation}
where $\pi(i)$ denotes the label in $L^{(r_2)}$ that is matched to label $i$ in $L^{(r_1)}$. This optimization is equivalent to solving a \textit{linear sum assignment problem}, which can be solved using the Hungarian algorithm in $O(k^3)$ time. 
After aligning $L^{(r_2)}$ onto $L^{(r_1)}$, the aligned labeling is denoted by $L^{(r_2  \rightarrow  r_1)}$ and the pointwise agreement between two clustering runs $(r_1,\ r_2)$ for a data point $x_i \in \Xp$ is defined as: 
\begin{equation}\label{eq:bernoullikernel}
A_i^{(r_1, r_2)} = \mathds{1}\left\{ L_i^{(r_1)}= L_i^{(r_2  \rightarrow  r_1)}\right\} \in \left\{0,1 \right\}.
\end{equation}
\textbf{Instance pairwise stability.} 
Given $R$ clustering runs, there are $\binom{R}{2} = \frac{R(R-1)}{2}$ distinct unordered pairs of partitions.
The pairwise stability score $c_i$ is defined for every point $x_i \in \Xp$ as the fraction of all unordered run-pairs in which $x_i$ is assigned to the same cluster (after alignment):
\begin{equation}\label{eq:stability}
c_i = \frac{2}{R(R-1)} \sum_{r_1 < r_2} A_i^{(r_1, r_2)} \in [0,1].
\end{equation}
\noindent A common alternative is to align all runs to a single reference partition (e.g., a medoid run) and compute vote-based uncertainty (such as entropy). This can be attractive in time-sensitive applications because it reduces alignment cost from quadratic to linear in $R$. However, reference-based scores can depend on the chosen run and may inherit its instability, effectively privileging one partition as canonical. In contrast, Eq.~\ref{eq:stability} averages agreement over \emph{all} unordered run-pairs, yielding a symmetric stability score that is less sensitive to any single run. Practically, $c_i$ can still be computed efficiently by streaming over run-pairs and accumulating agreements without storing all pairwise indicators.

\medskip

\noindent By definition, $c_i$ rewards points that are consistently assigned to the same cluster across the ensemble. Points with high $c_i$ exhibit minimal assignment variation and can be viewed as stable members of the underlying clustering structure.
Moreover, since $A_i^{(r_1,r_2)}$ is a Bernoulli indicator taking values in $\{0,1\}$, the quantity $c_i$ is an order-2 U-statistic with a binary (Bernoulli) kernel~\parencite{lee1990u}. As such, $c_i$ is an \textit{unbiased estimator of the true pairwise assignment stability} $\mathbb{E}[A_i^{(r_1, r_2)}]$ (i.e., the expected agreement over pairs of runs, which reflects how reproducibly $x_i$ is clustered across the ensemble).

\subsection{CAKE Score}\label{subsec:cake}
\noindent From the clustering ensemble, two complementary per-instance diagnostics are derived: the assignment-stability estimate $c_i$ (Eq.~\ref{eq:stability}) and the Silhouette-based reliability score $\tilde S_i$ (Eq.~\ref{eq:silhouette}).
CAKE combines these signals into a single confidence value in $[0,1]$ (outlined in Fig.~\ref{fig:pipeline} \& Algorithm~\ref{alg}). It rewards observations that are repeatedly assigned to the same cluster across runs and are strongly supported by their local geometry, while penalizing points with unstable assignments or poor local separation from other clusters. 
The two formulations are:
\begin{subequations}\label{eq:cake}
\noindent\begin{minipage}{.48\linewidth}
\begin{equation}\label{eq:cake:a}
\mathrm{CAKE}^{\text{(PR)}}_i = c_i\,\tilde S_i
\end{equation}
\end{minipage}\hfill
\begin{minipage}{.48\linewidth}
\begin{equation}\label{eq:cake:b}
\mathrm{CAKE}^{\text{(HM)}}_i = \frac{2\,c_i\,\tilde S_i}{c_i+\tilde S_i}
\end{equation}
\end{minipage}
\end{subequations}
The product (Eq.~\ref{eq:cake:a}) attains high values only when both components are large and rapidly downweights points that are weak in either stability or geometric fit. The harmonic mean (Eq.~\ref{eq:cake:b}) provides a smoother trade-off, remaining sensitive to low values while being less dominated by a single moderate component, which can make the combined score easier to interpret. This mirrors the role of the F1 score in classification, which balances precision and recall via a harmonic mean~\parencite{powers2011evaluation}, and similarly treats stability and geometric fit as complementary requirements for high confidence (see the score distribution and percentiles on synthetic data in Fig.~\ref{fig:cake_side_by_side}).

\begin{algorithm}[H]
\caption{CAKE scores computation}
\label{alg}
\begin{algorithmic}
\Require Dataset $\Xp$, Clustering algorithm(s) $\Fp$, Ensemble size $R > 1$
\Ensure $\mathrm{CAKE}$ scores $\forall x_i \in \Xp$

\State Initialize label assignments matrix $L \in \mathbb{N}^{n \times R}$  \Comment{\textcolor{darkgray}{Assigned labels per point/clustering run}}
\State Initialize Silhouette score matrix $S \in \mathbb{R}^{n \times R}$ \Comment{\textcolor{darkgray}{Silhouette per point/clustering run}}
\State Initialize Agreement Vector $A \in \mathbb{N}^{n}$ \Comment{\textcolor{darkgray}{Pairwise (clustering) run agreements}}
\For{$r = 1$ to $R$}
    \State Run clustering algorithm $\Fp$ on $\Xp$ with random seed $r$ \Comment{\textcolor{darkgray}{Ensemble construction (\S\ref{subsec:notation})}}
    \State Let $L[:, r] \gets$ predicted labels
    \State Compute Silhouette scores $S[:, r]$ (Eq.~\ref{eq:sil}) 
\EndFor
\State $\mu \gets \frac{1}{R} \sum_{r=1}^R S[:, r], \ \sigma \gets \sqrt{ \frac{1}{R} \sum_{r=1}^R (S[:, r] - \mu)^2 }$ (Eq.~\ref{eq:silstas}) \Comment{\textcolor{darkgray}{ Ensemble Silhouette statistics} (\S\ref{subsec:silhouette})}
\State  Geometric component $\tilde{S} \gets (\mu - \sigma)_+$ (Eq.~\ref{eq:silhouette})
\For{each clustering pair $(r_1, r_2)$ with $r_1 < r_2$} \Comment{\textcolor{darkgray}{Ensemble assignment stability} (\S\ref{subsec:stability})}
    \State Compute optimal label mapping $\pi^{(r_2 \rightarrow r_1)}$ (Eq.~\ref{eq:P})
    \State $A \gets A + \mathds{1}\left\{L[:, r_1] =\pi^{(r_2 \rightarrow r_1)}(L[:, r_2])\right\}$ (Eq.~\ref{eq:bernoullikernel})
\EndFor
\State Stability component $c \gets 2A[R(R-1)]^{-1}$ (Eq.~\ref{eq:stability}) 
\State $ \mathrm{CAKE} \gets \
\mathrm{CAKE}^{\text{(PR)}} \ \text{ or } \ \mathrm{CAKE}^{\text{(HM)}}$ (Eq.~\ref{eq:cake}) \Comment{\textcolor{darkgray}{CAKE scores computation} (\S\ref{subsec:cake})}
\State \Return $\mathrm{CAKE}$ 
\end{algorithmic}
\end{algorithm}

\begin{center}
{\captionsetup{type=figure} 

  \begin{subfigure}[b]{0.5\linewidth}
    \centering
    \includegraphics[width=0.8\linewidth]{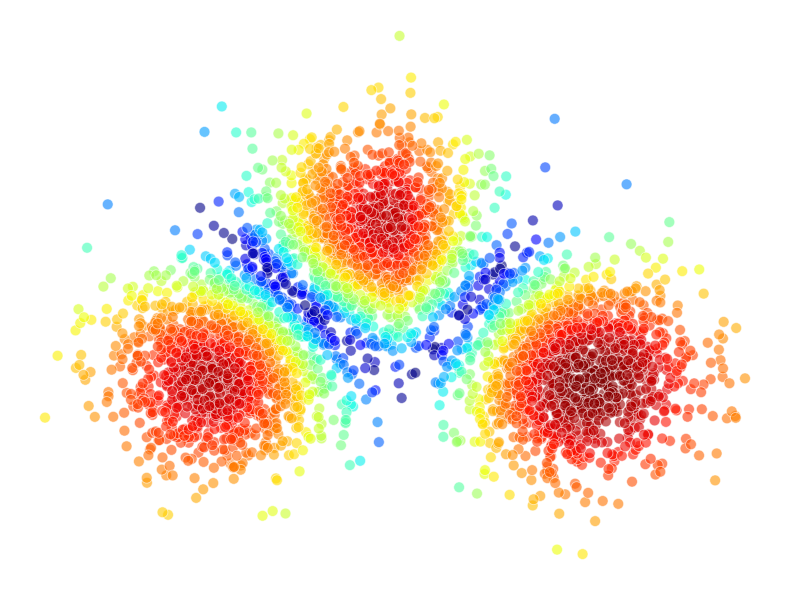}
    \caption{\textbf{\textcolor{blue}{Low}} $\rightarrow$ \textbf{\textcolor{red}{high}} $\mathrm{CAKE}$ scores.}
    \label{fig:cakedist}
  \end{subfigure}\hfill
  \begin{subfigure}[b]{0.5\linewidth}
    \centering
    \includegraphics[width=0.8\linewidth]{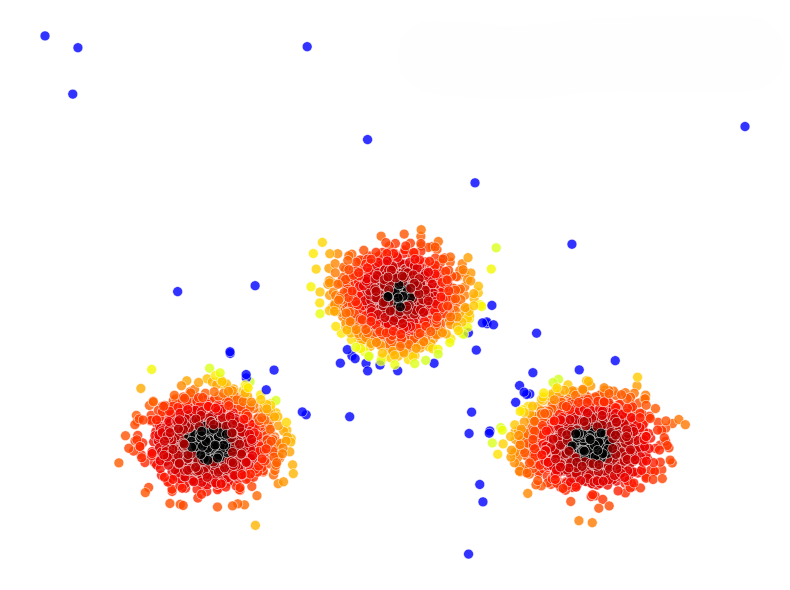}
    \caption{\textbf{\textcolor{blue}{Bottom 1\%}} $\rightarrow$ \textbf{top 90\%} $\mathrm{CAKE}$ scores}
    \label{fig:cake_percentile_synthetic}
  \end{subfigure}

  \caption{$\mathrm{CAKE}^{\text{(PR)}}$ (Eq.~\ref{eq:cake:a}) scores distribution (left) and percentiles (right) on synthetic data.}
  \label{fig:cake_side_by_side}
}
\end{center}

\noindent \textbf{Complexity.} With the centroid proxy, computing Silhouette scores over $R$ runs costs $O(Rnkd)$, while exact computations cost $O(R n^2 d )$. Label alignment over all run-pairs builds a $k\times k$ contingency and solves a Hungarian assignment per pair, for $O\!\big(\binom{R}{2}(n + k^3)\big)$ time; thus CAKE runs in $O(Rnk d + \binom{R}{2}(n + k^3))$ overall, or $O(Rn^2 d + \binom{R}{2}(n + k^3))$ using exact Silhouette scores, and uses $O(nR)$ memory (see runtime vs. $R, n$ on synthetic data in Fig.~\ref{fig:runtime_cake} and Tables~\ref{tab:runtimer}--\ref{tab:runtimen}). 

\medskip

\noindent \textbf{Non-convex data.}
The silhouette component (Eqs.~\ref{eq:sil}--\ref{eq:silhouette}) is typically computed with Euclidean distances and thus favors convex, isotropic clusters.
For non-convex structure, Euclidean silhouette can be misleading. CAKE accommodates such settings by replacing Euclidean distances in the silhouette computation (\S\ref{subsec:silhouette}) with distances induced by a \textit{positive semidefinite (PSD) kernel}.
Specifically, a \textit{kernelized Silhouette} is used, replacing Euclidean distances by RKHS distances to cluster means under a PSD kernel $\kappa$ (see Fig.~\ref{fig:app:moons_cake} for an application to non-convex data):
\begin{equation}\label{eq:kerneldist}
d_\kappa^2(x_i, C)=\kappa(x_i,x_i)-\frac{2}{|C|}\sum_{x_j\in C}\kappa(x_i,x_j)
              +\frac{1}{|C|^2}\sum_{x_p\in C}\sum_{x_q\in C}\kappa(x_p,x_q).
\end{equation}
Here, $\kappa(\cdot,\cdot)$ is a \emph{positive semidefinite (PSD) kernel}, i.e., a symmetric similarity function such that the Gram matrix
$[\kappa(x_i,x_j)]_{i,j}$ is PSD. Intuitively, $\kappa(x,y)$ measures how similar $x$ and $y$ are, and it corresponds to an inner product
in some feature space: $\kappa(x,y)=\langle \phi(x),\phi(y)\rangle$.
\vspace{-2mm}
\begin{center}
  \includegraphics[width=0.5\linewidth]{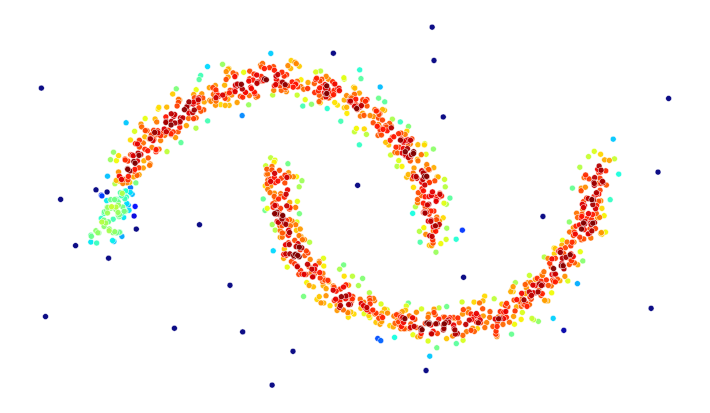}
  \captionsetup{hypcap=false}
\captionof{figure}{
$\mathrm{CAKE}^{\text{(PR)}}$ distribution on the two-moons dataset ($n{=}1200$) with $3\%$ uniform outliers.
An ensemble of $R{=}25$ partitions is constructed using \textbf{spectral clustering} on a $k$-NN graph, jittering $n_{\text{neighbors}}\in[10,15]$ per run to induce diversity.
For the geometric component, the kernelized Silhouette (\S\ref{subsec:silhouette}; Eq.~\ref{eq:kerneldist}) is used with a self-tuning RBF kernel:
$\kappa(x_i,x_j)=\exp\!\big[-\|x_i-x_j\|^2/(\sigma_i\sigma_j)\big]$.
The Gram matrix $K$ is built once and reused across runs, setting each $\sigma_i$ to the distance from $x_i$ to its $k_{\text{nn}}{=}7$th nearest neighbor.
Points are colored by their CAKE scores ($\uparrow$ \textbf{\textcolor{red}{red}}, $\downarrow$ \textbf{\textcolor{blue}{blue}}).
}
  \label{fig:app:moons_cake}
\end{center}
\vspace{-7mm}
\section{Theoretical Analysis}
\label{sec:theory}
\noindent Two non-asymptotic results are provided for the finite-ensemble stability scores $c_i$ (Eq.~\ref{eq:stability}); full derivations and analogous discussion for the geometric component $\tilde S_i$ (Eq.~\ref{eq:silhouette}) are given in Appendix~\ref{app:analysis}.
\vspace{-2mm}
\subsection{Ranking-Error Bound for Finite Ensembles}
\label{subsec:rankingerror}
\noindent Let $R\ge 2 $ be the number of independent clustering runs performed on $\Xp$. For a point $x_i \in \Xp$, the true pairwise stability is $\theta_i = \mathbb{E}\left[A_i^{(r_1, r_2)} \right]$, where $A_i^{(r_1, r_2)}$ (Eq.~\ref{eq:bernoullikernel}) is the Bernoulli kernel used in the order-2 U-statistic estimator $c_i$ (Eq.~\ref{eq:stability}). Intuitively, the empirical scores should rank points in the same order as their true stability $\theta_i$ and any reversed ordering should be random noise that fades when more runs are added.\footnote{By standard concentration for bounded U-statistics, $c_i$ converges to $\theta_i$ exponentially in $R$ (Appendix~\ref{app:bound1}).} The following bound shows how the misordering probability shrinks with $R$ (derivations in Appendix~\ref{app:bound1}).
For any two points $x_i, x_j\in \Xp$ and margin $\gamma>0$, it holds:
\begin{equation}
\mathbb{E}\left[A_i^{(r_1, r_2)} \right] \ge \mathbb{E}\left[A_j^{(r_1, r_2)} \right] + \gamma 
\Rightarrow \ \Pr\left[ c_i < c_j \right] \le 2 \exp\left\{ -R\frac{\gamma^2}{8}\right\}.
\end{equation}

\subsection{False-Positive Bound for Uniform-Noise Points}
\label{subsec:falsepositive}
\noindent Consider a uniform-noise observation, i.e., a point whose cluster label in each run is drawn independently and uniformly from $\left\{1,\dots,k\right\}$. Then $\theta_i=\frac{1}{k}$ and deviations of $c_i$ from $\theta_i$ decay exponentially in $R$~\parencite{hoeffding1963probability}. Thus, the probability that a noise point falsely attains a high score decreases rapidly with ensemble size (Appendix~\ref{app:bound2}):
\begin{equation}
\Pr[c_i>\tau \ | \ x_i \text{ uniform-noise}] \le \exp\left\{-\frac{R}{2}\left( \tau - \frac{1}{k} \right)^2 \right\} \quad \forall \text{  threshold } \tau>\frac{1}{k}.
\end{equation}
More generally, for any label distribution with expected agreement $\theta_i = \mathbb{E}[A_i^{(r_1, r_2)}]$: $\Pr[c_i>\tau] \le \exp\left\{ -\frac{R}{2}(\tau - \theta_i)^2 \right\}, $ ($\forall \tau>\theta_i$) so the probability still decays exponentially in $R$ (detailed analysis in Appendix~\ref{app:bound2}).
\vspace{-2mm}
\section{Empirical Validation}\label{empiricalvalidation}
\subsection{Synthetic \& Real-World Datasets}
\noindent CAKE is evaluated on synthetic and real-world datasets.
The \textbf{synthetic datasets} are described next (visualizations in Fig.~\ref{fig:app:synth_overview}). \textbf{S1} has 4,000 points sampled from three Gaussian clusters with equal size and std of 2.0; \textbf{S2} has 3,000 points from three clusters with std of 2.0, 2.5, and 1.5. \textbf{S3} has 4,500 points, where 3,000 belong to three Gaussian clusters with unit std, and 1,500 are uniformly distributed noise points. \textbf{S4} has 3,000 points arranged in four wide Gaussian clusters, located at the corners of a square. \textbf{S5} has 4,000 points from three clusters with strong density contrast: one tight cluster with std of 0.2, one wide with std of 3.0, and one intermediate with unit std. \textbf{S6} has 4,000 points from three clusters with std of 0.4, 2.5, and 0.4, where a sparse central cluster overlaps denser ones on each side. \textbf{S7} has 4,000 points, distributed across three clusters of size 500, 1,000, and 2,500, with std of 0.3, 1.5, and 2.5 respectively. 
\vspace{-2mm}
\begin{center}
  \includegraphics[width=0.7\linewidth]{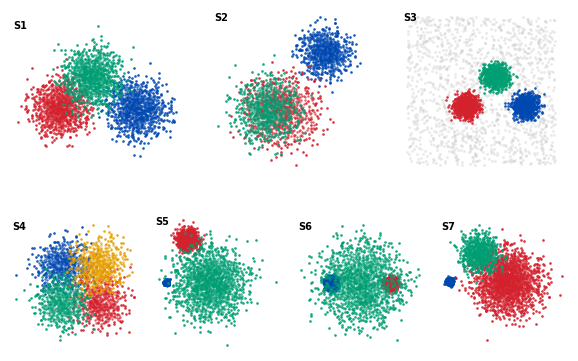}
  \captionsetup{hypcap=false}
  \captionof{figure}{Synthetic datasets (\textbf{S1}-\textbf{S7}), colored by ground-truth cluster labels; noise in \textcolor{gray}{light gray}.}
  \label{fig:app:synth_overview}
\end{center}

\noindent The \textbf{real-world datasets}, which are publicly available via OpenML~\parencite{OpenML2013}, \textsc{scikit-learn}~\parencite{scikit-learn}, or TensorFlow~\parencite{tensorflow-datasets}, span a wide range of domains and structures (sizes, dimensionality, and preprocessing are summarized in Table~\ref{tab:realdata}):  \textbf{Iris} (\textsc{Ir}), \textbf{Breast Cancer} (\textsc{Bc}), \textbf{Pendigits} (\textsc{Pd}), \textbf{Letter} (\textsc{Lt}), \textbf{Digits} (\textsc{Dg}), \textbf{Fashion MNIST} (\textsc{Fm}), \textbf{Satimage} (\textsc{Sa}), and \textbf{20~Newsgroups} (\textsc{Ng}).

\begin{center}
\footnotesize
\setlength{\tabcolsep}{6pt}
\begin{tabularx}{\textwidth}{l l r r r X}
\toprule
\textbf{Dataset (abbrev.)} & \textbf{Modality} & $\boldsymbol{n}$ & $\boldsymbol{d}$ & $\boldsymbol{k}$ & \textbf{Preprocessing / Representation} \\
\midrule
Iris (\textsc{Ir}) & Tabular         & 150      & 4   & 3  & None \\
Breast Cancer (\textsc{Bc}) & Tabular         & 569      & 10  & 2  & Standardize; PCA \\
Pendigits (\textsc{Pd}) & Tabular         & 10{,}992 & 16  & 10 & None \\
Letter (\textsc{Lt}) & Tabular         & 20{,}000 & 16  & 26 & Standardize \\
Digits (\textsc{Dg}) & Image           & 1{,}797  & 64  & 10 & Flatten 8$\times$8 grayscale images \\
Fashion MNIST (\textsc{Fm}) & Image           & 60{,}000 & 784 & 10 & Flatten 28$\times$28 grayscale images \\
Satimage (\textsc{Sa}) & Remote sensing  & 6{,}435  & 30  & 6  & PCA \\
20~Newsgroups (\textsc{Ng}) & Text            & 18{,}846 & 100 & 20 & all-MiniLM-L6-v2 embeddings; PCA\\
\bottomrule
\end{tabularx}
\captionsetup{hypcap=false}
\captionof{table}{Real datasets. $n$: number of samples; $d$: post-preprocessing dimensionality; $k$: number of classes (clusters).}
\label{tab:realdata}
\end{center}

\subsection{Evaluation Setup}\label{subse:evalset}
\medskip
\noindent \textbf{Instance removal.}
Instances are ranked according to their CAKE scores (\S\ref{subsec:cake}), and the highest-confidence subset is then selected in order to assess the effect of retaining the most reliable points on downstream clustering quality.
For each dataset, CAKE scores are computed using an ensemble of 20 independent \emph{k}-means runs at the ground-truth number of clusters $k$ (random initialization for synthetic datasets; \emph{k}-means++ for real datasets).
\emph{k}-means is used because it is widely used, easy to interpret, and known to be sensitive to initialization, making it a natural candidate for assessing assignment-level confidence; CAKE itself is model-agnostic and applies to any hard-assignment clustering ensemble (see Figs.~\ref{fig:app:moons_cake}, \ref{fig:ensembles-real-synth} and Appendix~\ref{app:sec:divens}).
In this setting, CAKE is treated primarily as a pointwise confidence score that induces a ranking over instances, while filtering is considered an application of that ranking under a specified retention fraction. Accordingly, the experiments are designed to assess whether the induced ranking distinguishes more reliable from less reliable instances under a fixed clustering resolution. The subsequent subset-selection analysis then provides a downstream test of whether high-confidence instances are also those whose assignments are more consistently supported by the local cluster structure. 

\medskip

\noindent For each dataset, six equal-size subsets are formed, each retaining 70\% of the points, $m=\lfloor 0.7n \rfloor$, according to different criteria: \textbf{Random} (sample $m$ points uniformly), \textbf{Consensus} (top $m$ by across-run label agreement after aligning all runs to a reference--medoid partition, i.e., the most representative run), $\tilde{\mathbf{S}}$ (top $m$ by the Silhouette component; Eq.~\ref{eq:silhouette}), $\mathbf{C}$ (top $m$ by the stability component; Eq.~\ref{eq:stability}), $\mathbf{CAKE}^{\textbf{(PR)}}$ (top $m$ by the product; Eq.~\ref{eq:cake:a}), and $\mathbf{CAKE}^{\textbf{(HM)}}$ (top $m$ by the harmonic mean; Eq.~\ref{eq:cake:b}). While this specific threshold is not necessarily optimal, it offers a reasonable coverage--confidence trade-off (see Fig.~\ref{fig:acc_cov_cake}; Appendix Fig.~\ref{app:fig:evaluation}). An adaptive thresholding variant, in which the retained fraction is selected automatically from the data, is reported separately in Appendix~\ref{app:sec:adafilter}.

\medskip

\noindent 
Clustering performance is compared on the full datasets and on their filtered subsets using new \emph{k}\nobreakdash-means runs (separate from the CAKE ensemble) at the ground-truth $k$. Performance is measured via Adjusted Rand Index (ARI), Adjusted Mutual Information (AMI), and clustering accuracy (ACC), after Hungarian alignment to ground-truth labels. 
To assess variability, the evaluation is repeated with multiple independent runs and means are reported with (Student’s $t$) 95\% confidence intervals (Table~\ref{tab:results}; see Appendix~\ref{app:sec:valid} Tables~\ref{tab:results_minibatchkmeans}--\ref{tab:results_gmm} for MiniBatchKMeans and GMM evaluation). 
Additional results for Silhouette, Normalized Mutual Information (NMI), and the correlation between CAKE score percentiles and clustering accuracy are reported in Appendix~\ref{app:sec:valid} (Table~\ref{app_extrares}).

\medskip

\noindent  
\textbf{Convergence.} The stabilization of CAKE scores with increasing ensemble size $R$ is studied. For each $R\!\in\!\{5,\dots,70\}$, $B{=}10$ sub-ensembles of size $R$ are drawn, CAKE scores are recomputed using the centroid-based Silhouette proxy (\S\ref{subsec:silhouette}), and the per-point standard deviation across the $B$ estimates is reported (Fig.~\ref{fig:convergence}). Lower variability suggests that CAKE scores are stabilizing (converging) as the clustering ensemble grows.

\begin{center}
\includegraphics[width=\linewidth]{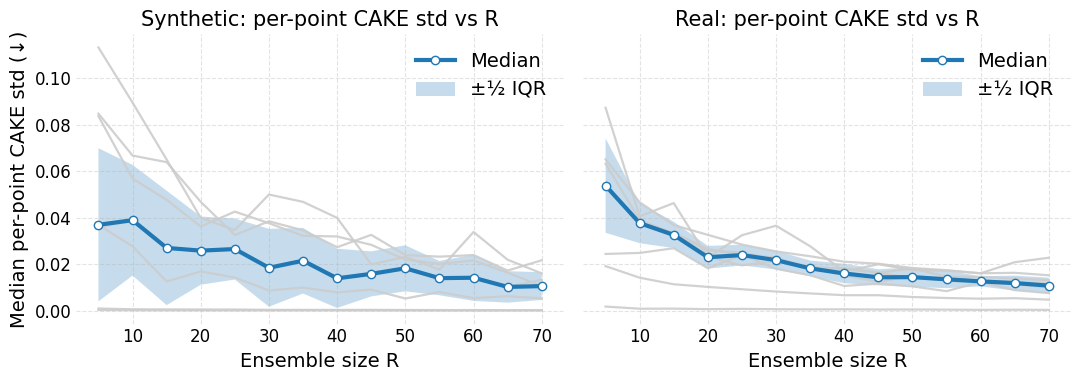}
  \captionsetup{hypcap=false}
  \captionof{figure}{Convergence vs $R$. Median per-point $\mathrm{CAKE}^{\text{(HM)}}$ scores (Eq.~\ref{eq:cake:b}) std vs.\ ensemble size $R$ for synthetic (left) and real (right) datasets. \textbf{Lines:} \textcolor{gray}{light gray}: individual datasets; \textcolor{MidnightBlue}{blue}: across-dataset median. \textbf{Band:} $\pm \tfrac{1}{2}$ IQR. Variability drops and stabilizes by $R{\approx}30$–$40$ across datasets.}
\label{fig:convergence}
\end{center}

\onecolumn
\footnotesize

\begin{longtable}{llccc}

\toprule
\textbf{Dataset} & \textbf{Subset} & \textbf{avg ARI} & \textbf{avg AMI} & \textbf{avg ACC} \\
\midrule

\multirow{7}{*}{S1} & Full & 0.879 [0.8785, 0.8791] & 0.833 [0.8331, 0.8338] & 0.958 [0.9578, 0.9580] 
\\
& Random & 0.879 [0.8790, 0.8790] & 0.834 [0.8339, 0.8339] & 0.958 [0.9579, 0.9579] \\
& Consensus & 0.884 [0.8842, 0.8842] & 0.839 [0.8394, 0.8394] & 0.960 [0.9596, 0.9596]
\\
& $C$ component & 0.884 [0.8842, 0.8842] & 0.839 [0.8394, 0.8394] & 0.960 [0.9596, 0.9596]
\\
& $\tilde S$ component & \textbf{0.992} [0.9921, 0.9921] & \textbf{0.983} [0.9827, 0.9827] & \textbf{0.997} [0.9971, 0.9971] 
\\
& $\mathrm{CAKE}^{\text{(PR)}}$ & \textbf{0.992} [0.9921, 0.9921] & \textbf{0.983} [0.9827, 0.9827] & \textbf{0.997} [0.9971, 0.9971]
\\
& $\mathrm{CAKE}^{\text{(HM)}}$ & \textbf{0.992} [0.9921, 0.9921] & \textbf{0.983} [0.9827, 0.9827] & \textbf{0.997} [0.9971, 0.9971]
\\

 \cmidrule(lr){2-5}

\multirow{7}{*}{S2} & Full & 0.552 [0.5225, 0.5812] & 0.606 [0.6056, 0.6068] & 0.762 [0.6993, 0.8244]
\\
& Random & 0.529 [0.4901, 0.5684] & 0.601 [0.5923, 0.6102] & 0.724 [0.6441, 0.8029]
\\
& Consensus & 0.528 [0.5070, 0.5492] & 0.628 [0.6031, 0.6530] & \textbf{0.797} [0.7838, 0.8094]
\\
& $C$ component & 0.528 [0.5070, 0.5492] & 0.628 [0.6031, 0.6530] & \textbf{0.797} [0.7838, 0.8094]
\\
& $\tilde S$ component & 0.679 [0.5731, 0.7840] & 0.698 [0.6789, 0.7162] & 0.759 [0.6359, 0.8814] 
\\
& $\mathrm{CAKE}^{\text{(PR)}}$ & \textbf{0.680} [0.5741, 0.7860] & \textbf{0.700} [0.6800, 0.7192] & 0.761 [0.6393, 0.8827]
\\
& $\mathrm{CAKE}^{\text{(HM)}}$ & \textbf{0.680} [0.5738, 0.7858] & 0.699 [0.6796, 0.7182] & 0.760 [0.6367, 0.8829]
\\

 \cmidrule(lr){2-5}

\multirow{7}{*}{S3} & Full & 0.498 [0.4941, 0.5016] & 0.614 [0.6098, 0.6178] & 0.733 [0.7268, 0.7384]
\\
& Random & 0.449 [0.3792, 0.5188] & 0.580 [0.5352, 0.6255] & 0.678 [0.6076, 0.7479]
\\
& Consensus & 0.753 [0.7090, 0.7973] & 0.757 [0.7301, 0.7836] & 0.842 [0.8013, 0.8828]
\\
& $C$ component & 0.719 [0.6957, 0.7432] & 0.721 [0.7041, 0.7379] & 0.830 [0.8128, 0.8473]
\\
& $\tilde S$ component & \textbf{0.795} [0.7667, 0.8241] & \textbf{0.781} [0.7591, 0.8022] & \textbf{0.862} [0.8348, 0.8886]
\\
& $\mathrm{CAKE}^{\text{(PR)}}$ & 0.734 [0.6444, 0.8232] & 0.751 [0.6982, 0.8031] & 0.784 [0.7085, 0.8591]
\\
& $\mathrm{CAKE}^{\text{(HM)}}$ & 0.784 [0.7538, 0.8137] & 0.776 [0.7597, 0.7932] & 0.837 [0.8022, 0.8724] 
\\

\cmidrule(lr){2-5}

\multirow{7}{*}{S4} & Full & 0.716 [0.7158, 0.7167] & 0.678 [0.6775, 0.6786] & 0.884 [0.8833, 0.8837] 
\\
& Random & 0.727 [0.7268, 0.7276] & 0.693 [0.6926, 0.6933] & 0.888 [0.8883, 0.8886]	
\\
& Consensus & 0.606 [0.5946, 0.6179]	 & 0.598 [0.5939, 0.6029] & 0.755 [0.7432, 0.7665]
\\
& $C$ component & 0.606 [0.5946, 0.6179]	 & 0.598 [0.5939, 0.6029] & 0.755 [0.7432, 0.7665] 
\\
& $\tilde S$ component & \textbf{0.902} [0.9022, 0.9022] & \textbf{0.867} [0.8667, 0.8667] & \textbf{0.962} [0.9624, 0.9624]
\\
& $\mathrm{CAKE}^{\text{(PR)}}$ & \textbf{0.902} [0.9022, 0.9022] & \textbf{0.867} [0.8667, 0.8667] & \textbf{0.962} [0.9624, 0.9624]
\\
& $\mathrm{CAKE}^{\text{(HM)}}$ & \textbf{0.902} [0.9022, 0.9022] & \textbf{0.867} [0.8667, 0.8667] & \textbf{0.962} [0.9624, 0.9624]
\\

 \cmidrule(lr){2-5}

\multirow{7}{*}{S5} & Full & 0.593 [0.4434, 0.7435] & 0.660 [0.5739, 0.7462] & 0.782 [0.6384, 0.9254]
\\
& Random & 0.549 [0.3891, 0.7089] & 0.634 [0.5415, 0.7267] & 0.743 [0.5933, 0.8933]
\\
& Consensus & 0.479 [0.4780, 0.4807] & 0.570 [0.5689, 0.5710] & 0.708 [0.7046, 0.7116]
\\
& $C$ component & 0.688 [0.5929, 0.7831] & 0.697 [0.6399, 0.7540] & 0.867 [0.7988, 0.9355]
\\
& $\tilde S$ component & 0.840 [0.6821, 0.9980] & 0.855 [0.7557, 0.9535] & 0.894 [0.7637, 1.0000]
\\
& $\mathrm{CAKE}^{\text{(PR)}}$ & 0.908 [0.9075, 0.9075] & 0.879 [0.8792, 0.8792] & 0.966 [0.9664, 0.9664] 
\\
& $\mathrm{CAKE}^{\text{(HM)}}$ & \textbf{0.925} [0.9245, 0.9245] & \textbf{0.898} [0.8979, 0.8979] & \textbf{0.973} [0.9729, 0.9729] 
\\

 \cmidrule(lr){2-5}

\multirow{7}{*}{S6} & Full & 0.303 [0.3006, 0.3045] & 0.473 [0.4720, 0.4748] & 0.681 [0.6794, 0.6834] 
\\
& Random & 0.315 [0.2978, 0.3325] & 0.480 [0.4690, 0.4920] & 0.689 [0.6748, 0.7032]
\\
& Consensus & 0.561 [0.5537, 0.5684] &0.586 [0.5780, 0.5943] & 0.779 [0.7712, 0.7859]
\\
& $C$ component & 0.565 [0.5640, 0.5654] & 0.590 [0.5896, 0.5907] & 0.782 [0.7815, 0.7828]
\\
& $\tilde S$ component & 0.566 [0.5506, 0.5811] & 0.590 [0.5770, 0.6036] & 0.781 [0.7654, 0.7960]
\\
& $\mathrm{CAKE}^{\text{(PR)}}$ & 0.558 [0.5302, 0.5860] & 0.583 [0.5573, 0.6095] & 0.777 [0.7524, 0.8006]
\\
& $\mathrm{CAKE}^{\text{(HM)}}$ & \textbf{0.573} [0.5662, 0.5798] & \textbf{0.596} [0.5893, 0.6034] & \textbf{0.789} [0.7817, 0.7962] 
\\

 \cmidrule(lr){2-5}

\multirow{7}{*}{S7} & Full & 0.615 [0.4786, 0.7523] & 0.649 [0.5642, 0.7333] & 0.802 [0.6727, 0.9317] 
\\
& Random & 0.602 [0.4685, 0.7363] & 0.636 [0.5464, 0.7253] & 0.797 [0.6691, 0.9257]
\\
& Consensus & 0.460 [0.4371, 0.4820]	 & 0.545 [0.5385, 0.5519] & 0.679 [0.6411, 0.7165]
\\
& $C$ component & 0.450 [0.4267, 0.4725] & 0.540 [0.5326, 0.5480] & 0.668 [0.6329, 0.7035]
\\
& $\tilde S$ component & 0.713 [0.5401, 0.8869]	 &0.762 [0.6529, 0.8713] & 0.820 [0.6817, 0.9582]	
\\
& $\mathrm{CAKE}^{\text{(PR)}}$ & \textbf{0.861} [0.7566, 0.9662] & \textbf{0.861} [0.7908, 0.9309] & \textbf{0.939} [0.8647, 1.0000] 
\\
& $\mathrm{CAKE}^{\text{(HM)}}$ & 0.678 [0.4982, 0.8581] & 0.741 [0.6239, 0.8579] & 0.797 [0.6623, 0.9308] 
\\

 \midrule

\multirow{7}{*}{\textsc{Ir}} & Full & 0.692 [0.6255, 0.7580] & 0.728 [0.6907, 0.7651] & 0.857 [0.7859, 0.9288] 
\\
& Random & 0.643 [0.5586, 0.7277] & 0.697 [0.6525, 0.7425] & 0.823 [0.7323, 0.9134]	 
\\
& Consensus & 0.658 [0.5951, 0.7206] & 0.733 [0.7047, 0.7620] & 0.835 [0.7681, 0.9024]
\\
& $C$ component & 0.658 [0.5951, 0.7206] & 0.733 [0.7047, 0.7620] & 0.835 [0.7681, 0.9024]
\\
& $\tilde S$ component & 0.824 [0.6491, 0.9985]	 & 0.871 [0.7671, 0.9741] & 0.862 [0.7138, 1.0000] 
\\
& $\mathrm{CAKE}^{\text{(PR)}}$ & 0.855 [0.7552, 0.9541] & 0.865 [0.8118, 0.9190] & \textbf{0.923} [0.8345, 1.0000] 
\\
& $\mathrm{CAKE}^{\text{(HM)}}$ & \textbf{0.874} [0.7222, 1.0000] & \textbf{0.900} [0.8097, 0.9908] & 0.909 [0.7849, 1.0000] 
\\

\cmidrule(lr){2-5}

\multirow{7}{*}{\textsc{Bc}} & Full & 0.663 [0.6558, 0.6698] & 0.543 [0.5343, 0.5518] & 0.908 [0.9058, 0.9101] 
\\
& Random & 0.660 [0.6598, 0.6598] & 0.534 [0.5337, 0.5337] & 0.907 [0.9070, 0.9070]
\\
& Consensus & \textbf{0.779} [0.7788, 0.7789] &\textbf{0.679} [0.6773, 0.6815] & 0.942 [0.9422, 0.9422]
\\
& $C$ component & \textbf{0.779} [0.7788, 0.7789] &\textbf{0.679} [0.6773, 0.6815] & 0.942 [0.9422, 0.9422]
\\
& $\tilde S$ component & 0.754 [0.7543, 0.7543] & 0.652 [0.6516, 0.6516] & \textbf{0.950} [0.9497, 0.9497] 
\\
& $\mathrm{CAKE}^{\text{(PR)}}$ & 0.754 [0.7543, 0.7543] & 0.652 [0.6516, 0.6516] & \textbf{0.950} [0.9497, 0.9497] 
\\
& $\mathrm{CAKE}^{\text{(HM)}}$ & 0.754 [0.7543, 0.7543] & 0.652 [0.6516, 0.6516] & \textbf{0.950} [0.9497, 0.9497] 
\\

 \cmidrule(lr){2-5}

\multirow{7}{*}{\textsc{Dg}} & Full & 0.630 [0.6032, 0.6568] & 0.731 [0.7204, 0.7421] & 0.747 [0.7153, 0.7785] 
\\
& Random & 0.594 [0.5726, 0.6145] & 0.712 [0.7009, 0.7240] & 0.713 [0.6918, 0.7351]
\\
& Consensus & 0.777 [0.7201, 0.8344] & 0.835 [0.8094, 0.8614] & 0.799 [0.7503, 0.8484]
\\
& $C$ component & 0.774 [0.7250, 0.8239] & 0.834 [0.8111, 0.8569] & 0.784 [0.7296, 0.8376]	
\\
& $\tilde S$ component & 0.763 [0.7175, 0.8092] & 0.852 [0.8316, 0.8725] & 0.779 [0.7312, 0.8268]
\\
& $\mathrm{CAKE}^{\text{(PR)}}$ & 0.741 [0.6807, 0.8013] & 0.842 [0.8126, 0.8711] & 0.744 [0.6829, 0.8055]	 
\\
& $\mathrm{CAKE}^{\text{(HM)}}$ & \textbf{0.793} [0.7466, 0.8392] & \textbf{0.866} [0.8452, 0.8875] & \textbf{0.815} [0.7730, 0.8575] 
\\

 \cmidrule(lr){2-5}
 
\multirow{7}{*}{\textsc{Ng}} & Full & 0.352 [0.3421, 0.3610] & 0.509 [0.5000, 0.5179] & 0.511 [0.4957, 0.5259] 
\\
& Random & 0.359 [0.3505, 0.3675] & 0.516 [0.5080, 0.5232] & 0.515 [0.4975, 0.5332]	
\\
& Consensus & 0.462 [0.4500, 0.4735] & 0.586 [0.5819, 0.5895] & 0.575 [0.5581, 0.5922]
\\
& $C$ component & 0.473 [0.4607, 0.4863] & 0.590 [0.5831, 0.5977] & 0.578 [0.5586, 0.5973]
\\
& $\tilde S$ component & 0.467 [0.4437, 0.4903] & 0.609 [0.5998, 0.6183] & 0.561 [0.5322, 0.5892]
\\
& $\mathrm{CAKE}^{\text{(PR)}}$ & \textbf{0.494} [0.4766, 0.5123] & \textbf{0.618} [0.6118, 0.6234] & \textbf{0.603} [0.5820, 0.6246] 
\\
& $\mathrm{CAKE}^{\text{(HM)}}$ & 0.467 [0.4437, 0.4903] & 0.609 [0.5998, 0.6183] & 0.561 [0.5322, 0.5892]
\\

 \cmidrule(lr){2-5}
 
\multirow{7}{*}{\textsc{Fm}} & Full & 0.353 [0.3416, 0.3641] & 0.506 [0.4966, 0.5163] & 0.514 [0.4880, 0.5391] 
\\
& Random & 0.362 [0.3495, 0.3746] & 0.509 [0.4994, 0.5193] & 0.537 [0.5127, 0.5611]
\\
& Consensus & 0.433 [0.4199, 0.4458]	 & 0.560 [0.5533, 0.5666] & 0.596 [0.5825, 0.6097]
\\
& $C$ component & 0.427 [0.4111, 0.4421] & 0.560 [0.5517, 0.5686] & 0.596 [0.5742, 0.6177]	
\\
& $\tilde S$ component & 0.447 [0.4295, 0.4637] & 0.585 [0.5740, 0.5961] & 0.568 [0.5357, 0.5999] 
\\
& $\mathrm{CAKE}^{\text{(PR)}}$ & 0.445 [0.4149, 0.4744] & 0.588 [0.5748, 0.6012] & 0.566 [0.5274, 0.6046]
\\
& $\mathrm{CAKE}^{\text{(HM)}}$ & \textbf{0.474} [0.4467, 0.5007] & \textbf{0.595} [0.5791, 0.6111] & \textbf{0.599} [0.5586, 0.6402] 
\\

 \cmidrule(lr){2-5}
 
\multirow{7}{*}{\textsc{Pd}} & Full & 0.541 [0.5121, 0.5690] & 0.676 [0.6667, 0.6846] & 0.669 [0.6326, 0.7057] 
\\
& Random & 0.566 [0.5336, 0.5976] & 0.682 [0.6725, 0.6917] & 0.712 [0.6633, 0.7611]
\\
& Consensus & 0.653 [0.6190, 0.6874] & 0.734 [0.7210, 0.7467] & 0.718 [0.6904, 0.7451]
\\
& $C$ component & 0.626 [0.5952, 0.6563] & 0.721 [0.7078, 0.7345] & 0.704 [0.6768, 0.7316]
\\
& $\tilde S$ component & 0.754 [0.6854, 0.8226] & \textbf{0.842} [0.8137, 0.8695] & 0.803 [0.7354, 0.8708]	
\\
& $\mathrm{CAKE}^{\text{(PR)}}$ & 0.724 [0.6749, 0.7740] & 0.819 [0.7981, 0.8406] & 0.784 [0.7372, 0.8309]
\\
& $\mathrm{CAKE}^{\text{(HM)}}$ & \textbf{0.759} [0.7166, 0.8010]	 & \textbf{0.842} [0.8242, 0.8597] & \textbf{0.810} [0.7665, 0.8536] 
\\

 \cmidrule(lr){2-5}

\multirow{7}{*}{\textsc{Lt}} & Full & 0.152 [0.1456, 0.1577]	 & 0.368 [0.3634, 0.3727] & 0.281 [0.2751, 0.2869] 
\\
& Random & 0.152 [0.1476, 0.1557] & 0.373 [0.3675, 0.3794] & 0.284 [0.2790, 0.2888]
\\
& Consensus & 0.199 [0.1939, 0.2031] & 0.430 [0.4264, 0.4333] & 0.324 [0.3177, 0.3306]	
\\
& $C$ component & 0.210 [0.2031, 0.2169] & 0.440 [0.4348, 0.4461] & 0.333 [0.3230, 0.3434]
\\
& $\tilde S$ component & 0.213 [0.2034, 0.2234] & 0.458 [0.4492, 0.4665] & 0.334 [0.3203, 0.3472]
\\
& $\mathrm{CAKE}^{\text{(PR)}}$ & 0.219 [0.2096, 0.2277] & 0.458 [0.4495, 0.4659] & 0.328 [0.3176, 0.3380] 
\\
& $\mathrm{CAKE}^{\text{(HM)}}$ & \textbf{0.222} [0.2163, 0.2267]	 & \textbf{0.460} [0.4555, 0.4643] & \textbf{0.335} [0.3277, 0.3422] \\

 \cmidrule(lr){2-5}
 
\multirow{7}{*}{\textsc{Sa}} & Full & 0.510 [0.4657, 0.5550] & 0.595 [0.5556, 0.6346] & 0.663 [0.6298, 0.6972] 
\\
& Random & 0.520 [0.4927, 0.5472] & 0.608 [0.5895, 0.6273] & 0.672 [0.6542, 0.6894]
\\
& Consensus & 0.523 [0.4794, 0.5669] &0.575 [0.5549, 0.5945] & 0.706 [0.6551, 0.7569]	
\\
& $C$ component & 0.523 [0.4794, 0.5669] &0.575 [0.5549, 0.5945] & 0.706 [0.6551, 0.7569]	
\\
& $\tilde S$ component & \textbf{0.710} [0.7080, 0.7125] & \textbf{0.764} [0.7584, 0.7703] & \textbf{0.799} [0.7842, 0.8138]
\\
& $\mathrm{CAKE}^{\text{(PR)}}$ & 0.682 [0.6383, 0.7249] & 0.749 [0.7268, 0.7704] & 0.772 [0.7317, 0.8129]
\\
& $\mathrm{CAKE}^{\text{(HM)}}$ & 0.653 [0.5997, 0.7067] & 0.734 [0.7087, 0.7594] & 0.750 [0.6964, 0.8037]
\\

\bottomrule

\caption{ARI, AMI, ACC (clustering accuracy) averages and $t$-95\% confidence intervals, computed over multiple independent clustering runs with varying random seeds. Results are reported for the full datasets (Full) and their filtered subsets mentioned in \S\ref{subse:evalset}. Highest values in \textbf{bold}.}
\label{tab:results}
\end{longtable}

\normalsize 
\vspace{-3mm}
\noindent \textbf{Coverage-accuracy evaluation.}
CAKE is evaluated as a pointwise reliability score for ensemble clustering by measuring clustering accuracy as a function of retained coverage. For each dataset, per-instance $\mathrm{CAKE}^{(\mathrm{PR})}$ and $\mathrm{CAKE}^{(\mathrm{HM})}$ scores are computed from an ensemble of $R{=}20$ \emph{k}-means partitions. 
A reference \emph{k}-means solution is fitted on the full dataset, and its accuracy is evaluated against ground truth after label alignment. For a grid of coverage levels $c \in [0.1,1.0]$, instances are ranked by CAKE and the reference accuracy restricted to the top-$c$ fraction of instances is measured. Plotting accuracy as a function of coverage yields coverage–accuracy curves, which show how well each CAKE variant orders instances from easier to harder and enables a direct comparison of $\mathrm{CAKE}^{(\mathrm{PR})}$ and $\mathrm{CAKE}^{(\mathrm{HM})}$ (Fig.~\ref{fig:acc_cov_cake}). Sharper curves (higher accuracy at lower coverage) indicate that CAKE concentrates correctly assigned points among the highest scores (see also Appendix~\ref{app:sec:valid} Fig.~\ref{app:fig:evaluation} for a percentile-based clustering accuracy view).

\begin{center}
{\captionsetup{type=figure} 

  \begin{subfigure}{\linewidth}
    \centering
    \includegraphics[width=0.95\linewidth]{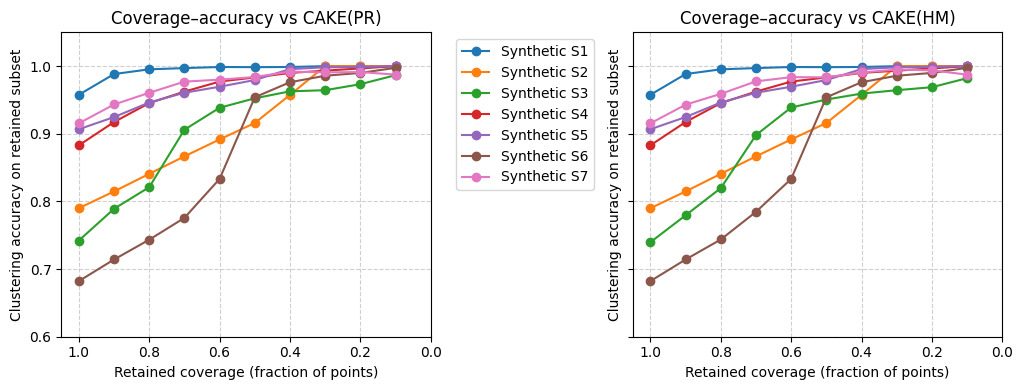}
    \caption{Coverage--accuracy on \textbf{synthetic} datasets: \textcolor{MidnightBlue}{S1}, \textcolor{orange}{S2}, \textcolor{OliveGreen}{S3}, \textcolor{red}{S4}, \textcolor{Plum}{S5}, \textcolor{RawSienna}{S6}, \textcolor{RubineRed}{S7}.}
    \label{fig:acc_cov_cake_synth}
  \end{subfigure}

  \vspace{0.6em}

  \begin{subfigure}{\linewidth}
    \centering
    \includegraphics[width=0.95\linewidth]{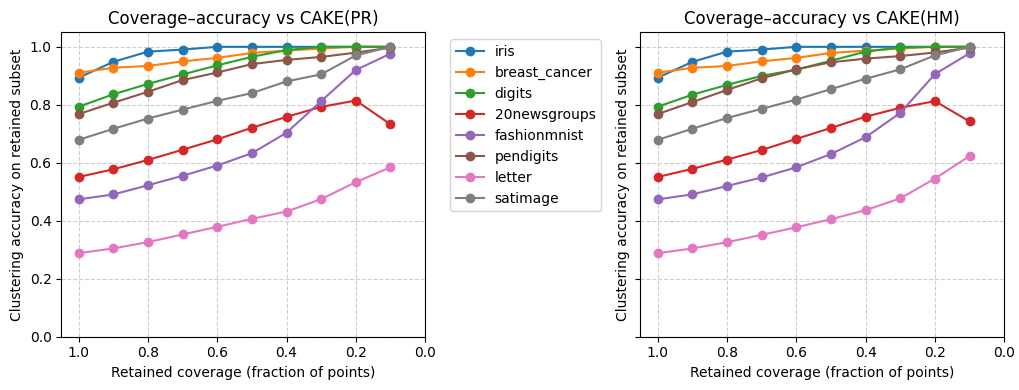}
    \caption{Coverage--accuracy on \textbf{real} datasets: \textcolor{MidnightBlue}{\textsc{Ir}}, \textcolor{orange}{\textsc{Bc}}, \textcolor{OliveGreen}{\textsc{Dg}}, \textcolor{red}{\textsc{Ng}}, \textcolor{Plum}{\textsc{Fm}}, \textcolor{RawSienna}{\textsc{Pd}}, \textcolor{RubineRed}{\textsc{Lt}}, \textcolor{darkgray}{\textsc{Sa}}.}
    \label{fig:acc_cov_cake_real}
  \end{subfigure}

  \caption{Clustering accuracy on the retained subset as a function of retained coverage, using $\mathrm{CAKE}^{(\mathrm{PR})}$ (left) and $\mathrm{CAKE}^{(\mathrm{HM})}$ (right).
  Top (\ref{fig:acc_cov_cake_synth}): synthetic datasets. Bottom (\ref{fig:acc_cov_cake_real}): real datasets.}
  \label{fig:acc_cov_cake}
}
\end{center}

\noindent \textbf{AUROC/AUPRC on consensus correctness.}
CAKE is evaluated by its ability to distinguish \emph{correctly} clustered instances from \emph{incorrect} ones. For each dataset, a consensus partition $z^\star$ is first computed from the clustering ensemble and a binary correctness label
$\mathds{1}\!\left[\pi(z^\star_i)=y_i\right]$ is defined, where $\pi$ denotes the Hungarian alignment mapping from consensus labels to ground-truth classes. Given a confidence score (higher indicates higher reliability), discriminative power is measured using \mbox{AUROC} (area under the ROC curve) and \mbox{AUPRC} (area under the precision-recall curve, i.e., average precision) when predicting $\mathds{1}\!\left[\pi(z^\star_i)=y_i\right]{=}1$.
Four signals are compared: (\textit{i}) $\mathrm{CAKE}^{(\mathrm{PR})}$ and (\textit{ii}) $\mathrm{CAKE}^{(\mathrm{HM})}$, which are computed from the same ensemble. (\textit{iii}) \emph{Entropy-based agreement}: for each run $r$, its labels are aligned to $z^\star$, yielding aligned labels $\tilde z^{(r)}$. This induces, for each point, an empirical distribution
$p_{i\ell} = \frac{1}{R}\sum_{r=1}^{R}\mathds{1}\!\left[\tilde z^{(r)}_i=\ell\right]$, with normalized entropy
$H_i \;=\; -\sum_{\ell=1}^{k} p_{i\ell}\log\!\bigl(p_{i\ell}+\varepsilon\bigr)
\ \Rightarrow \
\hat H_i \;=\; 1 - (H_i/\log k),$
so that $\hat H_i\in[0,1]$ is high when aligned votes concentrate on a single label and low when the ensemble is uncertain. (\textit{iv}) \emph{Bootstrap-based stability}: $B$ subsamples of size $\lfloor 0.8n \rfloor$ are drawn and \emph{k}-means is run on each subsample, obtaining partitions $\{z^{(b)}\}_{b=1}^B$.
For each point $i$, all pairs of subsamples $(b,b')$ in which $i$ appears are considered. For each such pair, $z^{(b')}$ is aligned to $z^{(b)}$ on the points shared by the two subsamples via Hungarian matching, and agreement is counted according to whether $i$ receives the same label in the aligned pair. The score is the average pairwise agreement:
${\mathrm{boot}}_i
\;=\;
\frac{1}{|\mathcal P_i|}
\sum_{(b,b')\in \mathcal P_i}
\mathds{1}\!\left[z^{(b)}_i=\tilde z^{(b')}_i\right],$
where $\mathcal P_i$ is the set of subsample-pairs containing $i$, and $\tilde z^{(b')}$ denotes the labels of
run $b'$ after alignment to $b$.\\ \noindent$B{=}20$ is used to match the default ensemble size $R{=}20$, yielding a comparable computational budget between baselines. This evaluation (Table~\ref{tab:pointwise_all}) quantifies discriminative power at the point level (see Appendix~\ref{app:sec:valid} Tables~\ref{tab:pointwise_all_minibatchkmeans}--\ref{tab:pointwise_all_gmm} for the same evaluation using MiniBatchKMeans/K-Medoids/GMM).
As a complement to Table~\ref{tab:pointwise_all}, it is also tested whether each signal yields a monotonic reliability ordering. Points are binned by score percentiles and Spearman’s correlation ($\rho_\text{Spear.}$) between percentile rank and within-bin clustering accuracy on representative datasets is reported (Fig.~\ref{fig:corr_scores_accuracy}).

\begin{center}
\small
\setlength{\tabcolsep}{6pt}
\begin{tabular}{llcccc}
\toprule
Dataset & Metric & $\mathrm{CAKE}^{(\mathrm{PR})}$ & $\mathrm{CAKE}^{(\mathrm{HM})}$ & Entropy agreement & Bootstrap stability \\

\midrule
\multirow{2}{*}{S1} & AUPRC   & \textbf{0.997} & \textbf{0.997} & 0.958 & 0.960 \\
                   & AUROC   & \textbf{0.932} & \textbf{0.932} & 0.506 & 0.519 \\
\cmidrule(lr){2-6}
\multirow{2}{*}{S2} & AUPRC   & 0.930 & \textbf{0.931} & 0.785 & 0.807 \\
                   & AUROC   & 0.761 & \textbf{0.764} & 0.488 & 0.553 \\
\cmidrule(lr){2-6}
\multirow{2}{*}{S3} & AUPRC   & \textbf{0.932} & \textbf{0.932} & 0.852 & 0.819 \\
                   & AUROC   & 0.842 & \textbf{0.844} & 0.723 & 0.665 \\
\cmidrule(lr){2-6}
\multirow{2}{*}{S4} & AUPRC   & \textbf{0.976} & \textbf{0.976} & 0.887 & 0.889 \\
                   & AUROC   & \textbf{0.843} & \textbf{0.843} & 0.514 & 0.528 \\
\cmidrule(lr){2-6}
\multirow{2}{*}{S5} & AUPRC   & 0.978 & \textbf{0.980} & 0.904 & 0.908 \\
                   & AUROC   & 0.823 & \textbf{0.834} & 0.499 & 0.509 \\
\cmidrule(lr){2-6}
\multirow{2}{*}{S6} & AUPRC   & 0.923 & \textbf{0.924} & 0.763 & 0.754 \\
                   & AUROC   & 0.821 & \textbf{0.823} & 0.657 & 0.621 \\
\cmidrule(lr){2-6}
\multirow{2}{*}{S7} & AUPRC   & 0.975 & \textbf{0.977} & 0.933 & 0.919 \\
                   & AUROC   & 0.817 & \textbf{0.826} & 0.602 & 0.575 \\
\midrule

\multirow{2}{*}{\textsc{Ir}}           & AUPRC  & 0.977 & \textbf{0.984} & 0.859 & 0.923 \\
                               & AUROC  & 0.840 & \textbf{0.888} & 0.328 & 0.675 \\
\cmidrule(lr){2-6}

\multirow{2}{*}{\textsc{Bc}} & AUPRC  & \textbf{0.971} & \textbf{0.971} & 0.915 & 0.929 \\
                               & AUROC  & \textbf{0.773} & \textbf{0.773} & 0.555 & 0.636 \\
\cmidrule(lr){2-6}

\multirow{2}{*}{\textsc{Dg}}         & AUPRC  & 0.909 & \textbf{0.912} & 0.734 & 0.869 \\
                               & AUROC  & 0.767 & \textbf{0.777} & 0.526 & 0.717 \\
\cmidrule(lr){2-6}

\multirow{2}{*}{\textsc{Ng}}   & AUPRC  & \textbf{0.638} & 0.624 & 0.623 & 0.637 \\
                               & AUROC  & \textbf{0.678} & 0.670 & 0.636 & 0.637 \\
\cmidrule(lr){2-6}

\multirow{2}{*}{\textsc{Fm}}   & AUPRC  & \textbf{0.758} & 0.755 & 0.631 & 0.751 \\
                               & AUROC  & 0.707 & 0.703 & 0.667 & \textbf{0.742} \\
\cmidrule(lr){2-6}

\multirow{2}{*}{\textsc{Pd}}      & AUPRC  & 0.951 & \textbf{0.954} & 0.900 & 0.864 \\
                               & AUROC  & 0.857 & \textbf{0.866} & 0.769 & 0.718 \\
\cmidrule(lr){2-6}

\multirow{2}{*}{\textsc{Lt}}         & AUPRC  & \textbf{0.465} & \textbf{0.465} & 0.402 & 0.409 \\
                               & AUROC  & 0.661 & \textbf{0.664} & 0.617 & 0.602 \\
\cmidrule(lr){2-6}

\multirow{2}{*}{\textsc{Sa}}       & AUPRC  & 0.890 & \textbf{0.894} & 0.706 & 0.725 \\
                               & AUROC  & 0.784 & \textbf{0.788} & 0.560 & 0.592 \\
\bottomrule
\end{tabular}
\captionsetup{hypcap=false}
\captionof{table}{Per-point \emph{consensus correctness} prediction (\mbox{AUPRC} and \mbox{AUROC}) with $\mathrm{CAKE}^{(\mathrm{PR})}$, $\mathrm{CAKE}^{(\mathrm{HM})}$, entropy-based agreement, and bootstrap-based stability. The highest value for each dataset and metric is highlighted in \textbf{bold}.}
\label{tab:pointwise_all}
\end{center}

\begin{center}
\centering
\includegraphics[width=\linewidth]{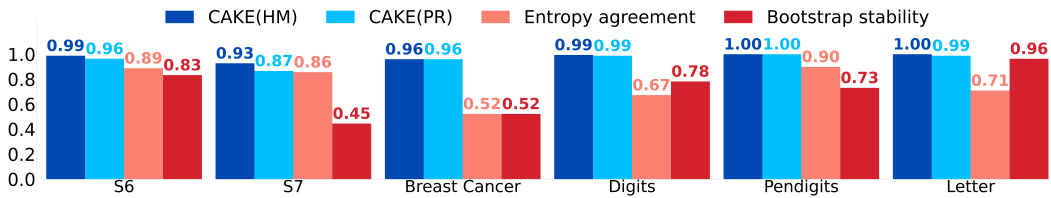}
\captionsetup{hypcap=false}
\captionof{figure}{
Correlation ($\rho_\text{Spear.}$) between confidence-score percentiles and per-bin clustering accuracy.}
\label{fig:corr_scores_accuracy}
\end{center}

\newpage

\noindent Beyond the ensemble-derived baselines above, additional comparisons were conducted against native soft and density-based confidence signals. In particular, two Fuzzy C-Means (FCM) confidence scores were considered: (\textit{i}) maximum membership, defined as the largest membership weight assigned to a point $\mathrm{FCM}^{(1)}_i = \max_{\ell} u_{i\ell}$, where $ u_{i\ell}$ is the FCM membership of point $x_i$ to cluster $\ell$, and (\textit{ii}) entropy-based confidence, defined as one minus the normalized entropy of the membership vector $\mathrm{FCM}^{(2)}_i
= 1-\frac{H_i}{\log k},
\text{ where }H_i = -\sum_{\ell=1}^{k} u_{i\ell}\log\!\bigl(u_{i\ell}+\varepsilon\bigr)$. In addition, two density-based baselines were examined. The first is a standard global Local Outlier Factor (LOF)~\parencite{lof} inlierness score $\mathrm{LOF}^{(1)}_i$, obtained from the negative outlier factor so that larger values indicate more typical (less outlier-like) observations. The second is an ensemble cluster-local LOF baseline $\mathrm{LOF}^{(2)}_i$, in which a LOF-based score is computed within the assigned cluster of each run in the same ensemble used by CAKE, and then averaged across runs. This yields a cluster-aware, ensemble-averaged density-based confidence score. Comparisons of these baselines against $\mathrm{CAKE}^{(\mathrm{HM})}$ are reported in Table~\ref{tab:pointwise_fcm_densitybased}. 
\noindent Supplementary comparisons against native GMM-based confidence signals, with $\mathrm{CAKE}^{(\mathrm{HM})}$ computed from GMM ensembles, are provided in Appendix~\ref{app:sec:gmm_confidence}.

\begin{center}
\small
\setlength{\tabcolsep}{6pt}
\begin{tabular}{llccccc}
\toprule
Dataset & Metric & $\mathrm{CAKE}^{(\mathrm{HM})}$ & $\mathrm{FCM}^{(1)}$ & $\mathrm{FCM}^{(2)}$ & $\mathrm{LOF}^{(1)}$ & $\mathrm{LOF}^{(2)}$\\

\midrule
\multirow{2}{*}{S1} & AUPRC   & \textbf{0.997} & 0.995 & 0.994 & 0.966 & 0.970 \\
                    & AUROC   & \textbf{0.932} & 0.911 & 0.895 & 0.541 & 0.619 \\
\cmidrule(lr){2-7}
\multirow{2}{*}{S2} & AUPRC   & \textbf{0.931} & 0.878 & 0.869 & 0.769 & 0.774 \\
                    & AUROC   & \textbf{0.764} & 0.682 & 0.655 & 0.442 & 0.459 \\
\cmidrule(lr){2-7}
\multirow{2}{*}{S3} & AUPRC   & 0.932 & \textbf{0.961} & \textbf{0.961} & 0.680 & 0.716 \\
                    & AUROC   & 0.844 & 0.942 & \textbf{0.943} & 0.550 & 0.587 \\
\cmidrule(lr){2-7}
\multirow{2}{*}{S4} & AUPRC   & \textbf{0.976} & 0.970 & 0.969 & 0.891 & 0.910 \\
                    & AUROC   & \textbf{0.843} & 0.826 & 0.819 & 0.515 & 0.598 \\
\cmidrule(lr){2-7}
\multirow{2}{*}{S5} & AUPRC   & 0.980 & \textbf{0.982} & \textbf{0.982} & 0.954 & 0.943 \\
                    & AUROC   & 0.834 & 0.855 & \textbf{0.857} & 0.682 & 0.631 \\
\cmidrule(lr){2-7}
\multirow{2}{*}{S6} & AUPRC   & 0.924 & \textbf{0.931} & 0.929 & 0.742 & 0.782 \\
                    & AUROC   & 0.823 & \textbf{0.832} & 0.827 & 0.617 & 0.643 \\
\cmidrule(lr){2-7}
\multirow{2}{*}{S7} & AUPRC   & \textbf{0.977} & 0.970 & 0.970 & 0.941 & 0.944 \\
                    & AUROC   & \textbf{0.826} & 0.767 & 0.768 & 0.635 & 0.630 \\
\midrule

\multirow{2}{*}{\textsc{Ir}} & AUPRC & 0.984 & \textbf{0.993} & 0.992 & 0.890 & 0.902 \\
                              & AUROC & 0.888 & \textbf{0.940} & 0.926 & 0.500 & 0.559 \\
\cmidrule(lr){2-7}

\multirow{2}{*}{\textsc{Bc}} & AUPRC & 0.971 & \textbf{0.985} & \textbf{0.985} & 0.948 & 0.960 \\
                              & AUROC & 0.773 & \textbf{0.889} & \textbf{0.889} & 0.639 & 0.704 \\
\cmidrule(lr){2-7}

\multirow{2}{*}{\textsc{Dg}} & AUPRC & \textbf{0.912} & 0.686 & 0.698 & 0.775 & 0.789 \\
                              & AUROC & \textbf{0.777} & 0.432 & 0.431 & 0.599 & 0.634 \\
\cmidrule(lr){2-7}

\multirow{2}{*}{\textsc{Ng}} & AUPRC & \textbf{0.624} & 0.489 & 0.489 & 0.598 & 0.598 \\
                              & AUROC & \textbf{0.670} & 0.459 & 0.456 & 0.589 & 0.608 \\
\cmidrule(lr){2-7}

\multirow{2}{*}{\textsc{Fm}} & AUPRC & \textbf{0.755} & 0.578 & 0.575 & 0.604 & 0.613 \\
                              & AUROC & \textbf{0.703} & 0.498 & 0.504 & 0.583 & 0.587 \\
\cmidrule(lr){2-7}

\multirow{2}{*}{\textsc{Pd}} & AUPRC & \textbf{0.954} & 0.939 & 0.930 & 0.783 & 0.822 \\
                              & AUROC & \textbf{0.866} & 0.823 & 0.807 & 0.537 & 0.598 \\
\cmidrule(lr){2-7}

\multirow{2}{*}{\textsc{Lt}} & AUPRC & \textbf{0.465} & 0.227 & 0.229 & 0.301 & 0.333 \\
                              & AUROC & \textbf{0.664} & 0.381 & 0.385 & 0.514 & 0.546 \\
\cmidrule(lr){2-7}

\multirow{2}{*}{\textsc{Sa}} & AUPRC & 0.894 & \textbf{0.898} & 0.896 & 0.707 & 0.713 \\
                              & AUROC & 0.788 & \textbf{0.795} & 0.788 & 0.555 & 0.561 \\
\bottomrule
\end{tabular}
\captionsetup{hypcap=false}
\captionof{table}{Per-point \emph{consensus correctness} prediction (\mbox{AUPRC} and \mbox{AUROC}) with $\mathrm{CAKE}^{(\mathrm{HM})}$, $\ \mathrm{FCM}^{(1)}$ (max membership), $\mathrm{FCM}^{(2)}$ (entropy confidence), $\mathrm{LOF}^{(1)}$ (global LOF), and $\mathrm{LOF}^{(2)}$ (ensemble-averaged cluster-local LOF). The highest value for each dataset and metric is shown in \textbf{bold}.}
\label{tab:pointwise_fcm_densitybased}
\end{center}

\noindent \textbf{Error discovery via confidence ranking.}
CAKE is tested for flagging clustering errors. For each dataset, a $k$-means++ model is fitted at the ground-truth $k$, predicted labels are aligned to ground truth, and misclustered points are marked as positives (``errors''). 
Three scores are compared: CAKE$^{\text{(HM)}}$ (Eq.~\ref{eq:cake:b}; $R{=}20$ random \emph{k}-means runs with centroid-based Silhouette approximation), \emph{GMM} $p_{\max}$ (max posterior under a $k$-component GMM), and their rank-average \mbox{fusion}. Ranking points from low to high confidence, AUPRC on the error class is reported (Fig.~\ref{fig:error_discovery}).
\vspace{-2mm}
\begin{center}
  \includegraphics[width=0.65\linewidth]{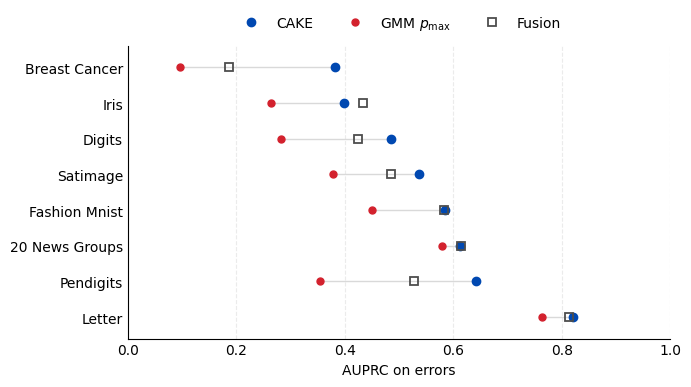}
  \captionsetup{hypcap=false}
  \captionof{figure}{AUPRC on the error class. Markers show per-dataset AUPRC; \textcolor{gray}{light gray connectors} join \textcolor{red}{GMM \(p_{\max}\)} to \textcolor{blue}{CAKE} to highlight the per-dataset CAKE score improvement. A simple fusion of these two is marked by $\square$.}
  \label{fig:error_discovery}
\end{center}

\noindent \textbf{Beyond \emph{k}-means clustering.}
CAKE is model-agnostic: it can be applied to ensembles produced by a range of base algorithms. To illustrate this, $\mathrm{CAKE}^{\text{(HM)}}$ is computed under three ensemble constructions: (\textit{i}) homogeneous \emph{k}-means, (\textit{ii}) homogeneous GMM, and (\textit{iii}) heterogeneous \emph{k}-means$+$GMM, on representative datasets (Fig.~\ref{fig:ensembles-real-synth}). A broader ablation on clustering ensemble design and its effect on CAKE score (ranking) quality is provided in Appendix~\ref{app:sec:divens}.
\vspace{-2mm}
\begin{center}
{\captionsetup{type=figure}

  \begin{subfigure}[b]{0.43\linewidth}
    \centering
    \includegraphics[width=\linewidth]{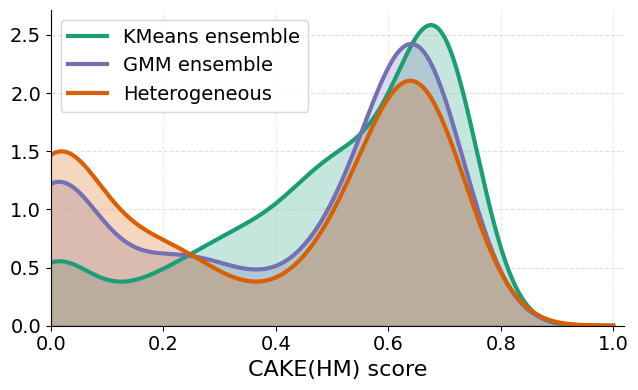}
    \caption{Breast Cancer}
    \label{fig:bc-ensembles}
  \end{subfigure}\hfill
  \begin{subfigure}[b]{0.43\linewidth}
    \centering
    \includegraphics[width=\linewidth]{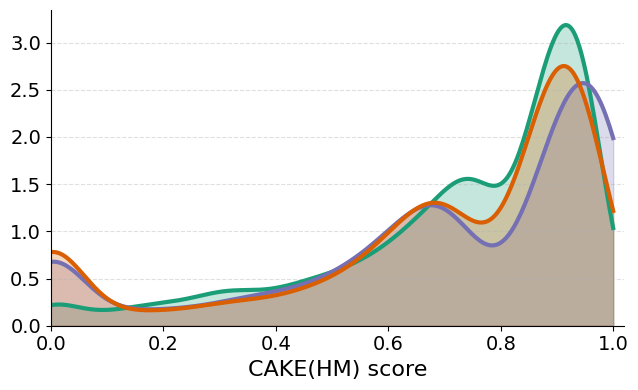}
    \caption{Synthetic S5}
    \label{fig:s5-ensembles}
  \end{subfigure}

  \begin{subfigure}[b]{0.43\linewidth}
    \centering
    \includegraphics[width=\linewidth]{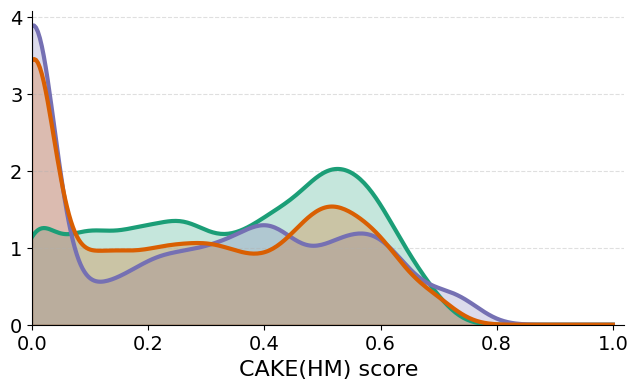}
    \caption{Pendigits}
    \label{fig:pen-ensembles}
  \end{subfigure}\hfill
  \begin{subfigure}[b]{0.43\linewidth}
    \centering
    \includegraphics[width=\linewidth]{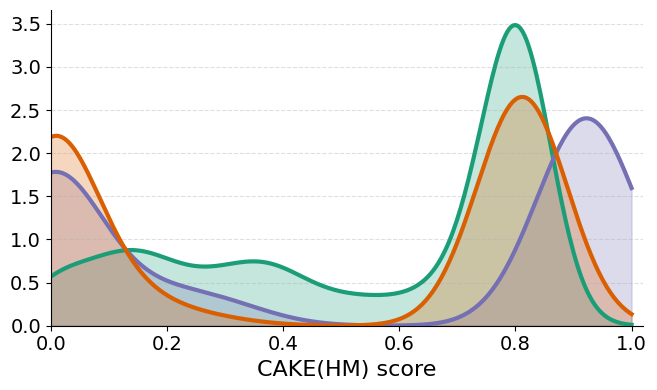}
    \caption{Synthetic S6}
    \label{fig:s6-ensembles}
  \end{subfigure}

  \caption{Distributions (KDEs) of $\mathrm{CAKE}^\text{(HM)}$ (Eq.~\ref{eq:cake:b}) under: \textcolor{ForestGreen}{homogeneous $k$-means ensembles}, \textcolor{RoyalPurple}{homogeneous GMM ensembles}, and \textcolor{Orange}{heterogeneous $k$-means$+$GMM ensembles}.}
  \label{fig:ensembles-real-synth}
}
\end{center}

\noindent \textbf{Misspecified number of clusters.} CAKE is tested under cluster-count misspecification, i.e., when the clustering uses a suboptimal number of clusters. For each $k'\!\in\!\{k\!-\!2,\dots, k\!+\!2\}$, where $k$ is the ground-truth number of clusters (Tab.~\ref{tab:realdata}), (\textit{i}) an ensemble of $R{=}30$ \emph{k}-means runs with $k'$ clusters is constructed and $\mathrm{CAKE}^{\text{(HM)}}$ scores are computed, and (\textit{ii}) an independent \emph{k}-means++ model with the same $k'$ is fitted for evaluation. 
The following are reported: (a) clustering quality via ARI with respect to ground truth and (b) CAKE ranking quality via AURC (area under the risk--coverage curve), where risk is the misclustering rate after Hungarian alignment. Higher ARI indicates better partitions, while lower AURC indicates that CAKE concentrates correct assignments among high-confidence points (Fig.~\ref{fig:kmisspec}).

\begin{center}
{\captionsetup{type=figure}

  \begin{subfigure}[b]{0.49\linewidth}
    \centering
    \includegraphics[width=0.8\linewidth]{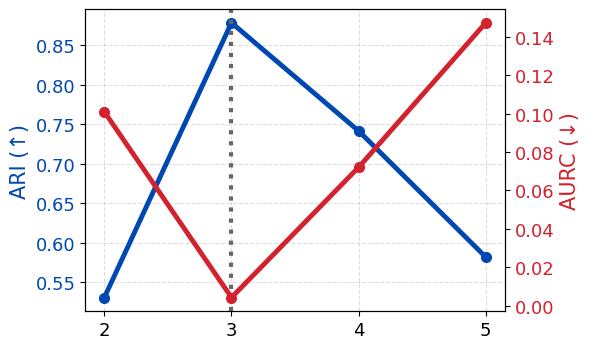}
    \caption{\textsc{S1}}
  \end{subfigure}\hfill
  \begin{subfigure}[b]{0.49\linewidth}
    \centering
    \includegraphics[width=0.8\linewidth]{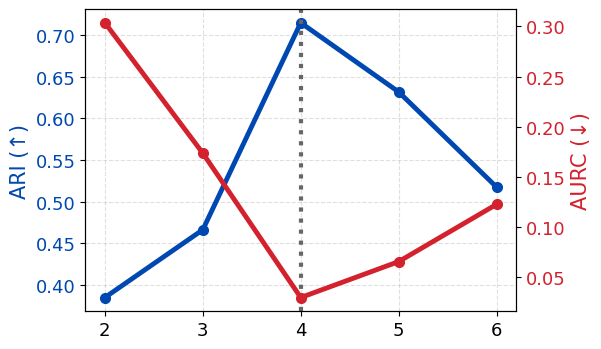}
    \caption{\textsc{S4}}
  \end{subfigure}

  \vspace{0.6em}

  \begin{subfigure}[b]{0.49\linewidth}
    \centering
    \includegraphics[width=0.8\linewidth]{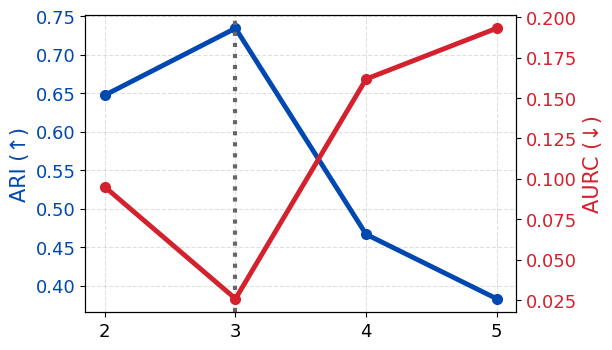}
    \caption{\textsc{S7}}
  \end{subfigure}\hfill
  \begin{subfigure}[b]{0.49\linewidth}
    \centering
    \includegraphics[width=0.8\linewidth]{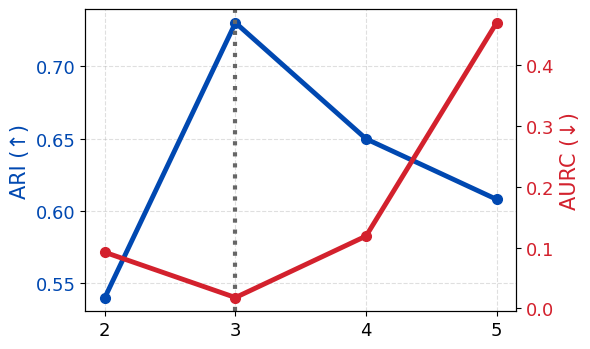}
    \caption{\textsc{Ir}}
  \end{subfigure}

  \vspace{0.6em}

  \begin{subfigure}[b]{0.49\linewidth}
    \centering
    \includegraphics[width=0.8\linewidth]{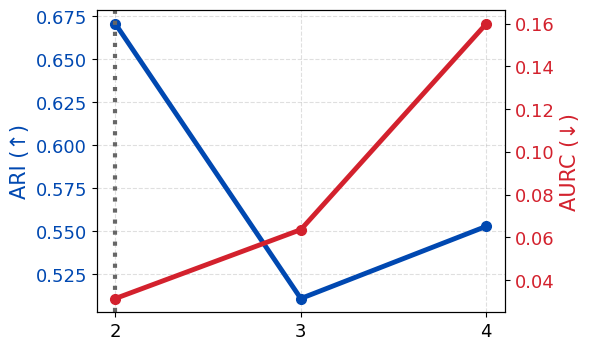}
    \caption{\textsc{Bc}}
  \end{subfigure}\hfill
  \begin{subfigure}[b]{0.49\linewidth}
    \centering
    \includegraphics[width=0.8\linewidth]{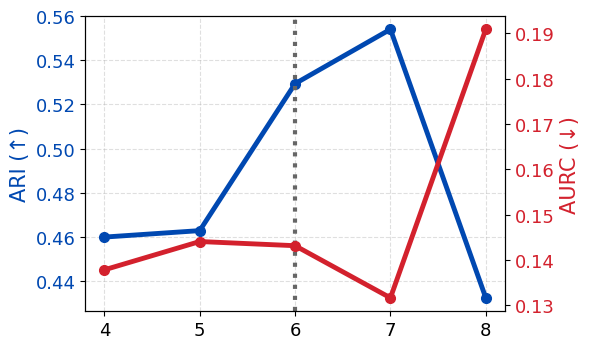}
    \caption{\textsc{Sa}}
  \end{subfigure}

  \vspace{0.6em}

  \begin{subfigure}[b]{0.49\linewidth}
    \centering
    \includegraphics[width=0.8\linewidth]{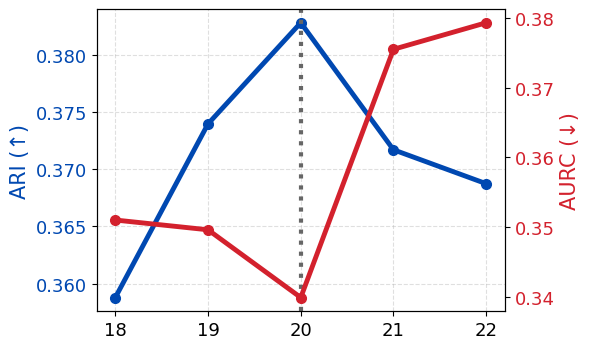}
    \caption{\textsc{Ng}}
  \end{subfigure}\hfill
  \begin{subfigure}[b]{0.49\linewidth}
    \centering
    \includegraphics[width=0.8\linewidth]{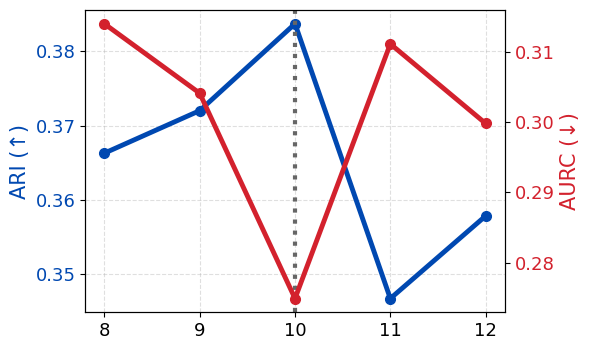}
    \caption{\textsc{Fm}}
  \end{subfigure}

  \caption{\textbf{Misspecified \emph{k}}. \textcolor{blue}{ARI} and \textcolor{red}{AURC} vs.\ $k'$. Vertical dotted line marks the ground-truth $k$.}
  \label{fig:kmisspec}
}
\end{center}

\noindent \textbf{Runtime.}
The computational cost of CAKE (Eq.~\ref{eq:cake}; Alg.~\ref{alg}) and the benefit of the centroid-based Silhouette proxy (\S\ref{subsec:silhouette}) are quantified by measuring runtime as a function of ensemble size $R$ and dataset size $n$ on synthetic Gaussian data (Fig.~\ref{fig:runtime_cake}). 
As shown in Fig.~\ref{fig:runtime_vs_R} (Table~\ref{tab:runtimer}), runtime grows approximately linearly with $R$. Figure~\ref{fig:runtime_vs_n} (Table~\ref{tab:runtimen}) shows that, when using the centroid-based proxy, runtime scales approximately linearly with $n$, whereas the exact Silhouette variant increases much more rapidly. 
Overall, these trends align with the complexity analysis in \S\ref{subsec:cake} and confirm that the proxy substantially reduces runtime while preserving high agreement with the exact scores (high Pearson correlation and low Mean Absolute Error; Tables~\ref{tab:runtimer}--\ref{tab:runtimen}).

\begin{center}
{\captionsetup{type=figure}

  \begin{subfigure}[b]{0.49\linewidth}
    \centering
    \includegraphics[width=\linewidth]{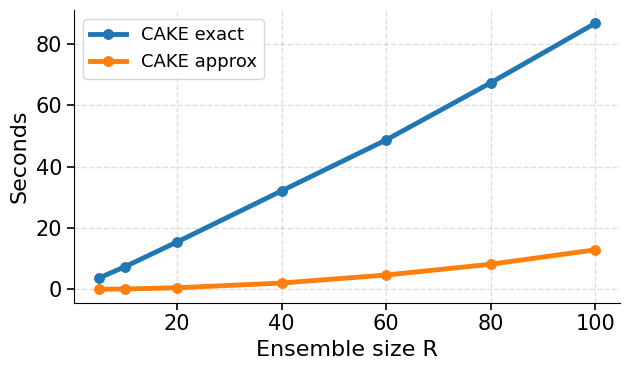}
    \caption{Runtime (s) vs. $R$ (ensemble size)}
    \label{fig:runtime_vs_R}
  \end{subfigure}\hfill
  \begin{subfigure}[b]{0.49\linewidth}
    \centering
    \includegraphics[width=\linewidth]{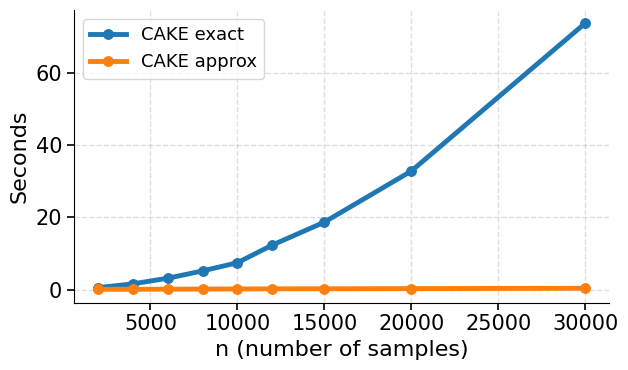}
    \caption{Runtime (s) vs. $n$ (number of samples)}
    \label{fig:runtime_vs_n}
  \end{subfigure}

  \caption{%
  Runtime in seconds (s) for $\mathrm{CAKE}^\text{(HM)}$ (Eq.~\ref{eq:cake:b}) using
  \textcolor{MidnightBlue}{exact pairwise Silhouette scores} vs.\ the
  \textcolor{orange}{centroid-based Silhouette proxy} on synthetic data.
  \textbf{Left} (\subref{fig:runtime_vs_R}): runtime as a function of ensemble size $R$
  ($n{=}10{,}000$, $d{=}20$, $k{=}10$; exact runtimes are reported in Table~\ref{tab:runtimer}).
  \textbf{Right} (\subref{fig:runtime_vs_n}): runtime as a function of the number of samples $n$
  ($d{=}20$, $k{=}10$, $R{=}20$; exact runtimes are reported in Table~\ref{tab:runtimen}).
  Experiments were run on a cloud-hosted notebook with $\sim$51\,GB RAM.%
  }
  \label{fig:runtime_cake}
}
\end{center}

\begin{center}
\centering
\small
\begin{tabular}{rccccc}
\toprule
$\mathbf{R}$ & \textbf{Ensemble construction (s)} & \textbf{CAKE exact (s)} & \textbf{CAKE approx. (s)} & \textbf{Pearson corr.} &  \textbf{MAE} \\
\midrule
 5  & 0.04 & 3.73  & 0.05 & 0.974  & 0.072  \\
10  & 0.06 & 7.42  & 0.16 & 0.972  & 0.069  \\
20  & 0.13 & 15.49 & 0.56 & 0.975  & 0.062  \\
40  & 0.25 & 32.13 & 2.10 & 0.955  & 0.066  \\
60  & 0.37 & 48.71 & 4.71 & 0.944  & 0.061  \\
80  & 0.49 & 67.32 & 8.22 & 0.936  & 0.062  \\
100 & 0.63 & 86.69 & 12.91 & 0.914 & 0.063  \\
\bottomrule
\end{tabular}
\captionsetup{hypcap=false}
\captionof{table}{Runtime in seconds (s) vs. ensemble size $R$ (Fig.~\ref{fig:runtime_vs_R}); Pearson correlation and Mean Absolute Error between exact and centroid-based Silhouette-proxy CAKE scores.}
\label{tab:runtimer}
\end{center}

\begin{center}
\centering
\small
\begin{tabular}{rccccc}
\toprule
$\mathbf{n}$ & \textbf{Ensemble construction (s)} & \textbf{CAKE exact (s)} & \textbf{CAKE approx. (s)} & \textbf{Pearson corr.} &  \textbf{MAE} \\
\midrule
2,000   & 0.02 & 0.53  & 0.06  & 0.977  & 0.065  \\
4,000   & 0.05 & 1.59  & 0.08  & 0.989  & 0.068  \\
8,000   & 0.06 & 5.18 & 0.14  & 0.974  & 0.077  \\
10,000  & 0.07 & 7.44 & 0.17  & 0.972  & 0.069  \\
15,000  & 0.08 & 18.69 & 0.22 & 0.994  & 0.054  \\
20,000  & 0.10 & 32.84 & 0.26 & 0.971  & 0.063  \\
30,000  & 0.14 & 73.84 & 0.35 & 0.982  & 0.056  \\
\bottomrule
\end{tabular}
\captionsetup{hypcap=false}
\captionof{table}{Runtime in seconds (s) vs. of number of samples $n$ (Fig.~\ref{fig:runtime_vs_n}); Pearson correlation and Mean Absolute Error between exact and centroid-based Silhouette-proxy CAKE scores.}
\label{tab:runtimen}
\end{center}

\subsection{Empirical Results}

\noindent Across synthetic and real-world datasets, clustering on the subset retained by CAKE consistently improves ARI, AMI, and ACC relative to using all points (top 70\%; Table~\ref{tab:results}, with an adaptive-thresholding approach reported in Appendix~\ref{app:sec:adafilter} Table~\ref{tab:adaptive_filtering_selected}), with similar gains in Silhouette and NMI (Appendix Table~\ref{app_extrares}). These improvements are typically accompanied by tighter confidence intervals, indicating reduced run-to-run variability.

\medskip

\noindent 
Although the strongest filtering criterion can be dataset-dependent, particularly when geometry (Eq.~\ref{eq:silhouette}) or stability (Eq.~\ref{eq:stability}) dominates, CAKE (Eq.~\ref{eq:cake}) provides the best overall trade-off as a confidence-based ranking criterion: it is best or tied-best for most dataset--metric combinations and remains near the top even when a baseline criterion is optimal on a given dataset (Tables~\ref{tab:results}, \ref{app_extrares}).

\medskip

\noindent 
The component baselines ($\tilde S, \ C$) can be competitive when one signal is highly informative. For instance, the geometric component alone is strongest on datasets where local separation is the main determinant of cluster assignments (e.g., S3 and \textsc{Sa}, where $\tilde S$ attains ARI $0.795$ and $0.710$ respectively), while agreement-focused signals can lead when cross-run consistency is most decisive (e.g., \textsc{Bc}, where Consensus/$C$ perform best in ARI/AMI).
However, these signals can fail in complementary ways: a point may be geometrically well-placed in each run but switch clusters across runs, or be consistently assigned yet weakly supported geometrically (Fig.~\ref{fig:failures}). CAKE reduces these failure modes by requiring both stable assignments and consistent geometric support under the learned cluster structure, yielding a more robust confidence signal than either component alone. CAKE outperforms both components on several synthetic datasets (S4--S6) and is also the top-performing filtering choice on most real datasets across ARI/AMI/ACC (e.g., CAKE$^{(\mathrm{HM})}$ on \textsc{Dg}, \textsc{Fm}, \textsc{Pd}, and \textsc{Lt}; CAKE$^{(\mathrm{PR})}$ on \textsc{Ng}). Overall, these trends further support the central claim that combining stability and geometry yields a more reliable confidence ranking than either component alone.

\medskip

\noindent 
CAKE produces an informative pointwise ranking.
Coverage--accuracy curves show that clustering accuracy improves as a function of retained coverage when points are ranked by CAKE scores (Fig.~\ref{fig:acc_cov_cake}), indicating that CAKE concentrates correctly clustered instances among the highest scores. This behavior is also visible in the percentile/decile analysis: accuracy rises nearly monotonically with CAKE score percentiles on both synthetic and real datasets (Appendix Fig.~\ref{app:fig:evaluation}). Overall, CAKE behaves as a reliability ordering from ambiguous/borderline points to stable core members.

\medskip

\noindent 
CAKE enables pointwise correctness prediction and effective error discovery. When predicting whether a consensus assignment is correct (after Hungarian alignment to ground truth), CAKE achieves strong discriminative performance and typically outperforms entropy-based agreement and bootstrap-based stability baselines in both AUPRC and AUROC (Table~\ref{tab:pointwise_all}). A similar picture emerges when the comparison is extended to native soft and density-based confidence signals (Table~\ref{tab:pointwise_fcm_densitybased}). $\mathrm{CAKE}^{(\mathrm{HM})}$ remains among the strongest overall methods, attaining the highest value on $4/7$ synthetic and $5/8$ real datasets across both metrics. The Fuzzy C-Means scores $\mathrm{FCM}^{(1)}$ and $\mathrm{FCM}^{(2)}$ are competitive, showing that soft memberships can already provide informative pointwise confidence when ambiguity is well captured by a single fuzzy partition. The density-based baselines are weaker overall. In particular, the ensemble cluster-local score $\mathrm{LOF}^{(2)}$ is consistently stronger than the global score $\mathrm{LOF}^{(1)}$, indicating that (ensemble-averaged) cluster-aware local density is more informative than dataset-level outlierness alone, but both remain below $\mathrm{CAKE}^{(\mathrm{HM})}$ in overall ranking quality. Related trends are also observed in the additional GMM-based comparison reported in Appendix~\ref{app:sec:gmm_confidence}, where $\mathrm{CAKE}^{(\mathrm{HM})}$ outperforms the native GMM-based confidence signals on most of the datasets considered (Appendix Table~\ref{tab:gmm_confidence}).
CAKE also yields a more monotonic reliability ordering than the ensemble-derived baselines: score percentiles correlate strongly with within-bin clustering accuracy (Fig.~\ref{fig:corr_scores_accuracy}).
In error discovery experiments, ranking by CAKE concentrates misclustered points more effectively than GMM posterior confidence $p_{\max}$ on most real datasets, while a simple fusion of the two signals is often competitive (Fig.~\ref{fig:error_discovery}).

\medskip

\noindent 
CAKE scores stabilize with moderate ensemble sizes. As the ensemble size $R$ increases, per-point score variability steadily decreases and the induced ranking becomes essentially stable, with diminishing returns beyond about $R \approx 30$--$40$ (Fig.~\ref{fig:convergence}). This indicates that a relatively small ensemble is sufficient to obtain a reliable confidence ordering. 

\medskip

\noindent 
Under misspecified cluster counts, CAKE's ranking quality also closely tracks clustering quality. Across datasets, Adjusted Rand Index (ARI) typically peaks at (or very near) the ground-truth number of clusters $k$, and the AURC is minimized at the same (or neighboring) $k$; equivalently, the $k$ values that yield the most accurate clustering also yield the sharpest CAKE ranking (Fig.~\ref{fig:kmisspec}).

\medskip

\noindent 
CAKE is model-agnostic. Across homogeneous \emph{k}-means, GMM, and heterogeneous \emph{k}-means \& GMM clustering ensembles (as well as spectral clustering ensembles for non-convex geometries; Fig.~\ref{fig:app:moons_cake}), CAKE score distributions are qualitatively similar and remain informative and well-shaped (Fig.~\ref{fig:ensembles-real-synth}), supporting that CAKE behaves consistently beyond purely \emph{k}-means ensembles. A more detailed analysis of ensemble diversity is reported in Appendix~\ref{app:sec:divens}, showing that CAKE remains informative across a broad range of homogeneous and heterogeneous ensemble constructions (Appendix Fig.~\ref{fig:ensabl}), while the effect of additional diversity depends on whether it contributes useful complementary variation or simply introduces less stable clustering behaviour.
As expected for any ensemble-derived confidence score, CAKE reflects the inductive biases of the base clustering algorithm, since it is intended to measure instance reliability with respect to that procedure and therefore remains comparable across different ensemble constructions.

\medskip

\noindent 
CAKE is computationally practical. Its runtime follows the expected scaling trends (\S\ref{subsec:cake}). It grows roughly linearly with the ensemble size $R$ and, when using the centroid-based Silhouette proxy, approximately linearly with the dataset size (Fig.~\ref{fig:runtime_cake}). The approximation closely matches the exact computation (high correlation and low MAE; Tables~\ref{tab:runtimer}--\ref{tab:runtimen}) while substantially reducing cost, making CAKE feasible for large datasets.

\section{Conclusion}
\noindent CAKE provides an interpretable, label-free per-point confidence score by fusing cross-run assignment stability with local geometric fit. It is simple to compute from a modest ensemble of standard clustering runs and yields a scalar in $[0,1]$ for every instance. More broadly, CAKE shows that ensemble diversity, long used for uncertainty estimation in supervised learning, transfers to unsupervised clustering: variability across runs provides a practical per-point confidence signal. 

\medskip

\noindent 
Promising directions include calibrated confidence (e.g., with a small labeled set), integration into semi- and self-supervised pipelines (e.g., uncertainty-aware pseudo-labeling/per-sample weighting), guiding the selection of $k$ via score distributions, and extending CAKE to multi-$k$ clustering ensembles where $k$ varies across runs. By making instance-level uncertainty explicit, CAKE provides an interpretable pointwise confidence score for clustering assignments, enabling more reliable ranking, prioritization, and downstream use of clustering in labeling, anomaly detection, and decision pipelines.

\section*{CRediT authorship statement}
\noindent \textbf{Aggelos Semoglou:} Conceptualization, Methodology, Software, Formal analysis, Writing - original draft.\\
\textbf{John Pavlopoulos:} Conceptualization, Methodology, Supervision, Writing - review \& editing. 

\section*{Declaration of competing interest}
\noindent The authors declare that they have no competing financial interests or personal relationships that could have appeared to influence the work reported in this paper.

\section*{Data availability}
\noindent Synthetic datasets and code are available at \url{https://github.com/semoglou/cake} and \url{https://pypi.org/project/cake-ensemble/}. The real-world datasets used in this study are publicly accessible via OpenML, \textsc{scikit-learn}, and TensorFlow.

\section*{Acknowledgments}
\noindent This work has been partially supported by project MIS 5154714 of the National Recovery and Resilience Plan Greece 2.0 funded by the European Union under the NextGenerationEU Program.

\printbibliography

\newpage
\appendix

\section{Appendix Theoretical Analysis}\label{app:analysis}
\normalsize
\subsection{Mis-ranking bound}\label{app:bound1}
\noindent Fix two points $x_i, x_j\in \Xp$ such that
$\theta_i \ge \theta_j + \gamma$ for some margin $\gamma > 0$.
Recall that the empirical score
$c_i$ (Eq.~\ref{eq:stability}) is an order-2 U-statistic whose (Bernoulli) kernel
$A_i^{(r_1,r_2)}\in[0,1]$ (Eq.~\ref{eq:bernoullikernel}) has mean $\mathbb{E}\left[A_i^{(r_1, r_2)}\right]= \theta_i$ (\S\ref{subsec:rankingerror}).
For each point, the estimation errors are defined as:
$e_i = c_i - \theta_i, \ \ e_j = c_j - \theta_j.$
Using Hoeffding's inequality for order-2 U-statistics with a kernel in $[0,1]$ for $e_i,e_j$: 
$$
\Pr\left(e_i \ge \epsilon\right) \le \exp\left( -\frac{R \epsilon^2}{2} \right), \quad
\Pr\left(e_i \le -\epsilon \right) \le \exp\left( -\frac{R \epsilon^2}{2} \right) \quad \forall \epsilon > 0 \text{ (same for } e_j).
$$
Choosing $\epsilon = \frac{\gamma}{2}$, the events are defined as:
$
E_1 = \left\{e_i \le - \frac{\gamma}{2}\right\}, \ E_2 = \left\{e_j\ge \frac{\gamma}{2}\right\}, 
$
with probabilities:
$$\Pr\left(E_1\right), \Pr\left( E_2 \right),  \text{ which are both bounded above by }\exp\left\{-R\left( \frac{(\gamma/2)^2}{2}\right) \right\} = \exp\left\{ -R \frac{\gamma^{2}}{8} \right\}.$$
Assume, for contradiction, that the event $\left\{c_i < c_j\right\}$ occurs while neither of $E_1$ nor $E_2$ takes place. \\Then $e_i > -\gamma/2$ and $e_j<\gamma/2$, so that:
$$
c_i = \theta_i + e_i \ge \theta_i - \frac{\gamma}{2} \ge \theta_j + \gamma - \frac{\gamma}{2} \ge \theta_j + \frac{\gamma}{2} \ge \theta_j + e_j = c_j \Rightarrow c_i \ge c_j, \text{ which contradicts } c_i < c_j.$$
Therefore, $\left\{c_i < c_j\right\} \subseteq E_1 \cup E_2$. 
Using the bounds on $E_1$ and $E_2$ obtained above:
\begin{equation}
\Pr\left[c_i<c_j\right] \le \Pr\left[E_1\right] + \Pr\left[E_2\right] \le 2\exp\left\{-R \cdot\frac{\gamma^2}{8}\right\}.
\end{equation}
Hence, the probability of a mis-ranking satisfies:
$
  \theta_i - \theta_j \ge \gamma
  \;\Rightarrow\;
  \Pr[c_i < c_j]
  \le 2\,\exp\!\left\{-R\frac{\gamma^{2}}{8}\right\}.
$
\subsection{False-stability bound}\label{app:bound2}
\noindent Consider a uniform-noise observation $x_i \in \Xp$, then $\text{for each } L^{(r)} \in \Lp, \ r \in \left\{1,\dots,R \right\}$, its label $L_i^{(r)} \in L^{(r)}$ is drawn independently and uniformly from $\left\{1,\dots,k \right\}$. For such a point the Bernoulli kernel in Eq.~\ref{eq:bernoullikernel} has expectation:
$$
\mathbb{E}\left[A_i^{(r_1, r_2)} \right] = \theta = \frac{1}{k}.
$$
Applying Hoeffding’s inequality to $c_i$ again gives:
$
\Pr\left[c_i - \mathbb{E}\left[A_i^{(r_1, r_2)} \right]\ge\epsilon\right] \le \exp{\left\{ -R \frac{\epsilon^2}{2} \right\} }.
$
Fix a stability threshold $\tau \in \left( \frac{1}{k},1\right)$ and select $\epsilon = (\tau - \theta)$. Then:
$$
\Pr\left[c_i> \tau \right] = \Pr\left[c_i - \theta > \tau - \theta \right] \le \exp{\left\{ -R \frac{(\tau-\theta)^2}{2} \right\} }.
$$
Since $x_i$ is a uniform-noise point: $\theta = 1/k$, and thus:
\begin{equation}
\Pr\left[c_i> \tau \ | \ x_i \text{ uniform-noise} \right] \le \exp{\left\{ -\frac{R}{2} \left(\tau- \frac{1}{k}\right)^2 \right\} .
}
\end{equation}
Additionally, if a fraction $\phi\in(0,1)$ of the $n$ points of the dataset are uniform-noise, linearity of expectation turns the single-point bound into a dataset bound immediately:
\begin{equation}
\mathbb{E}\left[ \#\left\{\ x_i \text{ uniform-noise, } c_i>\tau \right\} \right] \le n\cdot \phi \cdot \exp{\left\{ -\frac{R}{2} \left(\tau- \frac{1}{k}\right)^2 \right\} }.
\end{equation}
In more general scenarios, a noisy observation may be stochastically biased, due to proximity to dominant clusters or algorithmic bias, resulting in a non-uniform label distribution across runs. In such cases, the same concentration bound applies with this generalized $\theta \ge 1/k$:
\begin{equation}
\Pr\left[c_i> \tau \right]  \le \exp{\left\{ -R \frac{(\tau-\theta)^2}{2} \right\} }
\end{equation}
and extends the false-positive bound beyond uniform noise to any setting where labelings follow a known or estimated marginal distribution.

\medskip

\noindent \textbf{Geometric component.}
An analogous mis-ranking guarantee holds for the Silhouette-based geometric component.
For each point $x_i$ and run $r\in\{1,\dots,R\}$, let $s_i^{(r)}\in[-1,1]$ denote its (possibly approximated) Silhouette score in run $r$. 
The randomness here comes from the clustering procedure (e.g., different random initializations), so for fixed $i$, $s_i^{(1)},\dots,s_i^{(R)}$ are viewed as i.i.d.\ bounded random variables.
Let
\[
\bar{s}_i := \frac{1}{R}\sum_{r=1}^R s_i^{(r)}, 
\qquad
\hat{\sigma}_i^2 := \frac{1}{R}\sum_{r=1}^R (s_i^{(r)} - \bar{s}_i)^2,
\]
and recall that the ensemble geometric component used in CAKE is
\[
\tilde S_i := \max\bigl(0,\ \bar{s}_i - \hat{\sigma}_i\bigr)
\]
(up to the affine rescaling into $[0,1]$ described in \S\ref{subsec:silhouette}).
The map
\[
(s_i^{(1)},\dots,s_i^{(R)}) \mapsto \tilde S_i
\]
is a bounded-differences (Lipschitz) functional of the $s_i^{(r)}$’s, so by standard concentration inequalities for bounded independent variables (e.g., McDiarmid’s inequality) $\tilde S_i$ concentrates around its population counterpart $g_i := \mathbb{E}[\tilde S_i]$ at the usual $1/\sqrt{R}$ rate, with sub-Gaussian tails.
Repeating the mis-ranking argument of Appendix~\ref{app:bound1}, but with $c_i$ and $\theta_i$ replaced by $\tilde S_i$ and $g_i$, shows that whenever the true geometric scores of two points differ by a margin $\gamma>0$ (i.e., $g_i \ge g_j + \gamma$), the probability that CAKE mis-ranks them (i.e., $\tilde S_i < \tilde S_j$) decays exponentially in $R\gamma^2$.
A false-positive bound for $\tilde S_i$ can also be derived under additional assumptions on the distribution of Silhouette scores for noise points, again yielding exponentially small probabilities in $R$.

\section{Appendix Results}\label{app:sec:valid}
\medskip
\noindent\textbf{Clustering accuracy vs.\ CAKE percentiles (decile analysis).} To complement the coverage--accuracy curves in Fig.~\ref{fig:acc_cov_cake}, clustering accuracy is also examined across CAKE score quantiles. For each dataset, \emph{k}-means is applied at the ground-truth $k$, the predicted labels are aligned to ground truth via Hungarian matching, points are partitioned into CAKE deciles, and accuracy is computed within each decile. This yields an accuracy--percentile trend that summarizes how accuracy changes from the lowest-scored (most ambiguous) instances to the highest-scored (most reliable) instances, providing a decile-level view of CAKE’s ranking behavior over the full dataset (Figs.~\ref{app:fig:synthetic},\ref{app:fig:real}).
\vspace{3mm}
\begin{center}
{\captionsetup{type=figure}

  \begin{subfigure}[b]{0.48\linewidth}
    \centering
    \includegraphics[width=\linewidth]{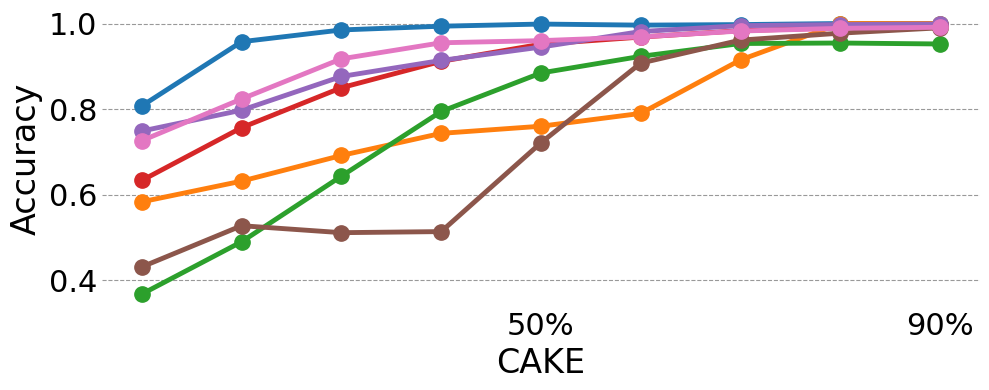}
    \caption{Synthetic datasets: \textcolor{MidnightBlue}{S1}, \textcolor{orange}{S2}, \textcolor{OliveGreen}{S3}, \textcolor{red}{S4}, \textcolor{Plum}{S5}, \textcolor{RawSienna}{S6}, \textcolor{RubineRed}{S7}.}
    \label{app:fig:synthetic}
  \end{subfigure}\hfill
  \begin{subfigure}[b]{0.48\linewidth}
    \centering
    \includegraphics[width=\linewidth]{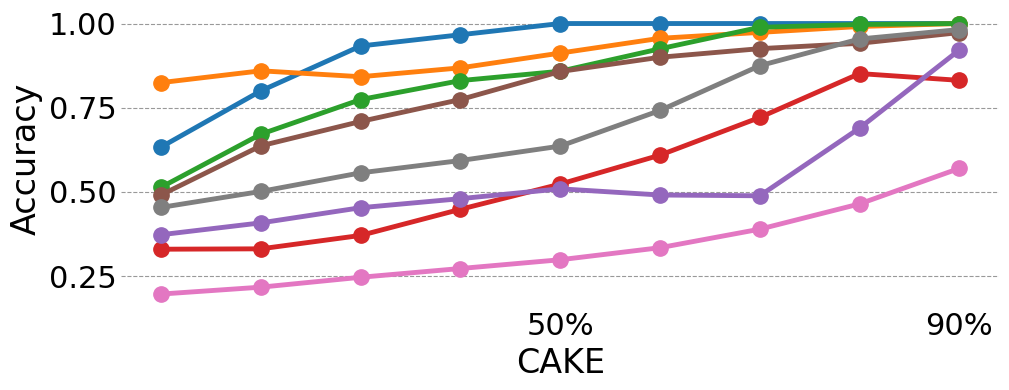}
    \caption{Real datasets: \textcolor{MidnightBlue}{\textsc{Ir}}, \textcolor{orange}{\textsc{Bc}}, \textcolor{OliveGreen}{\textsc{Dg}}, \textcolor{red}{\textsc{Ng}}, \textcolor{Plum}{\textsc{Fm}}, \textcolor{RawSienna}{\textsc{Pd}}, \textcolor{RubineRed}{\textsc{Lt}}, \textcolor{darkgray}{\textsc{Sa}}.}
    \label{app:fig:real}
  \end{subfigure}

  \caption{\textbf{Accuracy vs.\ CAKE score percentile.} Clustering accuracy within CAKE deciles (higher percentiles correspond to higher CAKE scores), after Hungarian alignment to ground-truth labels. This complements Fig.~\ref{fig:acc_cov_cake} by showing accuracy trends across score quantiles.}
  \label{app:fig:evaluation}
}
\end{center}

\onecolumn
\footnotesize

\begin{longtable}{llcc c}

\toprule
\textbf{Dataset} & \textbf{Subset} & \textbf{avg Silhouette} & \textbf{avg NMI} & \textbf{CORR } (CAKE $\sim $ ACC) \\
\midrule

\multirow{7}{*}{S1} & Full & 0.539 [0.5390, 0.5391] & 0.834 [0.8332, 0.8339]	 & \multirow{7}{*}{0.881} \\

& Random & 0.546 [0.5460, 0.5460]& 0.834 [0.8340, 0.8340] \\

& Consensus & 0.539 [0.5389, 0.5389] &0.839 [0.8395, 0.8395]  \\

& $C$ component & 0.539 [0.5389, 0.5389] & 0.839 [0.8395, 0.8395]  \\

& $\tilde S$ component & \textbf{0.692} [0.6921, 0.6921] & \textbf{0.983} [0.9828, 0.9828]	  \\

& $\mathrm{CAKE}^{\text{(PR)}}$ & \textbf{0.692} [0.6921, 0.6921] &  \textbf{0.983} [0.9828, 0.9828]	  \\

& $\mathrm{CAKE}^{\text{(HM)}}$ & \textbf{0.692} [0.6921, 0.6921] &  \textbf{0.983} [0.9828, 0.9828]	 \\

 \cmidrule(lr){2-5}

\multirow{7}{*}{S2} & Full &0.462 [0.4499, 0.4736] & 0.606 [0.6059, 0.6070]	 & \multirow{7}{*}{0.975} \\

& Random & 0.466 [0.4487, 0.4828] & 0.602 [0.5927, 0.6105]  \\

& Consensus & 0.459 [0.4499, 0.4673] & 0.628 [0.6034, 0.6533]\\

& $C$ component & 0.459 [0.4499, 0.4673] & 0.628 [0.6034, 0.6533] \\

& $\tilde S$ component & \textbf{0.582} [0.5254, 0.6394] & 0.698 [0.6792, 0.7165] \\

& $\mathrm{CAKE}^{\text{(PR)}}$ & \textbf{0.582} [0.5253, 0.6384]& \textbf{0.700} [0.6803, 0.7195] \\

& $\mathrm{CAKE}^{\text{(HM)}}$ & \textbf{0.582} [0.5252, 0.6392] & 0.699 [0.6799, 0.7185] \\

 \cmidrule(lr){2-5}

\multirow{7}{*}{S3} & Full & 0.537 [0.5270, 0.5472] & 0.614 [0.6101, 0.6181] & \multirow{7}{*}{0.975} \\

& Random & 0.503 [0.4694, 0.5357] & 0.581 [0.5357, 0.6259]  \\

& Consensus & 0.613 [0.5678, 0.6578] & 0.757 [0.7304, 0.7838] \\

& $C$ component & 0.599 [0.5657, 0.6327] & 0.721 [0.7044, 0.7382]	\\

& $\tilde S$ component & \textbf{0.639} [0.5904, 0.6867]	 & \textbf{0.781} [0.7593, 0.8024] \\

& $\mathrm{CAKE}^{\text{(PR)}}$ & 0.589 [0.5355, 0.6416] & 0.751 [0.6986, 0.8033]\\

& $\mathrm{CAKE}^{\text{(HM)}}$ & 0.613 [0.5858, 0.6407] & 0.777 [0.7600, 0.7934]\\

 \cmidrule(lr){2-5}

\multirow{7}{*}{S4} & Full & 0.434 [0.4339, 0.4339] & 0.678 [0.6778, 0.6790] & \multirow{7}{*}{0.999} \\

& Random & 0.441 [0.4410, 0.4410] & 0.693 [0.6931, 0.6938]  \\

& Consensus & 0.391 [0.3845, 0.3981] &0.599 [0.5946, 0.6036]	 \\

& $C$ component & 0.391 [0.3845, 0.3981] & 0.599 [0.5946, 0.6036]  \\

& $\tilde S$ component & \textbf{0.602} [0.6021, 0.6021] & \textbf{0.867} [0.8669, 0.8669]  \\

& $\mathrm{CAKE}^{\text{(PR)}}$ & \textbf{0.602} [0.6021, 0.6021] & \textbf{0.867} [0.8669, 0.8669]\\

& $\mathrm{CAKE}^{\text{(HM)}}$ & \textbf{0.602} [0.6021, 0.6021] & \textbf{0.867} [0.8669, 0.8669]\\

 \cmidrule(lr){2-5}

\multirow{7}{*}{S5} & Full & 0.537 [0.4467, 0.6267] & 0.660 [0.5741, 0.7463] & \multirow{7}{*}{0.996} \\

& Random & 0.512 [0.4115, 0.6127]	 & 0.634 [0.5419, 0.7268]  \\

& Consensus & 0.495 [0.4943, 0.4951] & 0.570 [0.5693, 0.5714]  \\

& $C$ component & 0.582 [0.5298, 0.6349]	 & 0.697 [0.6401, 0.7542] \\

& $\tilde S$ component & 0.724 [0.6001, 0.8486]	 & 0.855 [0.7559, 0.9536]  \\

& $\mathrm{CAKE}^{\text{(PR)}}$ & \textbf{0.805} [0.8049, 0.8049] & 0.879 [0.8793, 0.8793] \\

& $\mathrm{CAKE}^{\text{(HM)}}$ & \textbf{0.805} [0.8051, 0.8051] & \textbf{0.898} [0.8980, 0.8980]  \\

 \cmidrule(lr){2-5}

\multirow{7}{*}{S6} & Full & 0.511 [0.5106, 0.5116] & 0.474 [0.4722, 0.4750] & \multirow{7}{*}{0.963} \\

& Random & 0.512 [0.5075, 0.5157] & 0.481 [0.4693, 0.4924]	  \\

& Consensus & 0.669 [0.6571, 0.6807]	 & 0.586 [0.5783, 0.5946] \\

& $C$ component & 0.675 [0.6745, 0.6748]	 & 0.590 [0.5898, 0.5910]  \\

& $\tilde S$ component & 0.691 [0.6545, 0.7281]	 & 0.591 [0.5773, 0.6039]  \\

& $\mathrm{CAKE}^{\text{(PR)}}$ & 0.697 [0.6491, 0.7454] & 0.584 [0.5576, 0.6098]\\

& $\mathrm{CAKE}^{\text{(HM)}}$ & \textbf{0.717} [0.6936, 0.7396] & \textbf{0.597} [0.5896, 0.6037] \\

 \cmidrule(lr){2-5}

\multirow{7}{*}{S7} & Full & 0.467 [0.4112, 0.5232] & 0.649 [0.5645, 0.7334] & \multirow{7}{*}{0.984} \\

& Random & 0.466 [0.4091, 0.5222] & 0.636 [0.5467, 0.7255]  \\

& Consensus & 0.420 [0.4160, 0.4246] & 0.546 [0.5389, 0.5523] \\

& $C$ component & 0.419 [0.4156, 0.4231]	 & 0.541 [0.5330, 0.5484]	\\

& $\tilde S$ component & 0.581 [0.5068, 0.6557] & 0.762 [0.6531, 0.8714]  \\

& $\mathrm{CAKE}^{\text{(PR)}}$ & \textbf{0.643} [0.5941, 0.6913] & \textbf{0.861} [0.7910, 0.9309]\\

& $\mathrm{CAKE}^{\text{(HM)}}$ & 0.562 [0.4828, 0.6410] & 0.741 [0.6242, 0.8580]  \\

 \midrule
 
 \multirow{7}{*}{\textsc{Ir}} & Full & 0.548 [0.5409, 0.5559] & 0.731 [0.6947, 0.7679] & \multirow{7}{*}{0.884} \\
 
& Random & 0.527 [0.5200, 0.5335] & 0.703 [0.6592, 0.7470] \\

& Consensus & 0.537 [0.5191, 0.5550] &0.739 [0.7106, 0.7664]\\

& $C$ component & 0.537 [0.5191, 0.5550] & 0.739 [0.7106, 0.7664]  \\

& $\tilde S$ component & 0.637 [0.5766, 0.6971] & 0.873 [0.7716, 0.9745]  \\

& $\mathrm{CAKE}^{\text{(PR)}}$ & \textbf{0.655} [0.6133, 0.6964] & 0.868 [0.8154, 0.9204] \\

& $\mathrm{CAKE}^{\text{(HM)}}$ & 0.648 [0.5942, 0.7015] & \textbf{0.902} [0.8134, 0.9909] \\

\cmidrule(lr){2-5}
 
 \multirow{7}{*}{\textsc{Bc}} & Full & 0.357 [0.3560, 0.3574]	 & 0.544 [0.5349, 0.5524] & \multirow{7}{*}{0.963}\\
 
& Random & 0.360 [0.3599, 0.3599]	 & 0.535 [0.5346, 0.5346] \\

& Consensus & 0.390 [0.3897, 0.3901] &\textbf{0.680} [0.6779, 0.6822]\\

& $C$ component & 0.390 [0.3897, 0.3901] & \textbf{0.680} [0.6779, 0.6822]  \\

& $\tilde S$ component & \textbf{0.557} [0.5568, 0.5568] & 0.653 [0.6526, 0.6526]  \\

& $\mathrm{CAKE}^{\text{(PR)}}$ & \textbf{0.557} [0.5568, 0.5568] & 0.653 [0.6526, 0.6526] \\

& $\mathrm{CAKE}^{\text{(HM)}}$ & \textbf{0.557} [0.5568, 0.5568] & 0.653 [0.6526, 0.6526] \\

 \cmidrule(lr){2-5}

 \multirow{7}{*}{\textsc{Dg}} & Full & 0.178 [0.1711, 0.1857]	 & 0.734 [0.7233, 0.7447] & \multirow{7}{*}{0.996} \\
 
& Random & 0.181 [0.1737, 0.1892] & 0.717 [0.7053, 0.7280] \\

& Consensus &0.213 [0.1953, 0.2311]	 & 0.838 [0.8123, 0.8635]	 \\

& $C$ component & 0.208 [0.1904, 0.2254] & 0.837 [0.8140, 0.8591] \\

& $\tilde S$ component & 0.236 [0.2156, 0.2557] & 0.854 [0.8341, 0.8744]  \\

& $\mathrm{CAKE}^{\text{(PR)}}$ & 0.230 [0.2110, 0.2482] & 0.844 [0.8155, 0.8731]	 \\

& $\mathrm{CAKE}^{\text{(HM)}}$ & \textbf{0.237} [0.2181, 0.2558] & \textbf{0.868} [0.8475, 0.8891]\\

 \cmidrule(lr){2-5}
 
 \multirow{7}{*}{\textsc{Ng}} & Full & 0.076 [0.0699, 0.0812] & 0.511 [0.5016, 0.5194] & \multirow{7}{*}{0.960} \\
 
& Random & 0.082 [0.0800, 0.0834] & 0.518 [0.5103, 0.5254]  \\

& Consensus &0.105 [0.1021, 0.1072]	 & 0.588 [0.5838, 0.5914] \\

& $C$ component & 0.106 [0.1028, 0.1094] & 0.592 [0.5850, 0.5996]	 \\

& $\tilde S$ component & \textbf{0.108} [0.1047, 0.1118] & 0.611 [0.6017, 0.6201]  \\

& $\mathrm{CAKE}^{\text{(PR)}}$ & \textbf{0.108} [0.0992, 0.1167] & \textbf{0.619} [0.6136, 0.6252]  \\

& $\mathrm{CAKE}^{\text{(HM)}}$ & \textbf{0.108} [0.1047, 0.1118] & 0.611 [0.6017, 0.6201] \\

 \cmidrule(lr){2-5}
 
 \multirow{7}{*}{\textsc{Fm}} & Full & 0.142 [0.1373, 0.1476] & 0.507 [0.4967, 0.5164] & \multirow{7}{*}{0.890} \\
 
& Random & 0.140 [0.1341, 0.1458] & 0.510 [0.4996, 0.5195]  \\

& Consensus & 0.173 [0.1662, 0.1796]	 &0.560 [0.5535, 0.5668]	 \\

& $C$ component & 0.173 [0.1677, 0.1786] & 0.560 [0.5519, 0.5688]	\\

& $\tilde S$ component & 0.184 [0.1716, 0.1961] & 0.585 [0.5741, 0.5963]  \\

& $\mathrm{CAKE}^{\text{(PR)}}$ & 0.185 [0.1732, 0.1967] & 0.588 [0.5750, 0.6014] \\

& $\mathrm{CAKE}^{\text{(HM)}}$ & \textbf{0.192} [0.1812, 0.2023] & \textbf{0.595} [0.5793, 0.6113]\\

 \cmidrule(lr){2-5}
 
 \multirow{7}{*}{\textsc{Pd}} & Full & 0.303 [0.2904, 0.3161] & 0.676 [0.6673, 0.6851] & \multirow{7}{*}{0.999} \\
 
& Random & 0.291 [0.2822, 0.3005] & 0.683 [0.6733, 0.6925]  \\

& Consensus & 0.355 [0.3425, 0.3684] & 0.734 [0.7217, 0.7473] \\

& $C$ component & 0.358 [0.3427, 0.3728] & 0.722 [0.7085, 0.7351]\\

& $\tilde S$ component & \textbf{0.395} [0.3818, 0.4087] & \textbf{0.842} [0.8141, 0.8698]  \\

& $\mathrm{CAKE}^{\text{(PR)}}$ & 0.379 [0.3548, 0.4028]  & 0.820 [0.7985, 0.8410] \\

& $\mathrm{CAKE}^{\text{(HM)}}$ & 0.394 [0.3777, 0.4095] & \textbf{0.842} [0.8246, 0.8601] \\

 \cmidrule(lr){2-5}
 
 \multirow{7}{*}{\textsc{Lt}} & Full & 0.141 [0.1383, 0.1433] & 0.371 [0.3665, 0.3757] & \multirow{7}{*}{0.987} \\
 
& Random & 0.142 [0.1401, 0.1444] & 0.378 [0.3719, 0.3838]  \\

& Consensus & 0.185 [0.1816, 0.1880] & 0.434 [0.4304, 0.4373]	 \\

& $C$ component & 0.190 [0.1864, 0.1938]	 & 0.444 [0.4387, 0.4500] \\

& $\tilde S$ component & \textbf{0.193}  [0.1892, 0.1976] & 0.462 [0.4531, 0.4703]  \\

& $\mathrm{CAKE}^{\text{(PR)}}$ & 0.190 [0.1849, 0.1956]	 & 0.461 [0.4533, 0.4697]  \\

& $\mathrm{CAKE}^{\text{(HM)}}$ & 0.192 [0.1880, 0.1954] & \textbf{0.464} [0.4593, 0.4681]  \\

 \cmidrule(lr){2-5}
 
 \multirow{7}{*}{\textsc{Sa}} & Full & 0.349 [0.3378, 0.3605] & 0.596 [0.5561, 0.6350] & \multirow{7}{*}{0.999} \\
 
& Random & 0.348 [0.3350, 0.3602] & 0.609 [0.5901, 0.6279]  \\

& Consensus & 	0.336 [0.3107, 0.3617]	 & 0.575 [0.5556, 0.5952]	  \\

& $C$ component & 0.336 [0.3107, 0.3617]	 & 0.575 [0.5556, 0.5952]	 \\

& $\tilde S$ component & \textbf{0.483} [0.4688, 0.4972] & \textbf{0.765} [0.7588, 0.7707]  \\

& $\mathrm{CAKE}^{\text{(PR)}}$ & 0.468 [0.4393, 0.4959] & 0.749 [0.7272, 0.7708]  \\

& $\mathrm{CAKE}^{\text{(HM)}}$ & 0.443 [0.4104, 0.4765] & 0.735 [0.7092, 0.7598] \\

\bottomrule

\caption{Average Silhouette and NMI (Normalized Mutual Information) with $t$-95\% confidence intervals and Spearman's rank correlation of CAKE percentiles with clustering accuracy percentiles. Results are reported for the full datasets (Full) and their filtered subsets mentioned in \S\ref{subse:evalset}. Highest values in \textbf{bold}.}
\label{app_extrares}
\end{longtable}

\normalsize

\begin{center}
\centering
\normalsize
\setlength{\tabcolsep}{6pt}
\begin{tabular}{llcccc}
\toprule
Dataset & Metric & $\mathrm{CAKE}^{(\mathrm{PR})}$ & $\mathrm{CAKE}^{(\mathrm{HM})}$ & Entropy agreement & Bootstrap stability \\

\midrule
\multirow{2}{*}{S1} & AUPRC   & 0.992 & \textbf{0.993} & 0.976 & 0.967 \\
                   & AUROC   & 0.884 & \textbf{0.894} & 0.733 & 0.619 \\
\cmidrule(lr){2-6}
\multirow{2}{*}{S2} & AUPRC   & \textbf{0.881} & \textbf{0.881} & 0.801 & 0.872 \\
                   & AUROC   & \textbf{0.695} & 0.692 & 0.598 & 0.688 \\
\cmidrule(lr){2-6}
\multirow{2}{*}{S3} & AUPRC   & \textbf{0.943} & 0.940 & 0.915 & 0.868 \\
                   & AUROC   & \textbf{0.896} & 0.890 & 0.879 & 0.820 \\
\cmidrule(lr){2-6}
\multirow{2}{*}{S4} & AUPRC   & 0.973 & \textbf{0.974} & 0.948 & 0.929 \\
                   & AUROC   & 0.835 & \textbf{0.836} & 0.773 & 0.697 \\
\cmidrule(lr){2-6}
\multirow{2}{*}{S5} & AUPRC   & 0.982 & \textbf{0.983} & 0.917 & 0.909 \\
                   & AUROC   & 0.855 & \textbf{0.856} & 0.546 & 0.530 \\
\cmidrule(lr){2-6}
\multirow{2}{*}{S6} & AUPRC   & \textbf{0.919} & \textbf{0.919} & 0.812 & 0.766 \\
                   & AUROC   & \textbf{0.809} & 0.808 & 0.720 & 0.648 \\
\cmidrule(lr){2-6}
\multirow{2}{*}{S7} & AUPRC   & 0.975 & \textbf{0.977} & 0.950 & 0.944 \\
                   & AUROC   & \textbf{0.836} & 0.834 & 0.735 & 0.650 \\
\midrule

\multirow{2}{*}{\textsc{Ir}}           & AUPRC  & 0.991 & \textbf{0.992} & 0.962 & 0.980 \\
                               & AUROC  & 0.940 & \textbf{0.941} & 0.843 & 0.907 \\
\cmidrule(lr){2-6}

\multirow{2}{*}{\textsc{Bc}} & AUPRC  & 0.961 & \textbf{0.963} & 0.920 & 0.936 \\
                               & AUROC  & 0.733 & \textbf{0.743} & 0.584 & 0.670 \\
\cmidrule(lr){2-6}

\multirow{2}{*}{\textsc{Dg}}         & AUPRC  & \textbf{0.934} & 0.931 & 0.917 & 0.916 \\
                               & AUROC  & \textbf{0.815} & 0.809 & 0.785 & 0.771 \\
\cmidrule(lr){2-6}

\multirow{2}{*}{\textsc{Ng}}   & AUPRC  & 0.661 & 0.658 & \textbf{0.682} & 0.660 \\
                               & AUROC  & \textbf{0.695} & 0.693 & 0.677 & 0.656 \\
\cmidrule(lr){2-6}

\multirow{2}{*}{\textsc{Fm}}   & AUPRC  & 0.743 & \textbf{0.745} & 0.652 & 0.713 \\
                               & AUROC  & 0.676 & \textbf{0.679} & 0.643 & 0.648 \\
\cmidrule(lr){2-6}

\multirow{2}{*}{\textsc{Pd}}      & AUPRC  & \textbf{0.929} & \textbf{0.929} & 0.903 & 0.853 \\
                               & AUROC  & \textbf{0.844} & 0.841 & 0.823 & 0.769 \\
\cmidrule(lr){2-6}

\multirow{2}{*}{\textsc{Lt}}         & AUPRC  & \textbf{0.428} & 0.423 & 0.415 & 0.350 \\
                               & AUROC  & \textbf{0.659} & 0.656 & 0.652 & 0.606 \\
\cmidrule(lr){2-6}

\multirow{2}{*}{\textsc{Sa}}       & AUPRC  & 0.876 & \textbf{0.879} & 0.825 & 0.870 \\
                               & AUROC  & 0.779 & 0.783 & 0.716 & \textbf{0.792} \\
\bottomrule
\end{tabular}
\captionsetup{hypcap=false}
\captionof{table}{Per-point consensus correctness prediction (equivalent to Table~\ref{tab:pointwise_all}, which uses $k$-means), using \textbf{MiniBatchKMeans}. CAKE scores are computed from an $R{=}20$ MiniBatchKMeans ensemble; entropy-based agreement ($R{=}20$) and bootstrap-based stability ($B{=}20$) are both obtained using MiniBatchKMeans runs. Highest per row in \textbf{bold}.}
\label{tab:pointwise_all_minibatchkmeans}
\end{center}

\begin{center}
\centering
\normalsize
\setlength{\tabcolsep}{6pt}
\begin{tabular}{llcccc}
\toprule
Dataset & Metric & $\mathrm{CAKE}^{(\mathrm{PR})}$ & $\mathrm{CAKE}^{(\mathrm{HM})}$ & Entropy agreement & Bootstrap stability \\

\midrule
\multirow{2}{*}{S1} & AUPRC   & \textbf{0.997} & \textbf{0.997} & 0.964 & 0.965 \\
                   & AUROC   & \textbf{0.933} & \textbf{0.933} & 0.576 & 0.589 \\
\cmidrule(lr){2-6}
\multirow{2}{*}{S2} & AUPRC   & \textbf{0.928} & \textbf{0.928} & 0.888 & 0.794 \\
                   & AUROC   & 0.752 & \textbf{0.753} & 0.720 & 0.544 \\
\cmidrule(lr){2-6}
\multirow{2}{*}{S3} & AUPRC   & \textbf{0.936} & 0.935 & 0.848 & 0.817 \\
                   & AUROC   & \textbf{0.853} & \textbf{0.853} & 0.731 & 0.673 \\
\cmidrule(lr){2-6}
\multirow{2}{*}{S4} & AUPRC   & 0.970 & \textbf{0.971} & 0.906 & 0.928 \\
                   & AUROC   & 0.834 & \textbf{0.837} & 0.613 & 0.676 \\
\cmidrule(lr){2-6}
\multirow{2}{*}{S5} & AUPRC   & \textbf{0.979} & \textbf{0.979} & 0.917 & 0.908 \\
                   & AUROC   & 0.797 & \textbf{0.799} & 0.504 & 0.533 \\
\cmidrule(lr){2-6}
\multirow{2}{*}{S6} & AUPRC   & \textbf{0.912} & \textbf{0.912} & 0.799 & 0.794 \\
                   & AUROC   & \textbf{0.686} & 0.685 & 0.546 & 0.532 \\
\cmidrule(lr){2-6}
\multirow{2}{*}{S7} & AUPRC   & 0.975 & \textbf{0.977} & 0.939 & 0.908 \\
                   & AUROC   & 0.793 & \textbf{0.805} & 0.575 & 0.458 \\
\midrule

\multirow{2}{*}{\textsc{Ng}}   & AUPRC  & 0.316 & 0.316 & \textbf{0.384} & 0.335 \\
                               & AUROC  & 0.514 & 0.514 & \textbf{0.591} & 0.556 \\
\cmidrule(lr){2-6}

\multirow{2}{*}{\textsc{Bc}} & AUPRC  & \textbf{0.969} & 0.967 & 0.937 & 0.930 \\
                               & AUROC  & \textbf{0.726} & 0.718 & 0.626 & 0.601 \\
\cmidrule(lr){2-6}

\multirow{2}{*}{\textsc{Dg}}         & AUPRC  & 0.925 & 0.920 & \textbf{0.934} & 0.885 \\
                               & AUROC  & 0.807 & 0.798 & \textbf{0.832} & 0.741 \\
\cmidrule(lr){2-6}

\multirow{2}{*}{\textsc{Fm}}   & AUPRC  & \textbf{0.770} & 0.765 & 0.755 & 0.693 \\
                               & AUROC  & 0.737 & 0.733 & \textbf{0.764} & 0.702 \\
\cmidrule(lr){2-6}

\multirow{2}{*}{\textsc{Ir}}           & AUPRC  & 0.985 & \textbf{0.986} & 0.930 & 0.900 \\
                               & AUROC  & 0.895 & \textbf{0.899} & 0.570 & 0.466 \\
\cmidrule(lr){2-6}

\multirow{2}{*}{\textsc{Lt}}         & AUPRC  & 0.368 & 0.372 & \textbf{0.373} & 0.340 \\
                               & AUROC  & 0.604 & 0.605 & \textbf{0.629} & 0.575 \\
\cmidrule(lr){2-6}

\multirow{2}{*}{\textsc{Pd}}      & AUPRC  & 0.942 & \textbf{0.945} & 0.845 & 0.897 \\
                               & AUROC  & 0.862 & \textbf{0.866} & 0.725 & 0.764 \\
\cmidrule(lr){2-6}

\multirow{2}{*}{\textsc{Sa}}       & AUPRC  & \textbf{0.915} & 0.913 & 0.882 & 0.842 \\
                               & AUROC  & \textbf{0.815} & 0.811 & 0.760 & 0.680 \\
\bottomrule
\end{tabular}
\captionsetup{hypcap=false}
\captionof{table}{Per-point consensus correctness prediction (equivalent to Table~\ref{tab:pointwise_all}, which uses $k$-means), using \textbf{K-Medoids}. CAKE scores are computed from an $R{=}20$ K-Medoids ensemble; entropy-based agreement ($R{=}20$) and bootstrap-based stability ($B{=}20$) are both obtained using K-Medoids runs. Highest per row in \textbf{bold}.}
\label{tab:pointwise_all_kmedoids}
\end{center}

\begin{center}
\centering
\normalsize
\setlength{\tabcolsep}{6pt}
\begin{tabular}{llcccc}
\toprule
Dataset & Metric & $\mathrm{CAKE}^{(\mathrm{PR})}$ & $\mathrm{CAKE}^{(\mathrm{HM})}$ & Entropy agreement & Bootstrap stability \\

\midrule
\multirow{2}{*}{S1} & AUPRC   & \textbf{0.997} & \textbf{0.997} & 0.958 & 0.960 \\
                   & AUROC   & \textbf{0.932} & \textbf{0.932} & 0.503 & 0.525 \\
\cmidrule(lr){2-6}
\multirow{2}{*}{S2} & AUPRC   & \textbf{0.932} & \textbf{0.932} & 0.791 & 0.794 \\
                   & AUROC   & \textbf{0.766} & \textbf{0.766} & 0.507 & 0.515 \\
\cmidrule(lr){2-6}
\multirow{2}{*}{S3} & AUPRC   & 0.853 & 0.845 & \textbf{0.945} & 0.555 \\
                   & AUROC   & 0.756 & 0.747 & \textbf{0.956} & 0.446 \\
\cmidrule(lr){2-6}
\multirow{2}{*}{S4} & AUPRC   & \textbf{0.976} & \textbf{0.976} & 0.882 & 0.895 \\
                   & AUROC   & \textbf{0.852} & \textbf{0.852} & 0.510 & 0.570 \\
\cmidrule(lr){2-6}
\multirow{2}{*}{S5} & AUPRC   & \textbf{0.995} & \textbf{0.995} & 0.992 & 0.992 \\
                   & AUROC   & \textbf{0.568} & \textbf{0.568} & 0.500 & 0.531 \\
\cmidrule(lr){2-6}
\multirow{2}{*}{S6} & AUPRC   & \textbf{0.976} & \textbf{0.976} & 0.967 & 0.969 \\
                   & AUROC   & 0.506 & 0.506 & 0.503 & \textbf{0.529} \\
\cmidrule(lr){2-6}
\multirow{2}{*}{S7} & AUPRC   & \textbf{0.982} & \textbf{0.982} & 0.967 & 0.953 \\
                   & AUROC   & \textbf{0.675} & \textbf{0.675} & 0.542 & 0.437 \\
\midrule

\multirow{2}{*}{\textsc{Ir}}           & AUPRC  & \textbf{0.998} & \textbf{0.998} & 0.967 & 0.968 \\
                               & AUROC  & \textbf{0.949} & \textbf{0.949} & 0.500 & 0.537 \\
\cmidrule(lr){2-6}

\multirow{2}{*}{\textsc{Bc}} & AUPRC  & \textbf{0.909} & \textbf{0.909} & 0.692 & 0.727 \\
                               & AUROC  & \textbf{0.800} & \textbf{0.800} & 0.499 & 0.567 \\
\cmidrule(lr){2-6}

\multirow{2}{*}{\textsc{Dg}}         & AUPRC  & \textbf{0.947} & 0.945 & 0.943 & 0.926 \\
                               & AUROC  & 0.797 & 0.791 & \textbf{0.799} & 0.734 \\
\cmidrule(lr){2-6}

\multirow{2}{*}{\textsc{Ng}}   & AUPRC  & \textbf{0.691} & 0.684 & 0.685 & 0.651 \\
                               & AUROC  & \textbf{0.711} & 0.704 & 0.676 & 0.634 \\
\cmidrule(lr){2-6}

\multirow{2}{*}{\textsc{Fm}}   & AUPRC  & 0.684 & \textbf{0.688} & 0.572 & 0.613 \\
                               & AUROC  & 0.679 & \textbf{0.683} & 0.596 & 0.608 \\
\cmidrule(lr){2-6}

\multirow{2}{*}{\textsc{Pd}}      & AUPRC  & 0.898 & \textbf{0.899} & 0.850 & 0.891 \\
                               & AUROC  & \textbf{0.833} & 0.832 & \textbf{0.833} & \textbf{0.833} \\
\cmidrule(lr){2-6}

\multirow{2}{*}{\textsc{Lt}}         & AUPRC  & 0.457 & 0.452 & \textbf{0.490} & 0.399 \\
                               & AUROC  & 0.628 & 0.627 & \textbf{0.673} & 0.586 \\
\cmidrule(lr){2-6}

\multirow{2}{*}{\textsc{Sa}}       & AUPRC  & \textbf{0.912} & \textbf{0.912} & 0.596 & 0.584 \\
                               & AUROC  & \textbf{0.879} & \textbf{0.879} & 0.507 & 0.483 \\
\bottomrule
\end{tabular}
\captionsetup{hypcap=false}
\captionof{table}{Per-point consensus correctness prediction (equivalent to Table~\ref{tab:pointwise_all}, which uses $k$-means), using \textbf{GMM}. CAKE scores are computed from an $R{=}20$ GMM ensemble; entropy-based agreement ($R{=}20$) and bootstrap-based stability ($B{=}20$) are both obtained using GMM runs. The highest value for each dataset and metric is highlighted in \textbf{bold}.}
\label{tab:pointwise_all_gmm}
\end{center}

\newpage

\onecolumn
\footnotesize
\begin{longtable}{p{0.5cm}p{2.2cm}ccccc}
\caption{Average Silhouette, NMI, ARI, AMI, and ACC over multiple independent MiniBatchKMeans runs (different random seeds). Results are reported for the full dataset and for subsets selected by the different filtering strategies (random, consensus-agree, $C$ component, $\tilde S$ component, $\mathrm{CAKE}^{\text{(PR)}}$, $\mathrm{CAKE}^{\text{(HM)}}$). Best performance per dataset and metric is shown in \textbf{bold}.}\label{tab:results_minibatchkmeans}\\
\toprule
\multicolumn{2}{c}{} & \multicolumn{5}{c}{Clustering results on filtered datasets} \\
 \cmidrule(lr){3-7}
Data & Subset & avg Silhouette & avg NMI & avg ARI & avg AMI & avg ACC \\
\midrule
\endfirsthead

\toprule
\multicolumn{2}{c}{} & \multicolumn{5}{c}{Clustering results on filtered datasets} \\
 \cmidrule(lr){3-7}
Data& Subset & avg Silhouette & avg NMI & avg ARI & avg AMI & avg ACC \\
\midrule
\endhead

\endfoot

\bottomrule
\endlastfoot
\multirow{7}{*}{\textsc{S1}} & Full & 0.486 & 0.742 & 0.732 & 0.742 & 0.848 \\
& Random & 0.546 & 0.833 & 0.878 & 0.833 & 0.958 \\
& Consensus & 0.581 & 0.858 & 0.889 & 0.858 & 0.937 \\
& $C$ component & 0.590 & 0.864 & 0.910 & 0.864 & 0.959 \\
& $\tilde S$ component & 0.632 & 0.928 & 0.928 & 0.928 & 0.952 \\
& $\mathrm{CAKE}^{\text{(PR)}}$ & 0.636 & 0.950 & 0.973 & 0.950 & 0.990 \\
& $\mathrm{CAKE}^{\text{(HM)}}$ & \textbf{0.646} & \textbf{0.955} & \textbf{0.975} & \textbf{0.955} & \textbf{0.991} \\
\cmidrule(lr){2-7}
\multirow{7}{*}{\textsc{S2}} & Full & 0.470 & 0.600 & 0.499 & 0.600 & 0.666 \\
& Random & 0.448 & 0.593 & 0.537 & 0.592 & 0.753 \\
& Consensus & 0.553 & 0.652 & 0.657 & 0.652 & 0.718 \\
& $C$ component & 0.509 & 0.638 & 0.604 & 0.638 & 0.715 \\
& $\tilde S$ component & 0.562 & 0.677 & 0.608 & 0.676 & 0.723 \\
& $\mathrm{CAKE}^{\text{(PR)}}$ & \textbf{0.580} & \textbf{0.687} & \textbf{0.667} & \textbf{0.686} & \textbf{0.779} \\
& $\mathrm{CAKE}^{\text{(HM)}}$ & 0.578 & 0.681 & 0.656 & 0.681 & 0.775 \\
\cmidrule(lr){2-7}
\multirow{7}{*}{\textsc{S3}} & Full & 0.501 & 0.588 & 0.464 & 0.587 & 0.700 \\
& Random & 0.510 & 0.604 & 0.490 & 0.604 & 0.719 \\
& Consensus & 0.581 & 0.757 & 0.763 & 0.757 & 0.821 \\
& $C$ component & 0.593 & 0.770 & \textbf{0.782} & 0.769 & \textbf{0.836} \\
& $\tilde S$ component & 0.601 & 0.742 & 0.716 & 0.742 & 0.799 \\
& $\mathrm{CAKE}^{\text{(PR)}}$ & \textbf{0.628} & \textbf{0.771} & 0.774 & \textbf{0.771} & 0.828 \\
& $\mathrm{CAKE}^{\text{(HM)}}$ & 0.587 & 0.743 & 0.729 & 0.742 & 0.802 \\
\cmidrule(lr){2-7}
\multirow{7}{*}{\textsc{S4}} & Full & 0.420 & 0.644 & 0.664 & 0.644 & 0.856 \\
& Random & 0.435 & 0.674 & 0.702 & 0.674 & 0.876 \\
& Consensus & 0.457 & 0.728 & 0.699 & 0.727 & 0.789 \\
& $C$ component & \textbf{0.535} & \textbf{0.796} & \textbf{0.812} & \textbf{0.795} & \textbf{0.900} \\
& $\tilde S$ component & 0.516 & 0.785 & 0.779 & 0.785 & 0.859 \\
& $\mathrm{CAKE}^{\text{(PR)}}$ & 0.511 & 0.777 & 0.775 & 0.776 & 0.854 \\
& $\mathrm{CAKE}^{\text{(HM)}}$ & 0.496 & 0.765 & 0.749 & 0.765 & 0.840 \\
\cmidrule(lr){2-7}
\multirow{7}{*}{\textsc{S5}} & Full & 0.504 & 0.615 & 0.532 & 0.615 & 0.746 \\
& Random & 0.536 & 0.651 & 0.579 & 0.651 & 0.774 \\
& Consensus & 0.614 & 0.689 & 0.655 & 0.689 & 0.836 \\
& $C$ component & 0.620 & 0.710 & 0.681 & 0.710 & 0.846 \\
& $\tilde S$ component & 0.745 & 0.856 & 0.839 & 0.855 & 0.892 \\
& $\mathrm{CAKE}^{\text{(PR)}}$ & \textbf{0.815} & \textbf{0.923} & \textbf{0.948} & \textbf{0.923} & \textbf{0.981} \\
& $\mathrm{CAKE}^{\text{(HM)}}$ & 0.779 & 0.892 & 0.896 & 0.892 & 0.938 \\
\cmidrule(lr){2-7}
\multirow{7}{*}{\textsc{S6}} & Full & 0.508 & 0.481 & 0.316 & 0.481 & 0.687 \\
& Random & 0.511 & 0.486 & 0.325 & 0.485 & 0.690 \\
& Consensus & 0.647 & 0.585 & 0.566 & 0.584 & 0.781 \\
& $C$ component & 0.654 & \textbf{0.590} & \textbf{0.570} & \textbf{0.590} & \textbf{0.786} \\
& $\tilde S$ component & \textbf{0.663} & 0.567 & 0.541 & 0.567 & 0.757 \\
& $\mathrm{CAKE}^{\text{(PR)}}$ & 0.658 & 0.580 & 0.555 & 0.580 & 0.771 \\
& $\mathrm{CAKE}^{\text{(HM)}}$ & 0.640 & 0.569 & 0.546 & 0.568 & 0.766 \\
\cmidrule(lr){2-7}
\multirow{7}{*}{\textsc{S7}} & Full & 0.419 & 0.571 & 0.502 & 0.571 & 0.729 \\
& Random & 0.384 & 0.532 & 0.434 & 0.532 & 0.642 \\
& Consensus & 0.430 & 0.587 & 0.506 & 0.587 & 0.703 \\
& $C$ component & 0.446 & 0.613 & 0.550 & 0.613 & 0.720 \\
& $\tilde S$ component & 0.577 & 0.787 & 0.734 & 0.787 & 0.810 \\
& $\mathrm{CAKE}^{\text{(PR)}}$ & \textbf{0.579} & \textbf{0.801} & \textbf{0.744} & \textbf{0.801} & \textbf{0.820} \\
& $\mathrm{CAKE}^{\text{(HM)}}$ & 0.553 & 0.773 & 0.699 & 0.773 & 0.778 \\
\midrule
\multirow{7}{*}{\textsc{Ir}} & Full & 0.533 & 0.730 & 0.682 & 0.727 & 0.825 \\
& Random & 0.526 & 0.715 & 0.658 & 0.710 & 0.844 \\
& Consensus & 0.628 & 0.846 & 0.829 & 0.843 & 0.886 \\
& $C$ component & 0.643 & \textbf{0.905} & \textbf{0.894} & \textbf{0.903} & \textbf{0.942} \\
& $\tilde S$ component & \textbf{0.664} & 0.903 & 0.879 & 0.901 & 0.910 \\
& $\mathrm{CAKE}^{\text{(PR)}}$ & \textbf{0.664} & 0.903 & 0.879 & 0.901 & 0.910 \\
& $\mathrm{CAKE}^{\text{(HM)}}$ & \textbf{0.664} & 0.903 & 0.879 & 0.901 & 0.910 \\
\cmidrule(lr){2-7}
\multirow{7}{*}{\textsc{Bc}} & Full & 0.356 & 0.537 & 0.654 & 0.536 & 0.905 \\
& Random & 0.357 & 0.532 & 0.656 & 0.531 & 0.906 \\
& Consensus & 0.445 & 0.590 & 0.707 & 0.589 & 0.922 \\
& $C$ component & 0.445 & 0.590 & 0.707 & 0.589 & 0.922 \\
& $\tilde S$ component & 0.496 & 0.556 & 0.654 & 0.555 & 0.916 \\
& $\mathrm{CAKE}^{\text{(PR)}}$ & 0.533 & \textbf{0.608} & \textbf{0.715} & \textbf{0.607} & \textbf{0.937} \\
& $\mathrm{CAKE}^{\text{(HM)}}$ & \textbf{0.534} & 0.605 & 0.707 & 0.604 & 0.936 \\
\cmidrule(lr){2-7}
\multirow{7}{*}{\textsc{Dg}} & Full & 0.152 & 0.667 & 0.536 & 0.663 & 0.668 \\
& Random & 0.169 & 0.692 & 0.566 & 0.688 & 0.691 \\
& Consensus & 0.199 & 0.800 & 0.716 & 0.797 & 0.757 \\
& $C$ component & 0.203 & 0.819 & 0.748 & 0.816 & 0.781 \\
& $\tilde S$ component & 0.206 & 0.816 & 0.703 & 0.813 & 0.731 \\
& $\mathrm{CAKE}^{\text{(PR)}}$ & \textbf{0.223} & \textbf{0.849} & \textbf{0.767} & \textbf{0.847} & \textbf{0.807} \\
& $\mathrm{CAKE}^{\text{(HM)}}$ & 0.219 & 0.830 & 0.728 & 0.828 & 0.763 \\
\cmidrule(lr){2-7}
\multirow{7}{*}{\textsc{Ng}} & Full & 0.068 & 0.490 & 0.327 & 0.488 & 0.478 \\
& Random & 0.069 & 0.480 & 0.310 & 0.477 & 0.454 \\
& Consensus & 0.108 & 0.613 & 0.488 & 0.611 & 0.575 \\
& $C$ component & 0.113 & 0.614 & 0.480 & 0.612 & 0.565 \\
& $\tilde S$ component & \textbf{0.113} & \textbf{0.650} & \textbf{0.518} & \textbf{0.648} & \textbf{0.601} \\
& $\mathrm{CAKE}^{\text{(PR)}}$ & \textbf{0.113} & \textbf{0.650} & \textbf{0.518} & \textbf{0.648} & \textbf{0.601} \\
& $\mathrm{CAKE}^{\text{(HM)}}$ & \textbf{0.113} & \textbf{0.650} & \textbf{0.518} & \textbf{0.648} & \textbf{0.601} \\
\cmidrule(lr){2-7}
\multirow{7}{*}{\textsc{Fm}} & Full & 0.171 & 0.505 & 0.353 & 0.505 & 0.515 \\
& Random & 0.173 & 0.499 & 0.346 & 0.498 & 0.510 \\
& Consensus & 0.212 & 0.551 & 0.414 & 0.550 & 0.540 \\
& $C$ component & 0.226 & 0.528 & 0.394 & 0.527 & 0.510 \\
& $\tilde S$ component & 0.226 & 0.573 & 0.427 & 0.573 & 0.561 \\
& $\mathrm{CAKE}^{\text{(PR)}}$ & \textbf{0.227} & 0.578 & 0.440 & 0.578 & 0.557 \\
& $\mathrm{CAKE}^{\text{(HM)}}$ & 0.226 & \textbf{0.588} & \textbf{0.456} & \textbf{0.588} & \textbf{0.581} \\
\cmidrule(lr){2-7}
\multirow{7}{*}{\textsc{Pd}} & Full & 0.258 & 0.633 & 0.488 & 0.632 & 0.635 \\
& Random & 0.267 & 0.641 & 0.501 & 0.641 & 0.650 \\
& Consensus & 0.360 & 0.741 & 0.630 & 0.740 & 0.704 \\
& $C$ component & 0.363 & 0.762 & 0.666 & 0.761 & 0.733 \\
& $\tilde S$ component & 0.360 & 0.782 & 0.632 & 0.782 & 0.703 \\
& $\mathrm{CAKE}^{\text{(PR)}}$ & \textbf{0.377} & \textbf{0.810} & \textbf{0.718} & \textbf{0.810} & \textbf{0.777} \\
& $\mathrm{CAKE}^{\text{(HM)}}$ & 0.356 & 0.787 & 0.667 & 0.787 & 0.717 \\
\cmidrule(lr){2-7}
\multirow{7}{*}{\textsc{Lt}} & Full & 0.128 & 0.356 & 0.140 & 0.353 & 0.262 \\
& Random & 0.122 & 0.360 & 0.141 & 0.356 & 0.268 \\
& Consensus & 0.165 & 0.425 & 0.187 & 0.421 & 0.311 \\
& $C$ component & \textbf{0.172} & 0.430 & 0.191 & 0.426 & 0.310 \\
& $\tilde S$ component & 0.170 & 0.440 & 0.202 & 0.436 & 0.317 \\
& $\mathrm{CAKE}^{\text{(PR)}}$ & 0.169 & \textbf{0.445} & \textbf{0.209} & \textbf{0.442} & \textbf{0.331} \\
& $\mathrm{CAKE}^{\text{(HM)}}$ & 0.168 & 0.439 & 0.196 & 0.435 & 0.313 \\
\cmidrule(lr){2-7}
\multirow{7}{*}{\textsc{Sa}} & Full & 0.326 & 0.571 & 0.482 & 0.570 & 0.641 \\
& Random & 0.313 & 0.561 & 0.456 & 0.560 & 0.640 \\
& Consensus & 0.363 & 0.641 & 0.559 & 0.640 & 0.692 \\
& $C$ component & 0.366 & 0.647 & 0.586 & 0.647 & 0.718 \\
& $\tilde S$ component & 0.437 & 0.716 & 0.637 & 0.715 & 0.741 \\
& $\mathrm{CAKE}^{\text{(PR)}}$ & \textbf{0.442} & \textbf{0.720} & \textbf{0.651} & \textbf{0.720} & 0.742 \\
& $\mathrm{CAKE}^{\text{(HM)}}$ & 0.425 & 0.715 & 0.650 & 0.714 & \textbf{0.749} \\
\end{longtable}
\twocolumn

\newpage

\onecolumn
\footnotesize
\begin{longtable}{p{0.5cm}p{2.2cm}ccccc}
\caption{Average Silhouette, NMI, ARI, AMI, and ACC over multiple independent GMM runs (different random seeds). Results are reported for the full dataset and for subsets selected by the different filtering strategies (random, consensus-agree, $C$ component, $\tilde S$ component, $\mathrm{CAKE}^{\text{(PR)}}$, $\mathrm{CAKE}^{\text{(HM)}}$). Best performance per dataset and metric is shown in \textbf{bold}.}\label{tab:results_gmm}\\
\toprule
\multicolumn{2}{c}{} & \multicolumn{5}{c}{Clustering results on filtered datasets} \\
 \cmidrule(lr){3-7}
Data & Subset & avg Silhouette & avg NMI & avg ARI & avg AMI & avg ACC \\
\midrule
\endfirsthead

\toprule
\multicolumn{2}{c}{} & \multicolumn{5}{c}{Clustering results on filtered datasets} \\
 \cmidrule(lr){3-7}
Data & Subset & avg Silhouette & avg NMI & avg ARI & avg AMI & avg ACC \\
\midrule
\endhead

\endfoot

\bottomrule
\endlastfoot
\multirow{7}{*}{\textsc{S1}} & Full & 0.539 & 0.834 & 0.879 & 0.834 & 0.958 \\
& Random & 0.546 & 0.833 & 0.878 & 0.833 & 0.957 \\
& Consensus & 0.538 & 0.836 & 0.881 & 0.836 & 0.959 \\
& $C$ component & 0.538 & 0.836 & 0.881 & 0.836 & 0.959 \\
& $\tilde S$ component & \textbf{0.692} & \textbf{0.983} & \textbf{0.992} & \textbf{0.983} & \textbf{0.997} \\
& $\mathrm{CAKE}^{\text{(PR)}}$ & \textbf{0.692} & \textbf{0.983} & \textbf{0.992} & \textbf{0.983} & \textbf{0.997} \\
& $\mathrm{CAKE}^{\text{(HM)}}$ & \textbf{0.692} & \textbf{0.983} & \textbf{0.992} & \textbf{0.983} & \textbf{0.997} \\
\cmidrule(lr){2-7}
\multirow{7}{*}{\textsc{S2}} & Full & 0.457 & 0.612 & 0.566 & 0.612 & 0.789 \\
& Random & 0.461 & 0.615 & 0.562 & 0.614 & 0.782 \\
& Consensus & 0.489 & 0.701 & 0.694 & 0.700 & 0.770 \\
& $C$ component & 0.489 & 0.701 & 0.694 & 0.700 & 0.770 \\
& $\tilde S$ component & 0.613 & 0.712 & 0.739 & 0.712 & 0.831 \\
& $\mathrm{CAKE}^{\text{(PR)}}$ & \textbf{0.615} & \textbf{0.716} & \textbf{0.743} & \textbf{0.716} & \textbf{0.833} \\
& $\mathrm{CAKE}^{\text{(HM)}}$ & \textbf{0.615} & \textbf{0.716} & \textbf{0.743} & \textbf{0.716} & \textbf{0.833} \\
\cmidrule(lr){2-7}
\multirow{7}{*}{\textsc{S3}} & Full & 0.510 & 0.617 & 0.500 & 0.617 & 0.720 \\
& Random & 0.474 & 0.616 & 0.502 & 0.616 & 0.717 \\
& Consensus & 0.673 & 0.840 & 0.864 & 0.839 & 0.932 \\
& $C$ component & 0.673 & 0.840 & 0.864 & 0.839 & 0.932 \\
& $\tilde S$ component & \textbf{0.729} & 0.871 & 0.906 & 0.871 & \textbf{0.949} \\
& $\mathrm{CAKE}^{\text{(PR)}}$ & 0.723 & 0.833 & 0.857 & 0.833 & 0.905 \\
& $\mathrm{CAKE}^{\text{(HM)}}$ & \textbf{0.729} & \textbf{0.872} & \textbf{0.907} & \textbf{0.872} & \textbf{0.949} \\
\cmidrule(lr){2-7}
\multirow{7}{*}{\textsc{S4}} & Full & 0.434 & 0.675 & 0.709 & 0.675 & 0.880 \\
& Random & 0.441 & 0.691 & 0.724 & 0.691 & 0.887 \\
& Consensus & 0.379 & 0.619 & 0.649 & 0.618 & 0.806 \\
& $C$ component & 0.379 & 0.619 & 0.649 & 0.618 & 0.806 \\
& $\tilde S$ component & \textbf{0.602} & \textbf{0.866} & \textbf{0.901} & \textbf{0.866} & \textbf{0.962} \\
& $\mathrm{CAKE}^{\text{(PR)}}$ & \textbf{0.602} & \textbf{0.866} & \textbf{0.901} & \textbf{0.866} & \textbf{0.962} \\
& $\mathrm{CAKE}^{\text{(HM)}}$ & \textbf{0.602} & \textbf{0.866} & \textbf{0.901} & \textbf{0.866} & \textbf{0.962} \\
\cmidrule(lr){2-7}
\multirow{7}{*}{\textsc{S5}} & Full & 0.573 & 0.954 & 0.973 & 0.954 & \textbf{0.992} \\
& Random & 0.577 & 0.954 & 0.972 & 0.954 & 0.991 \\
& Consensus & 0.457 & 0.682 & 0.575 & 0.681 & 0.725 \\
& $C$ component & 0.457 & 0.682 & 0.575 & 0.681 & 0.725 \\
& $\tilde S$ component & \textbf{0.827} & \textbf{0.962} & \textbf{0.976} & \textbf{0.962} & 0.991 \\
& $\mathrm{CAKE}^{\text{(PR)}}$ & \textbf{0.827} & \textbf{0.962} & \textbf{0.976} & \textbf{0.962} & 0.991 \\
& $\mathrm{CAKE}^{\text{(HM)}}$ & \textbf{0.827} & \textbf{0.962} & \textbf{0.976} & \textbf{0.962} & 0.991 \\
\cmidrule(lr){2-7}
\multirow{7}{*}{\textsc{S6}} & Full & 0.344 & 0.852 & 0.894 & 0.852 & 0.966 \\
& Random & 0.349 & 0.855 & \textbf{0.896} & 0.855 & \textbf{0.967} \\
& Consensus & 0.424 & 0.628 & 0.546 & 0.628 & 0.679 \\
& $C$ component & 0.424 & 0.628 & 0.546 & 0.628 & 0.679 \\
& $\tilde S$ component & \textbf{0.725} & \textbf{0.863} & 0.895 & \textbf{0.863} & 0.960 \\
& $\mathrm{CAKE}^{\text{(PR)}}$ & \textbf{0.725} & \textbf{0.863} & 0.895 & \textbf{0.863} & 0.960 \\
& $\mathrm{CAKE}^{\text{(HM)}}$ & \textbf{0.725} & \textbf{0.863} & 0.895 & \textbf{0.863} & 0.960 \\
\cmidrule(lr){2-7}
\multirow{7}{*}{\textsc{S7}} & Full & 0.476 & 0.759 & 0.754 & 0.759 & 0.879 \\
& Random & 0.510 & 0.842 & 0.881 & 0.842 & 0.965 \\
& Consensus & 0.441 & 0.601 & 0.508 & 0.601 & 0.710 \\
& $C$ component & 0.439 & 0.601 & 0.508 & 0.601 & 0.703 \\
& $\tilde S$ component & 0.614 & 0.832 & 0.792 & 0.832 & 0.862 \\
& $\mathrm{CAKE}^{\text{(PR)}}$ & \textbf{0.693} & \textbf{0.921} & \textbf{0.934} & \textbf{0.921} & \textbf{0.980} \\
& $\mathrm{CAKE}^{\text{(HM)}}$ & 0.615 & 0.826 & 0.786 & 0.826 & 0.858 \\
\midrule
\multirow{7}{*}{\textsc{Ir}} & Full & 0.553 & 0.789 & 0.750 & 0.786 & 0.903 \\
& Random & 0.516 & 0.756 & 0.722 & 0.752 & 0.894 \\
& Consensus & 0.452 & 0.622 & 0.544 & 0.613 & 0.855 \\
& $C$ component & 0.452 & 0.622 & 0.544 & 0.613 & 0.855 \\
& $\tilde S$ component & \textbf{0.720} & \textbf{0.961} & \textbf{0.979} & \textbf{0.960} & \textbf{0.990} \\
& $\mathrm{CAKE}^{\text{(PR)}}$ & \textbf{0.720} & \textbf{0.961} & \textbf{0.979} & \textbf{0.960} & \textbf{0.990} \\
& $\mathrm{CAKE}^{\text{(HM)}}$ & \textbf{0.720} & \textbf{0.961} & \textbf{0.979} & \textbf{0.960} & \textbf{0.990} \\
\cmidrule(lr){2-7}
\multirow{7}{*}{\textsc{Bc}} & Full & 0.277 & 0.031 & 0.069 & 0.030 & 0.649 \\
& Random & 0.241 & 0.033 & 0.074 & 0.031 & 0.651 \\
& Consensus & 0.291 & 0.302 & 0.404 & 0.301 & 0.819 \\
& $C$ component & 0.291 & 0.302 & 0.404 & 0.301 & 0.819 \\
& $\tilde S$ component & \textbf{0.292} & \textbf{0.543} & \textbf{0.690} & \textbf{0.542} & \textbf{0.920} \\
& $\mathrm{CAKE}^{\text{(PR)}}$ & \textbf{0.292} & \textbf{0.543} & \textbf{0.690} & \textbf{0.542} & \textbf{0.920} \\
& $\mathrm{CAKE}^{\text{(HM)}}$ & \textbf{0.292} & \textbf{0.543} & \textbf{0.690} & \textbf{0.542} & \textbf{0.920} \\
\cmidrule(lr){2-7}
\multirow{7}{*}{\textsc{Dg}} & Full & 0.128 & 0.619 & 0.498 & 0.615 & 0.658 \\
& Random & 0.132 & 0.612 & 0.464 & 0.604 & 0.645 \\
& Consensus & 0.196 & 0.808 & 0.758 & 0.804 & \textbf{0.779} \\
& $C$ component & 0.170 & 0.794 & 0.731 & 0.789 & 0.735 \\
& $\tilde S$ component & \textbf{0.222} & \textbf{0.834} & \textbf{0.775} & \textbf{0.831} & 0.770 \\
& $\mathrm{CAKE}^{\text{(PR)}}$ & \textbf{0.222} & \textbf{0.834} & \textbf{0.775} & \textbf{0.831} & 0.770 \\
& $\mathrm{CAKE}^{\text{(HM)}}$ & \textbf{0.222} & \textbf{0.834} & \textbf{0.775} & \textbf{0.831} & 0.770 \\
\cmidrule(lr){2-7}
\multirow{7}{*}{\textsc{Ng}} & Full & 0.071 & 0.505 & 0.301 & 0.503 & 0.503 \\
& Random & 0.071 & 0.505 & 0.285 & 0.503 & 0.489 \\
& Consensus & 0.109 & 0.576 & 0.430 & 0.574 & 0.564 \\
& $C$ component & 0.112 & 0.580 & 0.438 & 0.578 & 0.567 \\
& $\tilde S$ component & \textbf{0.122} & \textbf{0.654} & \textbf{0.514} & \textbf{0.652} & \textbf{0.602} \\
& $\mathrm{CAKE}^{\text{(PR)}}$ & \textbf{0.122} & \textbf{0.654} & \textbf{0.514} & \textbf{0.652} & \textbf{0.602} \\
& $\mathrm{CAKE}^{\text{(HM)}}$ & \textbf{0.122} & \textbf{0.654} & \textbf{0.514} & \textbf{0.652} & \textbf{0.602} \\
\cmidrule(lr){2-7}
\multirow{7}{*}{\textsc{Fm}} & Full & 0.124 & 0.521 & 0.332 & 0.521 & 0.510 \\
& Random & 0.119 & 0.516 & 0.325 & 0.516 & 0.492 \\
& Consensus & 0.222 & 0.602 & 0.442 & 0.602 & 0.590 \\
& $C$ component & 0.230 & 0.600 & 0.464 & 0.600 & 0.608 \\
& $\tilde S$ component & \textbf{0.258} & \textbf{0.632} & \textbf{0.467} & \textbf{0.631} & \textbf{0.611} \\
& $\mathrm{CAKE}^{\text{(PR)}}$ & \textbf{0.258} & \textbf{0.632} & \textbf{0.467} & \textbf{0.631} & \textbf{0.611} \\
& $\mathrm{CAKE}^{\text{(HM)}}$ & \textbf{0.258} & \textbf{0.632} & \textbf{0.467} & \textbf{0.631} & \textbf{0.611} \\
\cmidrule(lr){2-7}
\multirow{7}{*}{\textsc{Pd}} & Full & 0.178 & 0.585 & 0.425 & 0.584 & 0.586 \\
& Random & 0.184 & 0.610 & 0.456 & 0.609 & 0.621 \\
& Consensus & 0.306 & 0.745 & 0.665 & 0.744 & 0.713 \\
& $C$ component & 0.322 & 0.753 & 0.672 & 0.752 & 0.715 \\
& $\tilde S$ component & \textbf{0.382} & \textbf{0.814} & \textbf{0.704} & \textbf{0.813} & \textbf{0.772} \\
& $\mathrm{CAKE}^{\text{(PR)}}$ & \textbf{0.382} & \textbf{0.814} & \textbf{0.704} & \textbf{0.813} & \textbf{0.772} \\
& $\mathrm{CAKE}^{\text{(HM)}}$ & \textbf{0.382} & \textbf{0.814} & \textbf{0.704} & \textbf{0.813} & \textbf{0.772} \\
\cmidrule(lr){2-7}
\multirow{7}{*}{\textsc{Lt}} & Full & 0.036 & 0.341 & 0.112 & 0.338 & 0.241 \\
& Random & 0.039 & 0.356 & 0.121 & 0.341 & 0.252 \\
& Consensus & 0.251 & 0.570 & 0.357 & 0.560 & 0.426 \\
& $C$ component & 0.261 & 0.600 & 0.382 & 0.590 & 0.450 \\
& $\tilde S$ component & 0.263 & 0.600 & 0.384 & 0.591 & 0.452 \\
& $\mathrm{CAKE}^{\text{(PR)}}$ & \textbf{0.281} & \textbf{0.601} & \textbf{0.391} & \textbf{0.592} & \textbf{0.460} \\
& $\mathrm{CAKE}^{\text{(HM)}}$ & \textbf{0.281} & \textbf{0.601} & \textbf{0.391} & \textbf{0.592} & \textbf{0.460} \\
\cmidrule(lr){2-7}
\multirow{7}{*}{\textsc{Sa}} & Full & 0.171 & 0.575 & 0.530 & 0.574 & 0.660 \\
& Random & 0.170 & 0.574 & 0.496 & 0.573 & 0.637 \\
& Consensus & 0.169 & 0.565 & 0.514 & 0.565 & 0.655 \\
& $C$ component & 0.169 & 0.565 & 0.514 & 0.565 & 0.655 \\
& $\tilde S$ component & \textbf{0.409} & \textbf{0.754} & \textbf{0.748} & \textbf{0.754} & \textbf{0.821} \\
& $\mathrm{CAKE}^{\text{(PR)}}$ & \textbf{0.409} & \textbf{0.754} & \textbf{0.748} & \textbf{0.754} & \textbf{0.821} \\
& $\mathrm{CAKE}^{\text{(HM)}}$ & \textbf{0.409} & \textbf{0.754} & \textbf{0.748} & \textbf{0.754} & \textbf{0.821} \\
\end{longtable}
\twocolumn

\newpage
\onecolumn

\normalsize
\section{Effect of Ensemble Diversity on CAKE Score Quality}\label{app:sec:divens}

\noindent The analysis in this section investigates how the construction of the clustering ensemble (\S \ref{subsec:notation}) influences the quality of the confidence scores produced by CAKE (Eq.~\ref{eq:cake}). The objective is to assess whether the proposed score (Eq.~\ref{eq:cake:b}) remains informative under different homogeneous and heterogeneous ensemble constructions (Fig.~\ref{fig:ensembles-real-synth}), and to clarify how changes in ensemble diversity relate to the quality of the induced confidence ranking. Rather than seeking a single universally optimal ensemble design, the goal is to characterize the extent to which the practical usefulness of CAKE depends on the way the ensemble is formed.

\medskip

\noindent For each dataset, multiple clustering ensembles were generated using different clustering methods and parameter settings. The considered constructions include both homogeneous ensembles, in which all runs are produced by the same clustering method, and heterogeneous ensembles, in which runs from different algorithms are combined within the same ensemble. More specifically, the analysis includes centroid-based constructions derived from \emph{k}-means, probabilistic constructions derived from Gaussian mixture models, and hierarchical constructions derived from BIRCH, together with heterogeneous combinations of these approaches.

\medskip

\noindent Variation within each clustering ``family'' was introduced through changes in key model settings, such as initialization choices, covariance structure, and threshold parameters, depending on the method under consideration. This design allows the effect of moderate variation within each construction, as well as stronger variation introduced by combining distinct clustering methods, to be examined in a unified way. For every ensemble construction, the number of runs was kept fixed at $R=20$ and the number of clusters was set to the ground-truth number of classes (Table~\ref{tab:realdata}). To reduce sensitivity to random initialization, the complete procedure was repeated across multiple random seeds and the reported values were averaged over these repetitions.

\medskip

\noindent Ensemble diversity was quantified through the disagreement among the partitions generated by the ensemble members. Let $\mathcal{E}=\{z^{(1)},z^{(2)},\dots,z^{(R)}\}$ denote an ensemble of \(R\) clustering runs, where \(z^{(r)}\) is the assignment vector produced by run \(r\). For each pair of runs, the normalized mutual information (NMI) between the corresponding partitions was computed, and ensemble diversity was defined as:
\begin{equation}
D(\mathcal{E}) \;=\; 1 - \frac{2}{R(R-1)} \sum_{r<s} \mathrm{NMI}\!\left(z^{(r)}, z^{(s)}\right).
\end{equation}
Under this definition, smaller values indicate that the ensemble members are highly similar to each other, whereas larger values indicate stronger disagreement and therefore a more diverse clustering ensemble. Since NMI is invariant to label permutations, this measure provides a label-independent summary of ensemble variability.

\medskip

\noindent For each ensemble construction, $\mathrm{CAKE}^\mathrm{HM}$ scores (Eq.~\ref{eq:cake:b}) were computed, yielding one confidence value per instance. The quality of the resulting confidence scores was then evaluated through a ranking-based coverage analysis. Instances were sorted from highest to lowest confidence, and for a given coverage level \(L\), the evaluation was restricted to the top-ranked \(L\%\) of the dataset. Clustering quality over this high-confidence portion was measured using the adjusted Rand index (ARI) between the corresponding consensus assignments (i.e., the final ensemble-level clustering labels obtained after aggregating the individual runs) and the ground-truth labels. This procedure produces an ARI--coverage curve as a function of the confidence coverage level. In the experiments, the coverage level was varied over the fixed grid $L \in \{0.40, 0.45, 0.50, \dots, 0.90\}.$
A single summary quantity, referred to as \emph{score quality}, was obtained as the normalized area under this ARI--coverage curve:
\begin{equation}
Q \;=\; \frac{1}{L_{\max}-L_{\min}} \int_{L_{\min}}^{L_{\max}} \mathrm{ARI}(L)\, dL .
\end{equation}
Intuitively, this quantity measures how well the confidence score orders instances according to assignment reliability. A high value indicates that points receiving higher confidence tend to have more reliable cluster assignments, so that the corresponding ARI remains high over the upper part of the confidence ranking.

\medskip

\noindent Illustratively, Fig.~\ref{fig:ensabl} reports, for the representative datasets, the mean CAKE score quality \(\bar{Q}\), the mean ARI of the consensus clustering on the dataset, and the mean ensemble diversity \(\bar{D}\) for each ensemble construction, all averaged across repetitions. This comparison makes it possible to examine how changes in ensemble diversity relate both to the overall clustering quality and to the usefulness of the confidence ranking induced by CAKE.

\begin{center}
  \includegraphics[width=\linewidth]{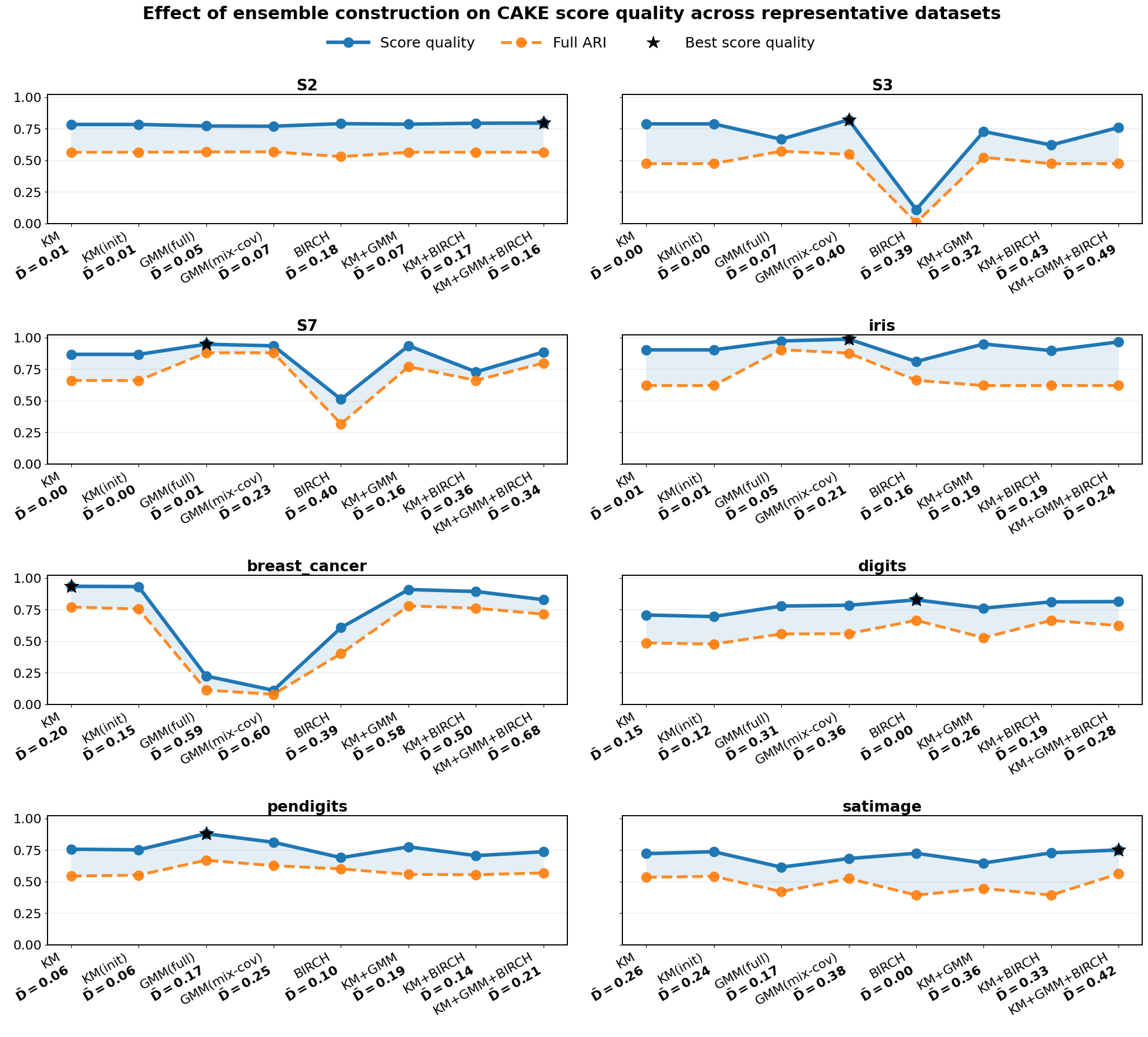}
  \captionsetup{hypcap=false}
  \captionof{figure}{Effect of ensemble construction on CAKE score quality across representative datasets. Each panel corresponds to one dataset. The \textcolor{MidnightBlue}{solid line} reports the mean score quality \(\bar{Q}\), while the \textcolor{orange}{dashed line} reports the mean ARI of the corresponding consensus clustering; both quantities are averaged across run repetitions. The black star marks the ensemble construction attaining the highest \(\bar{Q}\). The horizontal axis lists the ensemble constructions: KM: \emph{k}-means; KM (init): \emph{k}-means with varied initialization settings; GMM (full): GMMs with full covariance; GMM (mix-cov): GMMs with varying covariance structures; BIRCH: the corresponding hierarchical construction; and KM+GMM, KM+BIRCH, and KM+GMM+BIRCH: heterogeneous ensembles combining runs from the indicated clustering methods. Beneath each ensemble label, the corresponding mean ensemble diversity \(\bar{D}\) is also reported.}
  \label{fig:ensabl}
\end{center}

\noindent The results in Fig.~\ref{fig:ensabl} indicate that the relation between ensemble diversity and CAKE score quality is nuanced rather than monotonic. In several cases, more diverse or heterogeneous ensemble constructions lead to a more informative confidence ranking (e.g., S2, S3, \textit{Iris}, and \textit{Satimage}), suggesting that structured variation among ensemble members can help CAKE better identify reliable assignments. At the same time, the figure also shows that increased diversity is not beneficial by itself: on some datasets, homogeneous constructions remain more effective (e.g., S7 and \textit{Pendigits}), while highly diverse or poorly matched constructions can degrade both clustering quality and score quality (e.g., \textit{Breast Cancer}). Additionally, in all the cases shown in Fig.~\ref{fig:ensabl}, the mean score quality is higher than the mean consensus-clustering ARI. This is expected because the two quantities describe different aspects of performance. The mean consensus-clustering ARI summarizes the quality of the final consensus clustering over the full dataset, averaged across repetitions, whereas score quality summarizes how well CAKE orders instances by looking at clustering quality over the higher-confidence retained subsets across coverage levels. At the same time, the two quantities often show similar trends across ensemble constructions, suggesting that stronger overall clustering structure is often accompanied by a more informative confidence ranking. The consistent gap therefore indicates that the points ranked near the top by CAKE tend to be more reliably assigned and easier to cluster than the dataset as a whole.

\medskip

\noindent Taken together, these findings support the view that CAKE is robust across a wide range of ensemble constructions, but that the benefit of additional diversity depends on whether it captures useful complementary information rather than introducing unstable or incompatible clustering behavior.

\section{Adaptive CAKE filtering}\label{app:sec:adafilter}

\noindent To complement the fixed-coverage instance-removal experiment in \S\ref{subse:evalset}, an additional adaptive thresholding analysis is included here. In the main experiment, filtering is evaluated at a fixed retained coverage of 70\%, which provides a controlled and directly comparable setting across datasets and confidence criteria, and is also practically relevant when a target retained fraction is specified in advance. This evaluation setting isolates ranking quality itself and allows it to be assessed separately from threshold-selection effects, since CAKE is intended primarily as a pointwise confidence score that induces a ranking over instances. A natural follow-up question is whether the retained fraction can also be selected in a fully data-driven manner rather than fixed in advance. The present section addresses that question.

\medskip

\noindent Rather than fixing the retained fraction \textit{a priori}, the retained coverage is selected automatically from a predefined grid
\[
c \in \left\{0.40, 0.45, 0.50, \dots, 0.90\right\}.
\]
For each candidate coverage \(c\), the top-\(c\) fraction of instances is retained according to the ranking induced by \(\mathrm{CAKE}^{(\mathrm{HM})}\) (Eq.~\ref{eq:cake:b}), \emph{k}-means is re-fitted on the retained subset, and the resulting partition is evaluated using the Silhouette score.  The selected coverage is the one that maximizes this internal, label-free clustering-quality criterion over the subsets induced by the CAKE ranking. The standard sample-averaged Silhouette is used here, so the adaptive rule reflects overall subset quality at the instance level. In settings where balanced per-cluster retention is of greater interest, a macro-averaged variant could also be considered by first averaging Silhouette values within each cluster and then averaging across clusters, thereby reducing the influence of cluster size on coverage selection. After this retained size is determined, all competing filtering criteria (\S\ref{subse:evalset}) are evaluated at the same matched subset size. This yields a simple adaptive filtering rule that remains fully unsupervised while avoiding a fixed global threshold.

\medskip

\noindent This experiment serves a different purpose from the fixed-coverage analysis. The fixed-coverage analysis evaluates CAKE under a controlled retained size that is shared across methods, so that the comparison focuses directly on ranking quality. The adaptive analysis instead considers the setting in which the retained fraction is selected automatically from the data, and asks whether CAKE remains competitive under such a data-driven selection rule. Because the retained fraction is chosen by optimizing an internal clustering-validity criterion over a grid of candidate coverages, this analysis is presented as a supplementary adaptive filtering experiment. In particular, because the adaptive threshold is determined through the internal Silhouette objective, the comparison naturally becomes more favorable to the geometric component \(\tilde S\) than in the fixed-coverage setting. This makes the adaptive setting a useful stress test: strong CAKE performance under this protocol is particularly informative, since the selection criterion itself is more favorable to geometry-driven filtering.

\medskip

\noindent Table~\ref{tab:adaptive_filtering_selected} reports representative results on six real-world datasets chosen to summarize the main empirical patterns observed under adaptive coverage selection: \textsc{Bc}, \textsc{Dg}, \textsc{Ng}, \textsc{Fm}, \textsc{Pd}, and \textsc{Sa}. Together, these datasets span multiple modalities, lower- and higher-dimensional settings, smaller and larger sample sizes, and problems with varying levels of structural complexity and class overlap. This allows the adaptive filtering behavior of CAKE to be examined under conditions in which the relative contribution of geometry and assignment stability may differ.

\medskip

\noindent Several trends are consistent across these results. CAKE remains highly competitive under adaptive threshold selection, even though the procedure itself is based on internal Silhouette and therefore tends to strengthen geometry-based filtering rules. Under this more favorable setting for the geometric baseline, \(\mathrm{CAKE}^{(\mathrm{HM})}\) still attains the best performance on all three external metrics on \textsc{Dg} and \textsc{Pd}, while \(\mathrm{CAKE}^{(\mathrm{PR})}\) attains the best performance on all three metrics on \textsc{Ng} and \textsc{Fm}. In addition, \(\mathrm{CAKE}^{(\mathrm{HM})}\) remains favorable on \textsc{Sa}, where it achieves the best performance on two of the three reported metrics. Exact ties with the geometric baseline are observed on \textsc{Bc}. At the same time, the confidence-based filtering criteria consistently remain well above the random baseline across all six datasets, indicating that the gains are not simply a consequence of using a smaller retained subset, but reflect a meaningful ordering of points by estimated assignment reliability.

\medskip

\noindent Overall, the experiment supports the same qualitative conclusion as the fixed-coverage analysis: CAKE continues to provide a strong pointwise confidence ranking under an automatically chosen retained fraction, while the strongest individual filtering rule may still depend on dataset structure. 

\footnotesize
\begin{longtable}{llcccc}
\caption{Adaptive filtering with coverage selected by maximizing the Silhouette score over a fixed grid of retained fractions. For each dataset, the retained size is chosen from the $\mathrm{CAKE}^{(\mathrm{HM})}$ ranking, and all competing filtering criteria are then evaluated at the same matched subset size. ARI, AMI, and ACC are reported as means over multiple independent runs, with $95\%$ Student's $t$-confidence intervals shown in brackets. The highest reported value for each metric is shown in \textbf{bold}.}
\label{tab:adaptive_filtering_selected}\\

\toprule
\textbf{Dataset} & \textbf{Subset} & \textbf{Retained} & \textbf{ARI} & \textbf{AMI} & \textbf{ACC} \\
\midrule
\endfirsthead

\toprule
\textbf{Dataset} & \textbf{Subset} & \textbf{Retained} & \textbf{ARI} & \textbf{AMI} & \textbf{ACC} \\
\midrule
\endhead

\endfoot
\bottomrule
\endlastfoot

\multirow{7}{*}{\textsc{Bc}}
& Full & 569/569  & 0.663 [0.6558, 0.6698] & 0.543 [0.5343, 0.5518] & 0.908 [0.9058, 0.9101] \\
& Random & 228/569  & 0.612 [0.6071, 0.6175] & 0.484 [0.4786, 0.4892] & 0.893 [0.8909, 0.8942] \\
& Consensus & 228/569  & 0.831 [0.8233, 0.8393] & 0.730 [0.7139, 0.7469] & 0.957 [0.9553, 0.9596] \\
& $C$ component & 228/569  & 0.831 [0.8233, 0.8393] & 0.730 [0.7139, 0.7469] & 0.957 [0.9553, 0.9596] \\
& $\tilde S$ component & 228/569  & \textbf{0.924} [0.9239, 0.9239] & \textbf{0.851} [0.8515, 0.8515] & \textbf{0.987} [0.9868, 0.9868] \\
& $\mathrm{CAKE}^{\text{(PR)}}$ & 228/569 & \textbf{0.924} [0.9239, 0.9239] & \textbf{0.851} [0.8515, 0.8515] & \textbf{0.987} [0.9868, 0.9868] \\
& $\mathrm{CAKE}^{\text{(HM)}}$ & 228/569  & \textbf{0.924} [0.9239, 0.9239] & \textbf{0.851} [0.8515, 0.8515] & \textbf{0.987} [0.9868, 0.9868] \\

\cmidrule(lr){2-6}

\multirow{7}{*}{\textsc{Dg}}
& Full & 1797/1797 & 0.630 [0.6032, 0.6568] & 0.731 [0.7204, 0.7421] & 0.747 [0.7153, 0.7785] \\
& Random & 719/1797 & 0.538 [0.5051, 0.5711] & 0.677 [0.6563, 0.6967] & 0.671 [0.6255, 0.7155] \\
& Consensus & 719/1797 & 0.766 [0.7140, 0.8187] & 0.833 [0.8098, 0.8560] & 0.769 [0.7269, 0.8105] \\
& $C$ component & 719/1797 & 0.738 [0.6838, 0.7913] & 0.821 [0.7969, 0.8450] & 0.727 [0.6801, 0.7736] \\
& $\tilde S$ component & 719/1797 & 0.808 [0.7486, 0.8673] & 0.901 [0.8763, 0.9266] & 0.780 [0.7281, 0.8316] \\
& $\mathrm{CAKE}^{\text{(PR)}}$ & 719/1797 & 0.746 [0.6787, 0.8125] & 0.884 [0.8602, 0.9074] & 0.731 [0.6735, 0.7885] \\
& $\mathrm{CAKE}^{\text{(HM)}}$ & 719/1797 & \textbf{0.823} [0.7407, 0.9046] & \textbf{0.911} [0.8772, 0.9451] & \textbf{0.788} [0.7028, 0.8733] \\

\cmidrule(lr){2-6}

\multirow{7}{*}{\textsc{Ng}}
& Full & 18846/18846 & 0.352 [0.3433, 0.3612] & 0.509 [0.5006, 0.5179] & 0.513 [0.4992, 0.5259] \\
& Random & 7538/18846 & 0.356 [0.3477, 0.3638] & 0.514 [0.5076, 0.5212] & 0.513 [0.4948, 0.5313] \\
& Consensus & 7538/18846 & 0.465 [0.4459, 0.4844] & 0.580 [0.5775, 0.5823] & 0.532 [0.5205, 0.5427] \\
& $C$ component & 7538/18846 & 0.441 [0.4201, 0.4612] & 0.570 [0.5685, 0.5721] & 0.488 [0.4722, 0.5036] \\
& $\tilde S$ component & 7538/18846 & 0.579 [0.5458, 0.6112] & 0.699 [0.6904, 0.7081] & 0.629 [0.6040, 0.6533] \\
& $\mathrm{CAKE}^{\text{(PR)}}$ & 7538/18846 & \textbf{0.592} [0.5602, 0.6246] & \textbf{0.701} [0.6877, 0.7134] & \textbf{0.639} [0.6090, 0.6695] \\
& $\mathrm{CAKE}^{\text{(HM)}}$ & 7538/18846 & 0.567 [0.5310, 0.6030] & 0.688 [0.6754, 0.7006] & 0.626 [0.5921, 0.6603] \\

\cmidrule(lr){2-6}

\multirow{7}{*}{\textsc{Fm}}
& Full & 60000/60000 & 0.353 [0.3416, 0.3641] & 0.506 [0.4966, 0.5163] & 0.514 [0.4880, 0.5391] \\
& Random & 24000/60000 & 0.356 [0.3433, 0.3681] & 0.517 [0.5071, 0.5260] & 0.505 [0.4835, 0.5263] \\
& Consensus & 24000/60000 & 0.520 [0.4765, 0.5629] & 0.598 [0.5798, 0.6165] & 0.572 [0.5318, 0.6122] \\
& $C$ component & 24000/60000 & 0.397 [0.3822, 0.4120] & 0.593 [0.5818, 0.6051] & 0.492 [0.4805, 0.5038] \\
& $\tilde S$ component & 24000/60000 & 0.540 [0.5070, 0.5724] & 0.653 [0.6422, 0.6634] & 0.599 [0.5660, 0.6311] \\
& $\mathrm{CAKE}^{\text{(PR)}}$ & 24000/60000 & \textbf{0.560} [0.5263, 0.5932] & \textbf{0.671} [0.6543, 0.6878] & \textbf{0.634} [0.6094, 0.6596] \\
& $\mathrm{CAKE}^{\text{(HM)}}$ & 24000/60000 & 0.541 [0.5015, 0.5803] & 0.663 [0.6468, 0.6793] & 0.620 [0.5774, 0.6631] \\

\cmidrule(lr){2-6}

\multirow{7}{*}{\textsc{Pd}}
& Full & 10992/10992 & 0.541 [0.5121, 0.5690] & 0.676 [0.6667, 0.6846] & 0.669 [0.6326, 0.7057] \\
& Random & 6046/10992 & 0.529 [0.5083, 0.5496] & 0.672 [0.6644, 0.6793] & 0.670 [0.6404, 0.6992] \\
& Consensus & 6046/10992 & 0.727 [0.6996, 0.7553] & 0.796 [0.7898, 0.8024] & 0.739 [0.7083, 0.7690] \\
& $C$ component & 6046/10992 & 0.714 [0.6913, 0.7361] & 0.789 [0.7720, 0.8050] & 0.741 [0.7156, 0.7665] \\
& $\tilde S$ component & 6046/10992 & 0.756 [0.7179, 0.7942] & 0.858 [0.8423, 0.8737] & 0.779 [0.7454, 0.8132] \\
& $\mathrm{CAKE}^{\text{(PR)}}$ & 6046/10992 & 0.723 [0.6647, 0.7821] & 0.833 [0.8093, 0.8559] & 0.756 [0.7095, 0.8019] \\
& $\mathrm{CAKE}^{\text{(HM)}}$ & 6046/10992 & \textbf{0.771} [0.7213, 0.8203] & \textbf{0.861} [0.8418, 0.8804] & \textbf{0.796} [0.7533, 0.8393] \\

\cmidrule(lr){2-6}

\multirow{7}{*}{\textsc{Sa}}
& Full & 6430/6430 & 0.510 [0.4657, 0.5550] & 0.595 [0.5556, 0.6346] & 0.663 [0.6298, 0.6972] \\
& Random & 2894/6430  & 0.524 [0.4974, 0.5504] & 0.614 [0.5951, 0.6333] & 0.670 [0.6526, 0.6869] \\
& Consensus & 2894/6430  & 0.540 [0.5008, 0.5793] & 0.597 [0.5799, 0.6134] & 0.712 [0.6715, 0.7524] \\
& $C$ component & 2894/6430  & 0.540 [0.5008, 0.5793] & 0.597 [0.5799, 0.6134] & 0.712 [0.6715, 0.7524] \\
& $\tilde S$ component & 2894/6430  & 0.722 [0.6517, 0.7919] & \textbf{0.780} [0.7466, 0.8125] & 0.806 [0.7618, 0.8495] \\
& $\mathrm{CAKE}^{\text{(PR)}}$ & 2894/6430  & 0.668 [0.5904, 0.7449] & 0.748 [0.7020, 0.7946] & 0.760 [0.6996, 0.8209] \\
& $\mathrm{CAKE}^{\text{(HM)}}$ & 2894/6430  & \textbf{0.731} [0.6573, 0.8048] & 0.778 [0.7373, 0.8191] & \textbf{0.808} [0.7478, 0.8679] \\

\end{longtable}
\normalsize
\newpage

\section{Comparison Against GMM Confidence Signals}
\label{app:sec:gmm_confidence}

\noindent As a complement to the pointwise \mbox{AUPRC}/\mbox{AUROC} experiments in Tables~\ref{tab:pointwise_all} and~\ref{tab:pointwise_fcm_densitybased}, additional comparisons are reported here for native GMM-based confidence signals. In this setting, correctness is defined with respect to a reference Gaussian Mixture Model (GMM) fitted at the ground-truth number of components $k$. Let $\hat z_i^{\mathrm{GMM}}$ denote the label assigned to point $x_i$ by the reference GMM, and let $\pi$ denote the Hungarian alignment mapping to ground-truth labels. The binary target is then $\mathds{1}\!\left[\pi(\hat z_i^{\mathrm{GMM}})=y_i\right]$. As in the main pointwise evaluation, discriminative performance is measured using \mbox{AUPRC} and \mbox{AUROC}.

\medskip

\noindent Three GMM confidence scores were considered. Let $p_{i\ell}$ denote the posterior responsibility of point $x_i$ for component $\ell$, with $\sum_{\ell=1}^{k} p_{i\ell}=1$. The first score is maximum posterior confidence,
\begin{equation}
\mathrm{GMM}^{(1)}_i=\max_{\ell} p_{i\ell},
\end{equation}
the second is posterior margin,
\begin{equation}
\mathrm{GMM}^{(2)}_i=p_{i(1)}-p_{i(2)},
\end{equation}
where $p_{i(1)}\ge p_{i(2)}$ are the two largest posterior responsibilities, and the third is entropy-based confidence,
\begin{equation}
\mathrm{GMM}^{(3)}_i
=
1-\frac{H_i}{\log k},
\qquad
H_i
=
-\sum_{\ell=1}^{k} p_{i\ell}\log\!\bigl(p_{i\ell}+\varepsilon\bigr).
\end{equation}

\noindent These scores are compared against $\mathrm{CAKE}^{(\mathrm{HM})}$ (Eq.~\ref{eq:cake:b}) computed from a GMM ensemble. Specifically, an ensemble of $R=20$ GMM runs is constructed using the same number of mixture components $k$ and different random seeds, and $\mathrm{CAKE}^{(\mathrm{HM})}$ scores are computed from the resulting assignments in the same way as in the main experiments. The corresponding results for the real-world datasets (Table~\ref{tab:realdata}) are reported in Table~\ref{tab:gmm_confidence}.

\begin{center}
\small
\setlength{\tabcolsep}{6pt}
\begin{tabular}{llcccc}
\toprule
Dataset & Metric & $\mathrm{GMM}^{(1)}$ & $\mathrm{GMM}^{(2)}$ & $\mathrm{GMM}^{(3)}$ & $\mathrm{CAKE}^{(\mathrm{HM})}$\\

\midrule
\multirow{2}{*}{\textsc{Ir}} & AUPRC & \textbf{0.997} & \textbf{0.997} & \textbf{0.997} & 0.996 \\
                             & AUROC & \textbf{0.927} & \textbf{0.927} & \textbf{0.927} & 0.919 \\
\cmidrule(lr){2-6}

\multirow{2}{*}{\textsc{Bc}} & AUPRC & 0.756 & 0.756 & 0.758 & \textbf{0.947} \\
                             & AUROC & 0.641 & 0.641 & 0.641 & \textbf{0.915} \\
\cmidrule(lr){2-6}

\multirow{2}{*}{\textsc{Dg}} & AUPRC & 0.847 & 0.848 & 0.841 & \textbf{0.967} \\
                             & AUROC & 0.533 & 0.537 & 0.511 & \textbf{0.853} \\
\cmidrule(lr){2-6}

\multirow{2}{*}{\textsc{Ng}} & AUPRC & 0.628 & 0.638 & 0.610 & \textbf{0.639} \\
                             & AUROC & 0.659 & 0.664 & 0.645 & \textbf{0.687} \\
\cmidrule(lr){2-6}

\multirow{2}{*}{\textsc{Fm}} & AUPRC & 0.418 & 0.418 & 0.413 & \textbf{0.684} \\
                             & AUROC & 0.500 & 0.500 & 0.487 & \textbf{0.715} \\
\cmidrule(lr){2-6}

\multirow{2}{*}{\textsc{Pd}} & AUPRC & 0.644 & 0.638 & 0.679 & \textbf{0.924} \\
                             & AUROC & 0.537 & 0.535 & 0.559 & \textbf{0.858} \\
\cmidrule(lr){2-6}

\multirow{2}{*}{\textsc{Lt}} & AUPRC & 0.518 & 0.516 & \textbf{0.544} & 0.444 \\
                             & AUROC & 0.686 & 0.686 & \textbf{0.689} & 0.609 \\
\cmidrule(lr){2-6}

\multirow{2}{*}{\textsc{Sa}} & AUPRC & 0.695 & 0.692 & 0.711 & \textbf{0.806} \\
                             & AUROC & 0.692 & 0.690 & 0.690 & \textbf{0.747} \\
\bottomrule
\end{tabular}
\captionsetup{hypcap=false}
\captionof{table}{Per-point GMM-correctness prediction (\mbox{AUPRC} and \mbox{AUROC}) with $\mathrm{GMM}^{(1)}$ (max posterior), $\mathrm{GMM}^{(2)}$ (posterior margin), $\mathrm{GMM}^{(3)}$ (entropy confidence), and $\mathrm{CAKE}^{(\mathrm{HM})}$ computed from GMM ensembles. The highest value for each dataset and metric is shown in \textbf{bold}.}
\label{tab:gmm_confidence}
\end{center}

\noindent The results in Table~\ref{tab:gmm_confidence} show that $\mathrm{CAKE}^{(\mathrm{HM})}$ performs more strongly on most of the real-world datasets. In particular, it achieves the highest values on \textsc{Bc}, \textsc{Dg}, \textsc{Ng}, \textsc{Fm}, \textsc{Pd}, and \textsc{Sa}. This suggests that CAKE confidence scores derived from GMM ensembles can provide a more informative pointwise reliability signal than native posterior-based confidence scores alone.

\end{document}